\documentclass[11pt]{article}

\usepackage[preprint]{acl}

\usepackage{times}
\usepackage{latexsym}

\usepackage[T1]{fontenc}
\usepackage[utf8]{inputenc}

\usepackage{microtype}

\usepackage{inconsolata}

\usepackage{graphicx}

\usepackage{booktabs}
\usepackage{multirow}
\usepackage{colortbl}
\usepackage{arydshln}

\usepackage{amsmath}
\usepackage{subcaption}

\usepackage{placeins} % preamble

\title{Ability Transfer and Recovery via Modularized Parameters Localization}

\author{Songyao Jin, Kun Zhou\thanks{Corresponding Author}, Wenqi Li, Peng Wang, Biwei Huang \\
        University of California San Diego
         \\ \texttt{$\{$soj007, kuzhou, wel118, pew025, bih007$\}$@ucsd.edu}}

\begin{document}
\maketitle
\begin{abstract}
Large language models can be continually pre-trained or fine-tuned to improve performance in specific domains, languages, or skills, but this specialization often degrades other capabilities and may cause catastrophic forgetting. We investigate how abilities are distributed within LLM parameters by analyzing module activations under domain- and language-specific inputs for closely related models. Across layers and modules, we find that ability-related activations are highly concentrated in a small set of channels (typically <5\%), and these channels are largely disentangled with good sufficiency and stability. Building on these observations, we propose ACT (Activation-Guided Channel-wise Ability Transfer), which localizes ability-relevant channels via activation differences and selectively transfers only the corresponding parameters, followed by lightweight fine-tuning for compatibility. Experiments on multilingual mathematical and scientific reasoning show that ACT can recover forgotten abilities while preserving retained skills. It can also merge multiple specialized models to integrate several abilities into a single model with minimal interference. %Overall, ACT achieves competitive or superior performance to fine-tuning and model-merging baselines while updating only a small fraction of parameters per language–task setting.
Our code and data will be publicly released.
\end{abstract}

\section{Introduction}

Large language models (LLMs) pretrained on large-scale corpora have demonstrated strong general-purpose capabilities across a wide range of tasks~\citep{brown2020languagemodelsfewshotlearners,touvron2023llama2openfoundation,zhao2025surveylargelanguagemodels}.
To further enhance performance in specific domains, languages and skills, these LLMs can be continually pre-trained or fine-tuned on related datasets~\citep{ouyang2022traininglanguagemodelsfollow,rafailov2024directpreferenceoptimizationlanguage}.
However, such specialization often degrades other abilities, and even causes catastrophic forgetting~\citep{luo2025empiricalstudycatastrophicforgetting} on relevant tasks.

To overcome the ability degradation issue, existing work mostly focuses on devising new fine-tuning strategy with special regularization~\citep{Kirkpatrick_2017} and parameter frozen methods~\citep{houlsby2019parameterefficienttransferlearningnlp}.
However, due to the unawareness of the ability knowledge distribution, it is challenging to disentangle the targeted and irrelevant abilities within the LLM.
Moreover, these method can not directly work on the ability recovery of fine-tuned LLMs, limiting the broader application on growing scenarios in the community.

In this work, we study how different abilities are stored in LLMs, and find a more efficient and effective way to transfer or recovery them.
To this end, we focus on localizing the ability-related parameters and test the sufficiency and stability.
Existing work~\citep{zhang2023emergentmodularitypretrainedtransformers,du2024unlockingcontinuallearningabilities,kumar2024shared} about the modularity of LLMs, has revealed that their inner parameters and activations are potentially highly disentangled for storing and representing different knowledge, respectively.
Inspired by it, we start from analyzing activation values within LLMs
\footnote{In this paper, \emph{activations} refer to outputs of trainable modules, such as attention (Q/K/V/O) and MLP (gate/up/down) projections, measured along their output channels.}.
By feeding domain- and language-specific samples and comparing closely related models, we empirically observe that the ability-related activations are highly concentrated on very few channels (e.g., $<5\%$), spanning all layers and modules.
Surprisingly, the ability-specific channels are highly disentangled with each other, and are very stable and consistent to resist fine-tuning.
It indicates that pre-trained LLMs contain modularized parameters for specific ability, which is promising to help efficiently and effectively transfer or recovery the target abilities.
%only a small subset of output channels exhibits large activation differences for a given language–task ability.
%These channels are largely disentangled across abilities, remain consistent across models derived from the same pretrained checkpoint, and typically do not correspond to the parameters with the largest weight changes during fine-tuning.

Motivated by these observations, we introduce \textbf{ACT} (\textbf{A}ctivation-Guided \textbf{C}hannel-wise Ability \textbf{T}ransfer), a selective ability transfer framework that localizes ability-relevant channels using cross-model activation differences and restricts parameter transfer to these channels. ACT performs ability transfer through activation-guided channel selection followed by masked task-vector merging. We additionally apply lightweight post-transfer fine-tuning to improve compatibility, and empirically find that the transferred parameters substantially enhance fine-tuning effectiveness. 

Extensive experiments show that ACT effectively recovers forgotten abilities while preserving retained skills.
In addition, experiments on merging multiple specialized models demonstrate that ACT can integrate several retained abilities into a single model.
Compared with existing fine-tuning and model merging baselines, ACT achieves competitive or superior performance while modifying only a small fraction of parameters, and enables multi-ability integration with minimal interference.
%\textcolor{blue}{xxx data xxx performance, xxx weights xxx degrade}

\section{Related Work}

\paragraph{Modularity in LLMs.}
A growing body of work suggests that large neural networks exhibit sparse and task-relevant subnetworks~\citep{frankle2019lotterytickethypothesisfinding, sanh2020movementpruningadaptivesparsity}.
As model scale increases, representations associated with different tasks or classes become increasingly orthogonal, indicating emergent functional separation~\citep{ramasesh2021effect}.
Activation- and gradient-based analyses further show that parameters in deeper layers of language models become increasingly sparse and specialized across domains and tasks~\citep{wang2024exploringactivationpatternsparameters, du2024unlockingcontinuallearningabilities}.
More direct evidence of modularity comes from interpretability studies of Transformers, which reveal that a small subset of neurons, channels, or components specialize in distinct semantic, factual, or algorithmic functions~\citep{zhang2023emergentmodularitypretrainedtransformers, elhage2021mathematical, olsson2022incontextlearninginductionheads, wang2025transformersrichapproximationdynamics}.
Recent dictionary learning and sparse decomposition approaches further show that model activations can be expressed in terms of approximately disentangled, monosemantic features, despite the presence of superposition~\citep{bricken2023monosemanticity, templeton2024scaling, elhage2022toymodelssuperposition, yan2024encourageinhibitmonosemanticityrevisit}.
Together, these findings indicate that model behaviors may be modular, motivating the use of activation-level signals to identify ability-relevant structures.
%Building on this perspective, our work leverages activation differences to localize specific modularized parameters for ability transfer and recovery.

% \paragraph{Modularity in LLMs.}
% A growing body of work suggests that neural networks exhibit sparse and task-relevant subnetworks~\citep{frankle2019lotterytickethypothesisfinding, sanh2020movementpruningadaptivesparsity}.
% In pretrained models, representations associated with different classes become increasingly orthogonal as model scale grows~\citep{ramasesh2021effect}.
% Gradient-based analyses further show that when inputs come from different domains, parameters in shallow layers tend to exhibit more similar behavior, whereas deeper layers become increasingly sparse and specialized~\citep{wang2024exploringactivationpatternsparameters}.
% In addition, the $\ell_1$-normalized output magnitude distributions of linear layers differ substantially across tasks in language models~\citep{du2024unlockingcontinuallearningabilities}.
% Together, these findings indicate that model behaviors may be modular, motivating the use of activation-level signals to identify ability-relevant structures.
% Building on this perspective, our work leverages activation differences to localize specific modularized parameters for ability transfer and recovery.

\paragraph{Catastrophic Forgetting.}
Catastrophic forgetting (CF) has been extensively studied, where models continually trained on new data or tasks tend to degrade performance on previously learned tasks.
Early work addresses CF through parameter regularization, often formulated as approximate Bayesian inference over model parameters~\citep{Kirkpatrick_2017, farquhar2019unifyingbayesianviewcontinual}.
Parameter-efficient approaches, such as adapters, introduce task-specific trainable modules to reduce interference across tasks~\citep{houlsby2019parameterefficienttransferlearningnlp}.
Beyond performance degradation, several studies show that internal representations can still drift substantially even when task accuracy is preserved~\citep{davari2022probingrepresentationforgettingsupervised, wu2022pretrained}.
Recent empirical analyses confirm that CF also arises during post-training LLMs~\citep{luo2025empiricalstudycatastrophicforgetting}.
%Existing approaches mostly aim to mitigate forgetting during training, while the post-hoc recovery of forgotten abilities remains largely unexplored.
%are developed and evaluated in classical continual learning settings, and primarily 

\paragraph{Model Merging.}
Model merging aims to combine multiple task-specific models into one generalist model.
Naive parameter averaging often leads to severe performance degradation~\citep{wortsman2022modelsoupsaveragingweights}.
To address this, several weighted merging methods have been proposed, including Fisher-Merging~\citep{matena2022mergingmodelsfisherweightedaveraging} and RegMean~\citep{jin2025datalessknowledgefusionmerging}, which compute merging coefficients based on parameter-level statistics such as Fisher information or inner-product matrices.
Task Arithmetic~\citep{ilharco2023editingmodelstaskarithmetic} represents task-specific behaviors as task vectors, the differences between fine-tuned and pretrained models, and composes them through linear addition.
Subsequent methods, such as TIES~\citep{yadav2023tiesmergingresolvinginterferencemerging}, DARE~\citep{yu2024languagemodelssupermario}, and LiNeS~\citep{wang2025linesposttraininglayerscaling}, further mitigate interference using parameter-space heuristics, including trimming, random dropping, or layer-wise scaling.
%Despite these, most existing approaches operate primarily in the parameter space and perform merging in a largely global manner, which may inadvertently transfer undesired behaviors together with target capabilities.
%In this work, we leverage activation-level differences to localize ability-relevant modularized parameters and merge them for more efficient ability combination.

\section{Empirical Analysis of Cross-Model Activation Differences}

In this section, we empirically analyze cross-model activation differences to characterize how activations vary among language or domain abilities.
Specifically, we focus on pair of closely related LLMs, and examine how activation differences change under inputs from different language-domain combinations.
This analysis aims to study whether these activation differences are informative to identify ability-specific modularized parameters.

\subsection{Analysis Setup}
\label{sec:analysis_setup}

\paragraph{Activation Extraction and Difference.}
To perform a comprehensive study, we extract \emph{activation vectors} from all layers in the LLM, including attention projections, MLP projections, layer normalization, token embedding layers, and the language modeling head.
An \emph{channel} corresponds to one dimension in the activation vector, which is a scalar computed typically by the dot product between one row of the weight matrix and the input hidden state.
%For linear projections, it is equivalent to a single row of the projection weight matrix (and its associated bias, if present), whose activation is the scalar output produced for an input token.
For each data sample, we record activations at every layer and token position.
Given a model pair and inputs for a specific ability, we compute token-level activation differences as
\begin{equation}
\Delta a_{i,t} = \left| a_{i,t}^{(\text{m1})} - a_{i,t}^{(\text{m2})} \right|,
\end{equation}
where $a_{i,t}^{(\cdot)}$ denotes the scalar activation of the $i$-th output channel at the $t$-th token position.
% We restrict this computation to answer tokens only.
For each output channel, we average activation differences over all answer tokens across samples, yielding a single scalar activation difference $\Delta a_i$ per channel for studying language-domain combination.

\paragraph{Model Pairs.}
We analyze activation differences for pair of related models that share the same pretrained backbone but differ in fine-tuning objectives.
Our analysis primarily focuses on two Qwen2.5 math model pairs at different scales: (1) Qwen2.5-Math-7B-Instruct vs.\ Qwen2.5-7B-Instruct, and (2) Qwen2.5-Math-1.5B-Instruct vs.\ Qwen2.5-1.5B-Instruct.
% All models support multilingual inference.
The math-specialized models exhibit strong performance on English and Chinese mathematical reasoning, but underperform their corresponding general-purpose instruct models in other domains and languages~\citep{yang2024qwen25mathtechnicalreportmathematical,qwen2025qwen25technicalreport}.
To assess the generality, we additionally analyze Qwen2.5-Coder models and LLaMA-2-13B-based fine-tuned variants.
Unless otherwise stated, we report representative results on Qwen2.5 math model pairs at 7B and 1.5B, while comprehensive results across different model families, scales, languages, and domains are in Appendix~\ref{app:additional_empirical_analysis}.

% For comparison between model pairs, we also use Qwen2.5 Coder models. Moreover, we analyze other model family (e.g., LLaMA2). The complete results are in \textcolor{blue}{Appendix~X}.

% Additional results on other model families (e.g., LLaMA2) are provided in \textcolor{blue}{Appendix~X}.

\paragraph{Languages and Domains.}
Our analysis primarily considers two domain-specific reasoning, mathematics and science, across 11 languages: English, Chinese, Arabic, Bengali, Hungarian, Japanese, Korean, Swahili, Telugu, Thai, and Vietnamese.
For each language-domain combination, both models in a pair are evaluated on the same set of input samples, ensuring that observed activation differences primarily reflect differences in model parameters rather than tokenization or input content.
For mathematical reasoning, we use the MetaMathQA dataset~\citep{yu2024metamathbootstrapmathematicalquestions}, translating 1{,}500 distinct samples per language to avoid biases from shared content.
For scientific reasoning, we adopt the MegaScience dataset~\citep{fan2025megasciencepushingfrontiersposttraining}, selecting physics, chemistry, and biology as representative subjects, and translating 500 samples per subject per language, resulting in 1{,}500 science samples per language.
All translations are performed using GPT-4o-mini~\citep{openai2024gpt4ocard}, except for mathematics in Telugu, which uses GPT-4o for improved quality.
Due to limited multilingual coverage, LLaMA2 experiments focus on English mathematics, science, and code-related domains.

% In the following, we report universal results using several representative combinations. For other more results and results on LLaMA2-based models in \textcolor{blue}{Appendix~X}. Due to limited multilingual coverage, LLaMA2 experiments focus on English mathematics, science, and code-related domains.

% For completeness, we additionally report results on LLaMA2-based models in \textcolor{blue}{Appendix~X}.
% Due to limited multilingual coverage, these experiments focus on English mathematics, science, and code-related tasks.

\begin{figure*}[htbp]
  \centering

  % -------- (a) Global --------
  \begin{subfigure}[t]{0.32\linewidth}
    \centering
    \includegraphics[width=\linewidth]{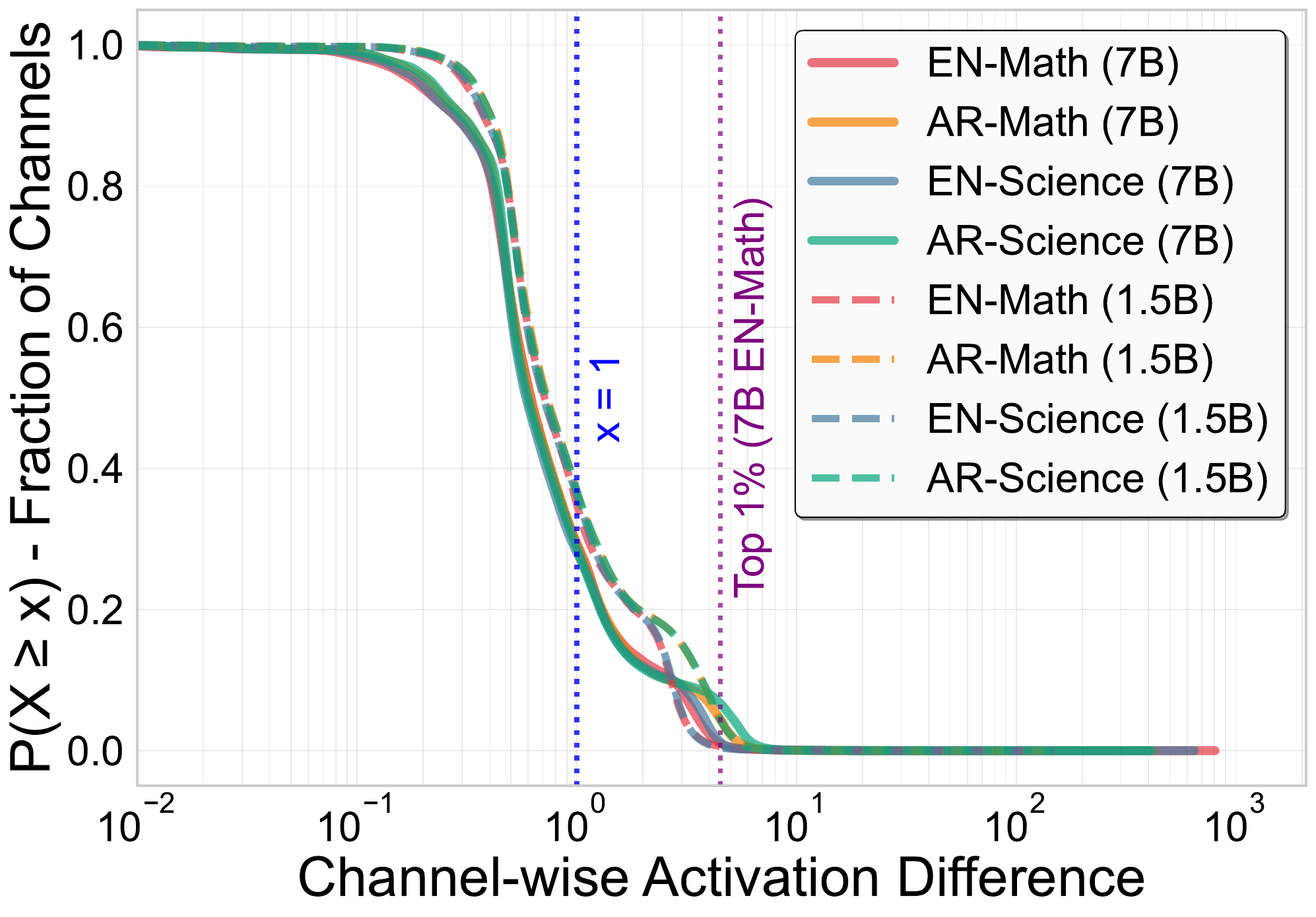}
    \caption{Global CCDF}
    \label{fig:activation_distribution}
  \end{subfigure}
  \hfill
  % -------- (b) Layer-wise --------
  \begin{subfigure}[t]{0.32\linewidth}
    \centering
    \includegraphics[width=\linewidth]{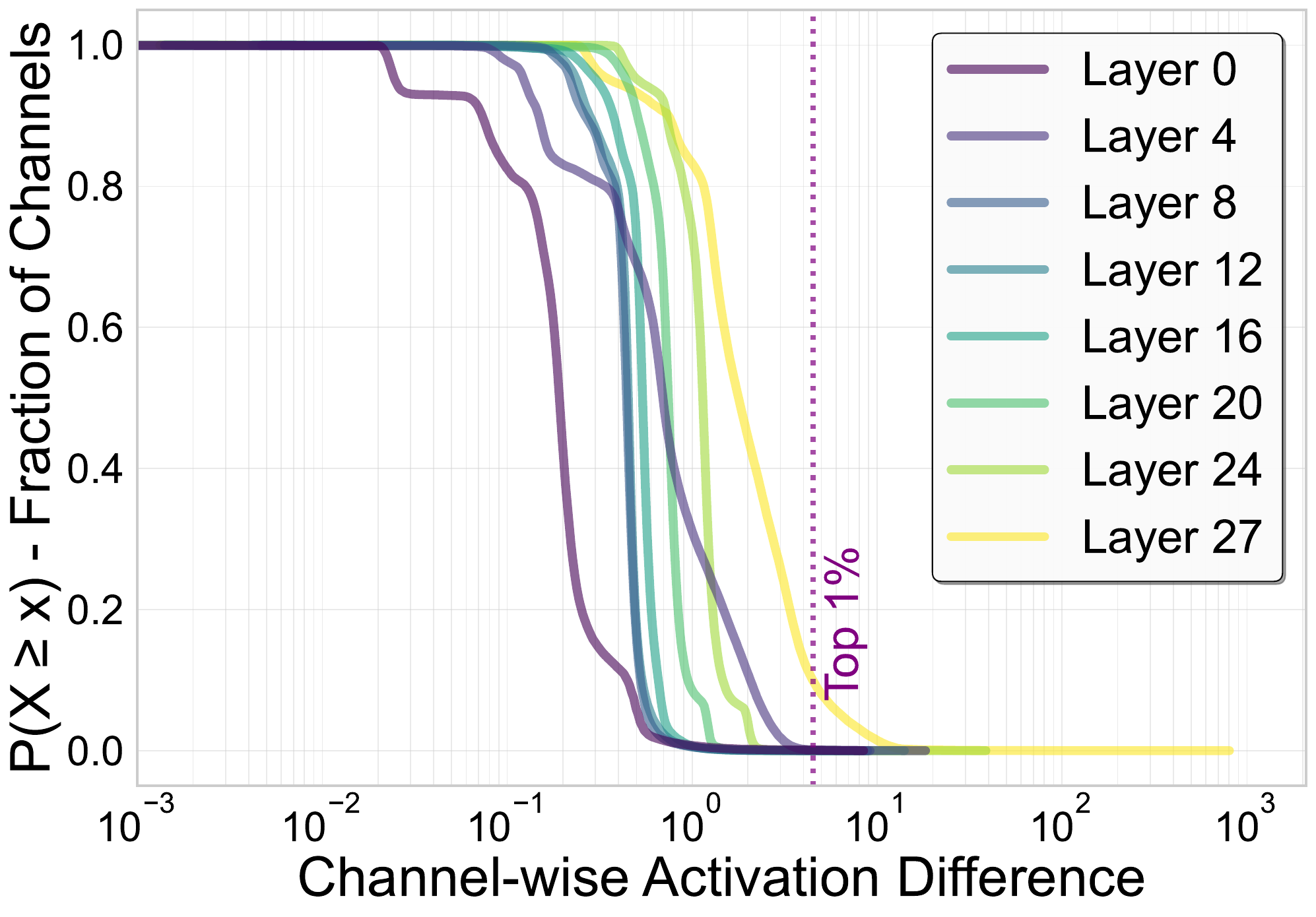}
    \caption{CCDF per layer (EN-Math 7B)}
    \label{fig:ccdf_by_layer}
  \end{subfigure}
  \hfill
  % -------- (c) Module-wise --------
  \begin{subfigure}[t]{0.32\linewidth}
    \centering
    \includegraphics[width=\linewidth]{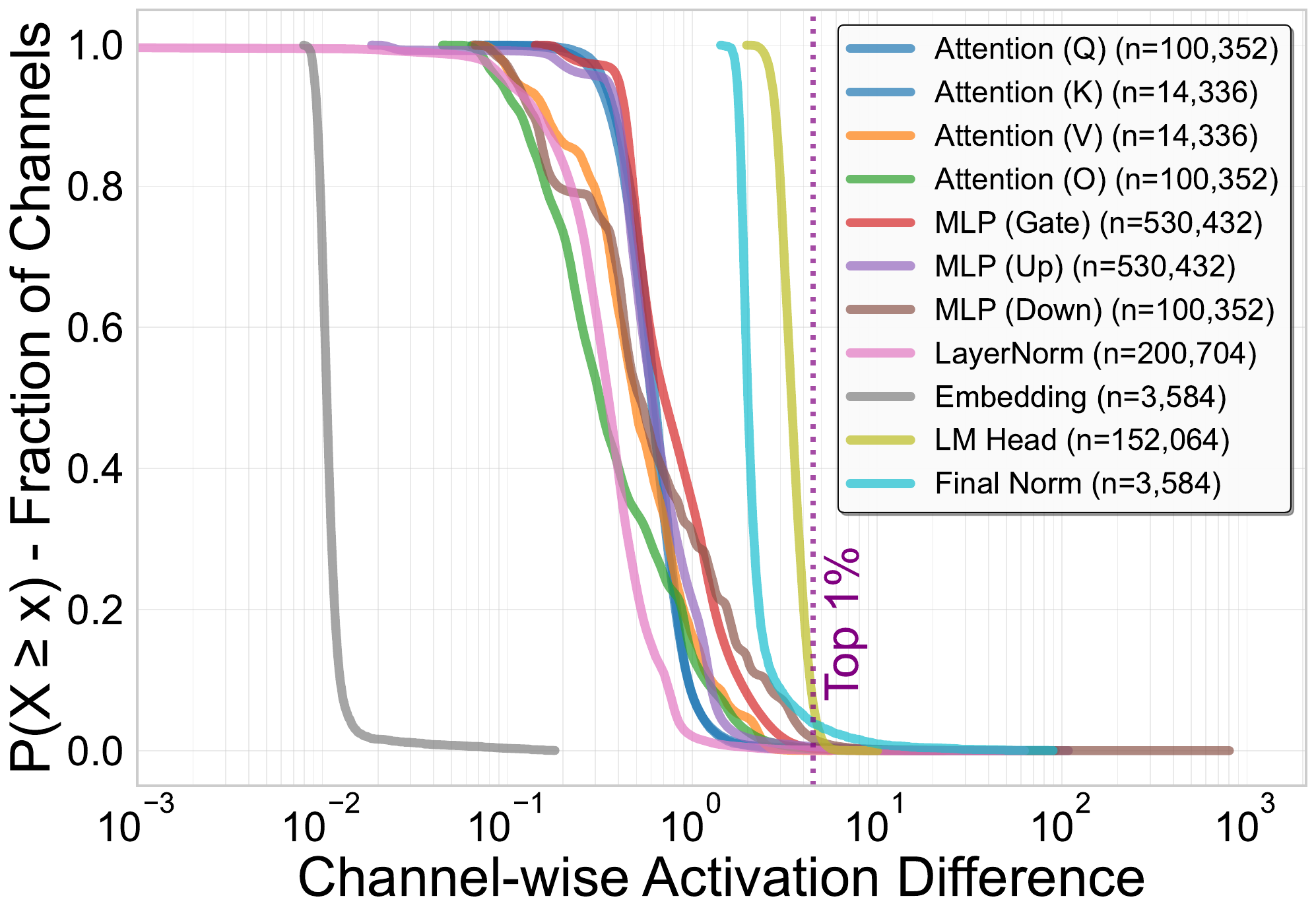}
    \caption{CCDF per module (EN-Math 7B)}
    \label{fig:ccdf_by_module}
  \end{subfigure}

\caption{
Complementary cumulative distribution function (CCDF), i.e., the fraction of output channels whose averaged activation difference exceeds a given threshold.
%(\textbf{a}) Global channel-wise CCDF aggregated across all layers and modules for representative language--domain combinations on Qwen2.5 math model pairs (7B and 1.5B).
%(\textbf{b,c}) The same CCDFs decomposed by decoder layer and module type, respectively (English math, Qwen2.5 7B).
Across all views, activation differences exhibit highly skewed, heavy-tailed distributions, with large deviations concentrated in a small subset of channels.
}
  \label{fig:activation_distribution_by_structure}
\end{figure*}

\subsection{Distribution of Activation Differences}

We first analyze the empirical distribution of activation differences across output channels, to study how large activation differences distribute across different layers and modules.
%obtained by averaging token-level differences within each channel.
% This yields a channel-wise statistic that characterizes how differently the two models respond internally to the same inputs.

\paragraph{Finding 1: Large activation differences are concentrated in a small fraction of channels.}
Figure~\ref{fig:activation_distribution} shows the empirical complementary cumulative distribution function (CCDF) of channel-wise activation differences for representative language-domain combinations, including English and Arabic for both mathematical and scientific reasoning.
% Results for additional combinations are reported in \textcolor{blue}{Appendix~X}.
Across all settings, the distributions are consistently skewed: approximately 99\% of output channels exhibit an averaged activation difference below 4.5, while only a small fraction displays substantially larger deviations.
This heavy-tailed behavior indicates that activation differences are concentrated in a small subset of channels.
This observation provides an intuitive basis for ability transfer.
If activation differences on high-deviation channels can be modeled and transferred, the ability gap between two models is promising to be bridged.
%become more closely aligned for the corresponding language-domain ability.
%We further observe that the 7B models tend to exhibit larger extreme activation differences than the 1.5B models, which is consistent with their greater depth and representational capacity.

\paragraph{Finding 2: Activation difference distributions across different layers exhibit consistent patterns.}
We further examine how activation differences are distributed across different layer and module types for a fixed language-domain combination.
As shown in Figure~\ref{fig:activation_distribution_by_structure}, both layer-wise and module-wise CCDFs exhibit highly consistent heavy-tailed patterns.
Although deeper layers and the language modeling head tend to show slightly larger extrema, no single layer or module dominates the distribution.
Overall, these results indicate that while large activation differences are sparse (Finding~1), they are broadly distributed throughout the network.

% Across layers, most channels display small activation differences, while a small fraction exhibits large deviations. 
% Although deeper layers tend to show mildly more and larger extrema, no single layer dominates the distribution.
% A similar pattern holds across module types: attention and MLP projections exhibit comparable distributions, with smaller activation differences in embedding layers and somewhat more large deviations in the LM head and its final normalization.
% Overall, these results indicate that while large activation differences are sparse (Finding~1), they are broadly distributed throughout the network. % with moderate concentration in the deep structure.
% % We observe the same qualitative trends across other language-domain combinations and additional model families (\textcolor{blue}{Appendix~X}).

\begin{table*}[htbp]
\centering
\setlength{\tabcolsep}{6pt}
\renewcommand{\arraystretch}{1.2}

\small
% \resizebox{1\textwidth}{!}{%
\begin{tabular}{lccccc}
\toprule
\textbf{Base Ability} 
& \textbf{EN-Math} 
& \textbf{AR-Math} 
& \textbf{EN-Science} 
& \textbf{AR-Science} 
& \textbf{Cross-Lingual} \\
\midrule
EN-Math
& 100.0\% (17{,}505)
& -- 
& -- 
& -- 
& 26.4\% (4{,}625) \\

AR-Math
& 25.3\% (4{,}434)
& 100.0\% (17{,}505)
& -- 
& -- 
& 35.7\% (6{,}253) \\

EN-Science
& 67.6\% (11{,}827)
& 31.3\% (5{,}476)
& 100.0\% (17{,}505)
& -- 
& 33.2\% (5{,}805) \\

AR-Science
& 21.0\% (3{,}669)
& 80.9\% (14{,}157)
& 26.5\% (4{,}635)
& 100.0\% (17{,}505)
& 30.5\% (5{,}346) \\
\bottomrule
\end{tabular}
% }

\caption{
Overlap of top-$1\%$ activation-difference channels across abilities for Qwen2.5-7B Math-Instruct vs.\ Instruct.
%Entries report the fraction (and count) of channels shared between ability-specific masks.
% normalized by the base (row) mask; ``--'' denotes symmetric entries.
The last column shows overlap with the union of both domains in the other language.
}

% \caption{
% Overlap of top-$0.5\%$ activation-difference channels across representative language-domain abilities for Qwen2.5-7B Math Instruct vs Instruct.
% Each entry reports the fraction (and absolute count) of channels shared between the ability-specific channel mask in the row and that in the column, normalized by the size of the base (row) mask.
% Diagonal entries correspond to self-overlap.
% Entries marked with ``--'' are omitted due to symmetry.
% The last column reports overlap with the union of both tasks in the corresponding the other language.
% }

\label{tab:mask_overlap}
\end{table*}

\subsection{Ability-specific Disentanglement}
\label{sec:disentanglement}
Based on the findings about the unbalanced activation difference distribution, we further study the ability-specific disentanglement for the top-ranked output channels.
Specifically, for each language-domain combination, we rank output channels by their average activation difference and select the top $1\%$ as the \emph{ability-specific channels}. Then, we empirically study:
%study channels with large cross-model activation differences from two complementary perspectives:
(i) whether channels associated with different language-domain abilities are easy to disentangle, and
(ii) whether channels associated with the same ability are consistent across different fine-tuned variations derived from the same pretrained model.
%Both aspects are critical for assessing the reliability of selective ability transfer.

% between masks.

\paragraph{Finding 3: Ability-specific channels are naturally disentangled.}
Table~\ref{tab:mask_overlap} reports the overlap of top activation difference channels for four abilities on the Qwen2.5-7B math model pair.
We observe low overlap between different language-domain abilities (e.g., English-Math vs.\ Arabic-Science), in the $20\%$--$30\%$ range.
Low overlap also persists across languages within the same domain, as well as when comparing a single ability to the union of both domains in another language.
%In contrast, overlaps between different domains within the same language are moderately higher (approximately $70\%$--$80\%$), indicating partial channel reuse across domains.
%Nevertheless, a substantial fraction of channels remains specific.
These results show the natural disentanglement of the language-specific channels with large activation differences.
% are primarily conditioned on language.

\begin{figure*}[t]
\centering
% ---------------- Left: Figure ----------------
\begin{minipage}[t]{0.66\textwidth}
    \vspace{0pt} 
    \centering
    \begin{subfigure}[t]{0.56\linewidth}
        \centering
        \includegraphics[width=\linewidth]{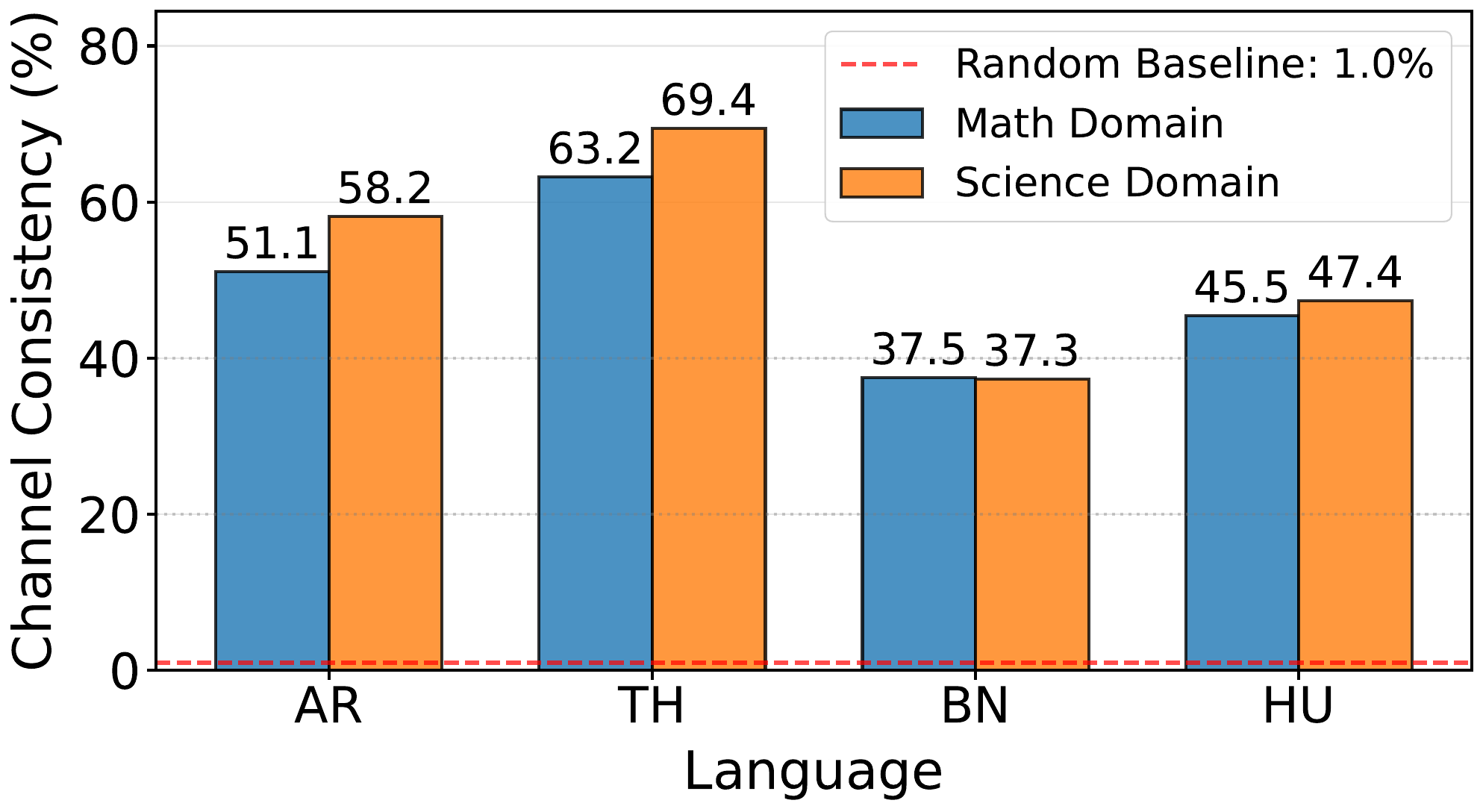}
        \label{fig:cross_finetune_consistency_filtered}
    \end{subfigure}
    \hfill
    \begin{subfigure}[t]{0.4\linewidth}
        \centering
        \includegraphics[width=\linewidth]{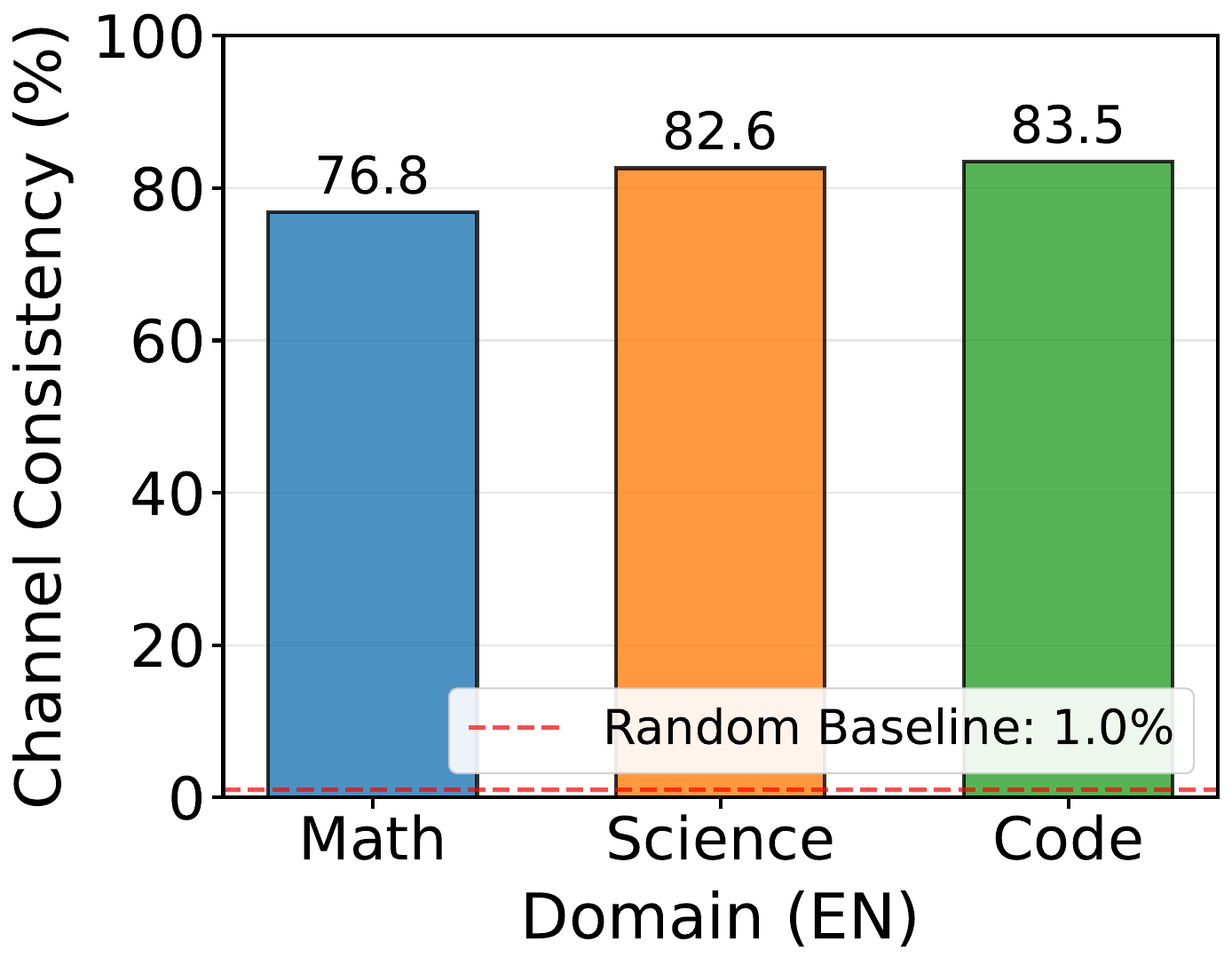}
        \label{fig:cross_model_consistency_llama2_tulu_wizardlm}
    \end{subfigure}
    \vspace{-20pt}
    \captionof{figure}{
    Cross-pair consistency of ability-specific channels.
    \textbf{Left:} Qwen2.5-7B overlap between top-1\% activation channels from (Math-Instruct vs.\ Instruct) and (Coder-Instruct vs.\ Instruct); % shown for representative languages; math and science are reported separately.
    \textbf{Right:} LLaMA-2-13B overlap from (Tulu-2-DPO vs.\ LLaMA-2) and (WizardLM vs.\ LLaMA-2).% evaluated on English math, science, and code.
    }
    \label{fig:cross_pair_consistency_two_examples}
\end{minipage}
\hfill
% ---------------- Right: Table ----------------
\begin{minipage}[t]{0.32\textwidth}
    \vspace{0pt} 
    \centering
    \small
    \setlength{\tabcolsep}{5pt}
    \renewcommand{\arraystretch}{1.15}

    \begin{tabular}{lcc}
    \toprule
    \textbf{Ability} & \textbf{Overlap (\%)} & \textbf{\#Ch.} \\
    \midrule
    EN-Math      & 4.66 & 816 \\
    EN-Science   & 5.35 & 936 \\
    AR-Math      & 5.62 & 983 \\
    AR-Science   & 3.83 & 670 \\
    \midrule
    Random       & 1.00 & -- \\
    \bottomrule
    \end{tabular}
    \vspace{2pt}
    \captionof{table}{
    Weight-activation top-1\% mask overlap (Qwen2.5-7B Math-Instruct vs.\ Instruct).
    }
    \label{tab:weight_activation_overlap}
\end{minipage}

\end{figure*}

\paragraph{Finding 4: Ability-specific channels exhibit consistency after fine-tuning.}
We compare top-1\% identified specific channels for the same ability across different fine-tuned LLM pairs derived from the same pretrained model.
As shown in Figure~\ref{fig:cross_pair_consistency_two_examples}, these channels exhibit significant overlap for the same language-domain ability, in both Qwen2.5-7B and LLaMA-2-13B models.
%, but with pronounced variation across languages.
It indicates that the ability-specific channels stably exist in the LLM, and fine-tuning will not greatly change that.
Thus, it is promising to transfer the specific channels across different fine-tuned variants from a pre-trained LLM, to recovery or enhance the ability.

\subsection{Comparison with Fine-tuning Updated Channels}
\label{sec:activation_vs_weight}

We examine if identified ability-specific channels are the same as largely updated ones in fine-tuning.
Using the same model pair as before (Qwen2.5-Math-7B-Instruct vs.\ Qwen2.5-7B-Instruct), we compute a channel-wise weight difference by calculating parameter differences with an $\ell_2$ norm.
Then, we compute the overlap between top $1\%$ channels selected by update and activation differences.
%The top $1\%$ channels form a \emph{weight-based mask}, which we compare against the top $1\%$ \emph{activation-based mask} using overlap.

% Channel-wise weight updates during fine-tuning also exhibit heavy-tailed distributions (Figure~\ref{fig:weight_ccdf_by_module}).

\paragraph{Finding 5: Ability-specific channels selected by activations differ from largely updated ones during fine-tuning.}
Figure~\ref{fig:weight_ccdf_by_structure} (Appendix~\ref{app:finding5}) shows that channel-wise weight updates also exhibit heavy-tailed distributions. 
% but with substantially weaker tails than activation differences. 
As shown in Table~\ref{tab:weight_activation_overlap}, the overlap between channels selected by activation differences and those with the largest parameter updates remains limited (3.8\%--5.6\% across language-domain abilities).
It indicates that channels with large activation differences do not simply correspond to those with the largest parameter updates.
This suggests that activation-based localization is able to identify the parameters that are highly important to the ability but may not be well trained during fine-tuning.
Therefore, it is promising to first use the activation-based localization and then move the identified parameters before fine-tuning, for efficient ability transfer or recovery.

% Additional results are provided in \textcolor{blue}{Appendix~X}.
%Although exceeding the random baseline ($1\%$), this overlap is far from complete, 

\begin{figure}[t]
  \centering
  \includegraphics[width=\linewidth]{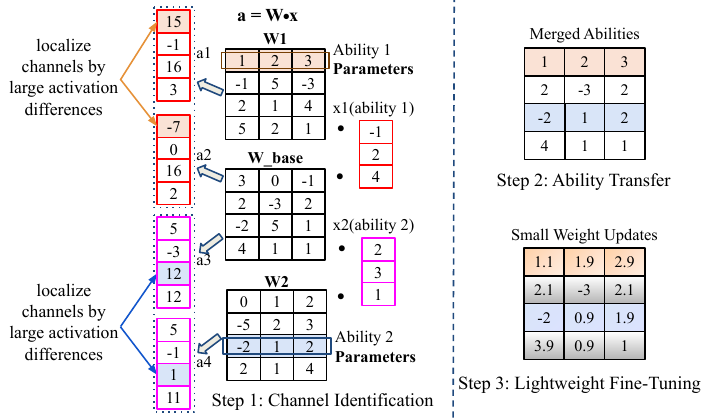}
  \caption{Illustration of ACT. 
    $W_{\text{base}}$ denotes a base model, while $W_1$ and $W_2$ denote two ability models.}
    %Step 1: For each ability, we feed ability-relevant inputs into the corresponding ability model and the shared base model, and identify ability-specific channels by large activation differences.
    %Step 2: Parameters from each ability model are selectively transferred to the base model, restricted to the identified channels, yielding a merged model with minimal ability conflicts.
    %Step 3: A brief fine-tuning is applied to stabilize the merged parameters and improve compatibility across abilities.}
  \label{fig:workflow}
\end{figure}

\section{Approach}
\label{sec:approach}
Motivated by our empirical findings, we propose \textbf{ACT} (\textbf{A}ctivation-Guided \textbf{C}hannel-wise Ability \textbf{T}ransfer), a selective ability transfer framework that transfers specific abilities across LLMs while minimizing interference with unrelated capabilities.
The core idea of ACT is to localize modularized parameters within the LLM corresponding to the target ability, to enable efficiently and effectively ability transfer or recovery.
%restrict parameter transfer to the channels most responsible for the observed cross-model behavioral differences.
As shown in Figure~\ref{fig:workflow}, ACT consists of three stages:
(1) identifying ability-specific channels via cross-model activation differences,
(2) selectively transferring parameters using masked task vectors, and
(3) applying lightweight post fine-tuning to stabilize the merged model.

% Given a target model and one or more source models that share the same architecture and tokenizer, ACT aims to selectively incorporate abilities from the source models into the target model without substantially degrading the target model’s existing strengths.

% We use the terms \emph{ability transfer--relevant channels} and \emph{ability-relevant channels} interchangeably for brevity.

\subsection{Ability-Transfer Channel Identification}
\label{sec:act_mask}

Let $M_{\text{trg}}$ denote a target model whose performance on a specific language-domain ability is to be improved, and $M_{\text{abl}}$ an ability model that exhibits strong performance on that ability.
Both models share the same architecture and tokenizer and are preferably derived from the same pretrained checkpoint. 
For a given language-domain combination, we feed the same set of input samples into both models and record the output activations of all trainable modules.
For each output channel $i$, we compute a token-averaged activation difference
\begin{equation}
\Delta a_i = \frac{1}{\left| T \right|}\sum_{t=1}^{T} \left| a_{i,t}^{(\text{abl})} - a_{i,t}^{(\text{trg})} \right|,
\end{equation}
where activations are averaged over answer tokens across all samples, as described in Section~\ref{sec:analysis_setup}.
This yields a channel-wise measure of how differently the two models respond internally to same inputs.

We then rank all output channels by $\Delta a_i$, irrespective of layer or module, and select the top $p\%$ to form an \emph{activation-based ability-specific mask}
\[
\mathcal{M}_{\text{act}} = \{ i \mid \Delta a_i \text{ ranks among the top } p\% \}.
\]
This mask defines a sparse set of channels that exhibit the larger cross-model behavioral discrepancies for the target ability.

% This mask defines a sparse set of channels most relevant to the transfer of target ability.

\subsection{Ability Transfer with Masks}
\label{sec:act_transfer}

Given a collection of activation-based channel masks
$\{\mathcal{M}_1, \dots, \mathcal{M}_K\}$,
each corresponding to a specific language--domain ability,
ACT performs selective parameter transfer from one or more \emph{ability models} to a target model.
Unlike global merging, parameter updates are restricted to channels identified as ability-relevant, reducing interference and improving parameter efficiency.

\paragraph{Model-level Mask Aggregation.}
In practice, multiple ability masks may originate from the same ability model.
For example, a single fine-tuned model may provide activation-based masks for several related abilities (e.g., multiple languages or domains).
To avoid repeatedly transferring the same task vector, ACT first aggregates masks at the \emph{model level}.
Specifically, for each ability model $m$, we define a unified mask
\begin{equation}
\mathcal{M}^{(m)} = \bigcup_{k \in \mathcal{K}_m} \mathcal{M}_k,
\end{equation}
where $\mathcal{K}_m$ indexes all ability masks of model $m$.
This unified mask identifies all channels to which knowledge from model $m$ will be transferred.

\paragraph{Masked Task Vector Transfer.}
Let $\theta_{\text{trg}}$ denote the parameters of the target model, and
$\theta_{\text{abl}}^{(m)}$ those of ability model $m$.
Following task arithmetic~\citep{ilharco2023editingmodelstaskarithmetic}, we compute a single task vector for each ability model:
\begin{equation}
\Delta \theta^{(m)} = \theta_{\text{abl}}^{(m)} - \theta_{\text{trg}}.
\end{equation}
ACT then applies the task vector only to parameter rows selected by the unified mask $\mathcal{M}^{(m)}$.
For each output channel $i$, the merged parameters are
\begin{equation}
\resizebox{.85\linewidth}{!}{$
\theta_i^{(\text{merged})}
=
\theta_i^{(\text{trg})}
+
\sum_{m}
\lambda_m \,
\Delta \theta_i^{(m)} \,
\mathbf{1}\!\left(i \in \mathcal{M}^{(m)}\right), 
$}
% \tag{\normalsize 5}
\end{equation}
where $\lambda_m$ are the scaling factors that control the transfer strength from ability model $m$, and
$\mathbf{1}(\cdot)$ denotes the indicator function.

By restricting parameter updates to the activation-based ability-specific channel masks,
ACT injects ability-relevant knowledge while preserving the majority of the target model’s parameters.
This design is directly motivated by our empirical findings that large activation differences are sparse and largely disentangled across abilities. 
% Moreover, aggregating masks at the model level explicitly reduces over-counting when multiple abilities share the same source model.
It is also worth noticing that this masked ability transfer is a versatile plug-and-play module and may be applied to any other model merging methods, such as Fisher \citep{matena2022mergingmodelsfisherweightedaveraging}, TIES \citep{yadav2023tiesmergingresolvinginterferencemerging}, and DARE \citep{yu2024languagemodelssupermario}.

\subsection{Lightweight Post-Transfer Fine-Tuning}
\label{sec:act_finetune}

After masked parameter transfer, ACT applies a lightweight post-transfer fine-tuning step to improve stability and compatibility.
Specifically, we perform a brief supervised fine-tuning on a small set of ability-relevant samples using a low learning rate and a limited number of optimization steps, i.e., 1500 instance per ability.
Although all model parameters are updated during this stage, the objective is not to relearn the target abilities from scratch, but to smooth the integration of the transferred parameters within the existing network.
As demonstrated in our experiments, this post-transfer refinement is substantially more effective than fine-tuning alone while largely preserving the target model’s original capabilities.

\begin{table*}
\centering
\setlength{\extrarowheight}{0pt}
\addtolength{\extrarowheight}{\aboverulesep}
\addtolength{\extrarowheight}{\belowrulesep}
\setlength{\aboverulesep}{0pt}
\setlength{\belowrulesep}{0pt}
\resizebox{1\textwidth}{!}{%
\begin{tabular}{lccccccccccccc} 
\toprule
\multirow{2}{*}{\textbf{Methods}} & \multicolumn{3}{c}{\textbf{Math (Keep)}}                        & \multicolumn{10}{c}{\textbf{Science (Multilingual Recovery)}}                                                     \\ 
\noalign{\kern-\cmidrulewidth}\cmidrule(l){2-14}
                                  & EN   & ZH   & {\cellcolor[rgb]{0.898,0.898,0.898}}Avg. Keep     & AR   & BN   & HU   & JA   & KO   & SW   & TE   & TH   & VI   & {\cellcolor[rgb]{0.898,0.898,0.898}}Avg. Recover   \\ 
\midrule
\multicolumn{14}{c}{\textit{Multi-lingual Large Language Models}}                                                                                                                                                       \\
\textbf{Qwen2.5-7B Instruct}      & 82.0 & 74.0 & {\cellcolor[rgb]{0.898,0.898,0.898}}78.0          & 26.6 & 20.5 & 26.1 & 27.5 & 26.8 & 21.4 & 21.4 & 28.4 & 30.6 & {\cellcolor[rgb]{0.898,0.898,0.898}}\textbf{25.5}  \\
\textbf{Qwen2.5-7B Math Instruct} & 92.4 & 84.0 & {\cellcolor[rgb]{0.898,0.898,0.898}}\textbf{88.2} & 14.3 & 9.6  & 13.0 & 11.6 & 11.8 & 15.4 & 11.8 & 12.3 & 11.2 & {\cellcolor[rgb]{0.898,0.898,0.898}}12.3           \\ 
\hdashline
\multicolumn{14}{c}{\textit{Ability Transfer Approaches}}                                                                                                                                                               \\
+ Task Arithmetic                 & 89.6 & 85.6 & {\cellcolor[rgb]{0.898,0.898,0.898}}87.6          & 16.5 & 4.2  & 17.0 & 16.3 & 13.2 & 13.8 & 12.5 & 13.2 & 13.6 & {\cellcolor[rgb]{0.898,0.898,0.898}}13.4           \\
+ TIES Merging                    & 92.4 & 84.4 & {\cellcolor[rgb]{0.898,0.898,0.898}}88.4          & 11.8 & 8.0  & 14.5 & 14.1 & 12.5 & 16.3 & 11.2 & 15.6 & 12.7 & {\cellcolor[rgb]{0.898,0.898,0.898}}13.0           \\
+ DARE Merging                    & 91.2 & 86.0 & {\cellcolor[rgb]{0.898,0.898,0.898}}\textbf{88.6} & 15.2 & 7.1  & 13.2 & 13.8 & 13.8 & 13.8 & 13.4 & 13.0 & 15.2 & {\cellcolor[rgb]{0.898,0.898,0.898}}13.2           \\
+ \textbf{Ours (transfer-only)}   & 90.0 & 84.8 & {\cellcolor[rgb]{0.898,0.898,0.898}}87.4          & 15.4 & 13.6 & 13.6 & 17.6 & 19.0 & 17.6 & 17.4 & 15.0 & 15.0 & {\cellcolor[rgb]{0.898,0.898,0.898}}\textbf{16.0}  \\ 
\hdashline
\multicolumn{14}{c}{\textit{With supervised fine-tuning}}                                                                                                                                                               \\
+ SFT                             & 87.6 & 74.8 & {\cellcolor[rgb]{0.898,0.898,0.898}}81.2          & 18.3 & 12.1 & 19.0 & 21.2 & 19.4 & 8.0  & 11.4 & 21.9 & 19.2 & {\cellcolor[rgb]{0.898,0.898,0.898}}16.7           \\
+ TA $\&$ SFT                     & 88.4 & 79.6 & {\cellcolor[rgb]{0.898,0.898,0.898}}84.0          & 21.2 & 15.4 & 18.3 & 17.4 & 20.8 & 10.0 & 13.4 & 19.4 & 17.9 & {\cellcolor[rgb]{0.898,0.898,0.898}}17.1           \\
+ TIES $\&$ SFT                   & 88.4 & 75.6 & {\cellcolor[rgb]{0.898,0.898,0.898}}82.0          & 20.1 & 11.2 & 17.4 & 16.1 & 17.9 & 10.5 & 13.0 & 18.5 & 16.5 & {\cellcolor[rgb]{0.898,0.898,0.898}}15.7           \\
+ DARE $\&$ SFT                   & 88.8 & 74.8 & {\cellcolor[rgb]{0.898,0.898,0.898}}81.8          & 20.8 & 13.0 & 18.1 & 19.2 & 20.8 & 10.7 & 13.4 & 19.4 & 21.6 & {\cellcolor[rgb]{0.898,0.898,0.898}}17.4           \\
+ \textbf{Ours w/ SFT}            & 90.4 & 79.6 & {\cellcolor[rgb]{0.898,0.898,0.898}}\textbf{85.0}          & 22.1 & 18.1 & 20.3 & 19.0 & 22.3 & 11.6 & 21.4 & 19.9 & 18.3 & {\cellcolor[rgb]{0.898,0.898,0.898}}\textbf{19.2}           \\
\bottomrule
\end{tabular}
}
\caption{Math preservation (EN/ZH) and multilingual science recovery. Math (Keep) reports English and Chinese math accuracy, and Science (Multilingual Recovery) reports science accuracy on the remaining 9 languages.}
\label{tab:math_keep_science_recover}
\end{table*}

\section{Experiments}
\label{sec:experiments}

\subsection{Recovering Forgotten Abilities}
\label{sec:exp_recover}

%We begin by evaluating ACT in a challenging yet practically relevant setting: recovering abilities that are degraded in a specialized model while preserving its strong performance on the retained skills.

\subsubsection{Experimental Setup}

\paragraph{Datasets and Models.}
Following the empirical analysis in Section~\ref{sec:analysis_setup}, we consider two reasoning domains, mathematics and science, across 11 languages.
Mathematics data are constructed from multilingual translations of MetaMathQA~\citep{yu2024metamathbootstrapmathematicalquestions}, while science data are derived from multilingual translations of MegaScience~\citep{fan2025megasciencepushingfrontiersposttraining}, covering physics, chemistry, and biology.
The same datasets are used for both ability localization and post-transfer fine-tuning.
Our experiments focus on the model pair Qwen2.5-Math-7B-Instruct and Qwen2.5-7B-Instruct, which share the same Qwen2.5-7B pretrained base and are obtained through continued pre-training processes.
The goal is to preserve the strong English and Chinese mathematical reasoning performance of Qwen2.5-Math-7B-Instruct, while recovering general scientific reasoning performance in the remaining languages.

\paragraph{Evaluation Metrics and Baselines.}
We evaluate all models using the multilingual and multi-ability evaluation framework BenchMax~\citep{huang2025benchmaxcomprehensivemultilingualevaluation}, where the mathematics and science evaluations are drawn from MGSM~\citep{shi2022languagemodelsmultilingualchainofthought} and GPQA~\citep{rein2023gpqagraduatelevelgoogleproofqa}, respectively. 
All evaluations are conducted under a zero-shot setting, and performance is measured using exact-match accuracy.
Inference is performed with \texttt{vLLM}~\citep{kwon2023efficientmemorymanagementlarge}.
We compare our method against three representative parameter-transfer baselines:
Task Arithmetic~\citep{ilharco2023editingmodelstaskarithmetic},
TIES~\citep{yadav2023tiesmergingresolvinginterferencemerging},
and DARE~\citep{yu2024languagemodelssupermario}.
For a fair comparison, we report results under two settings:
(i) parameter transfer without post-transfer fine-tuning, and
(ii) parameter transfer followed by lightweight supervised fine-tuning.
In the latter case, the same fine-tuning protocol is applied to all baselines. Detailed implementations and hyperparameters are in Appendix~\ref{app:implementation_recovery}.

\subsubsection{Main Results}

Table~\ref{tab:math_keep_science_recover} summarizes the performance of our method and baseline approaches.
Our evaluation targets a challenging setting: preserving the strong English and Chinese mathematical reasoning of Qwen2.5-7B Math-Instruct while recovering scientific reasoning in the remaining languages.

\paragraph{Transfer Only.}
Under the transfer-only setting, our method achieves the highest average science accuracy ($16.0$) among all approaches by transferring only 4.73\% of the model parameters, while preserving English and Chinese mathematical performance at the level of the Math-Instruct model.
In contrast, baseline merging methods yield only marginal improvements in science performance when math accuracy is maintained.
The limited effectiveness of these baseline methods is likely due to the fact that Qwen2.5-7B Math-Instruct is obtained via continuous pre-training on large-scale mathematical data~\citep{yang2024qwen25mathtechnicalreportmathematical}, which induces substantial parameter drift relative to the Instruct model.
Such large parameter shifts, together with configuration mismatches between the two models, violate the assumptions underlying most existing merging methods, which are effective only when parameter updates remain relatively small~\citep{yu2024languagemodelssupermario}.

\paragraph{Transfer with Lightweight Fine-Tuning.}
With lightweight post-transfer fine-tuning using 1{,}500 samples per language-domain combination, science performance further improves to $19.22$, while English and Chinese mathematical accuracy remains close to that of the Math-Instruct model.
This indicates that lightweight fine-tuning effectively smooths the integration of transferred parameters and alleviates residual interference.
Notably, the resulting model outperforms direct fine-tuning of Qwen2.5-7B Math-Instruct alone on both mathematics (85.0 vs.\ 81.2) and science (19.22 vs.\ 16.7).
In contrast, baseline methods derive substantially less benefit from fine-tuning: although their science performance increases, it remains comparable to fine-tuning without prior parameter transfer, suggesting limited synergy between merging and fine-tuning in these approaches.

\begin{figure}[t]
    \centering
    % --- Row 1: Transfer Ratio Comparison ---
    \begin{subfigure}[b]{0.48\linewidth}
        \centering
        \includegraphics[width=\linewidth]{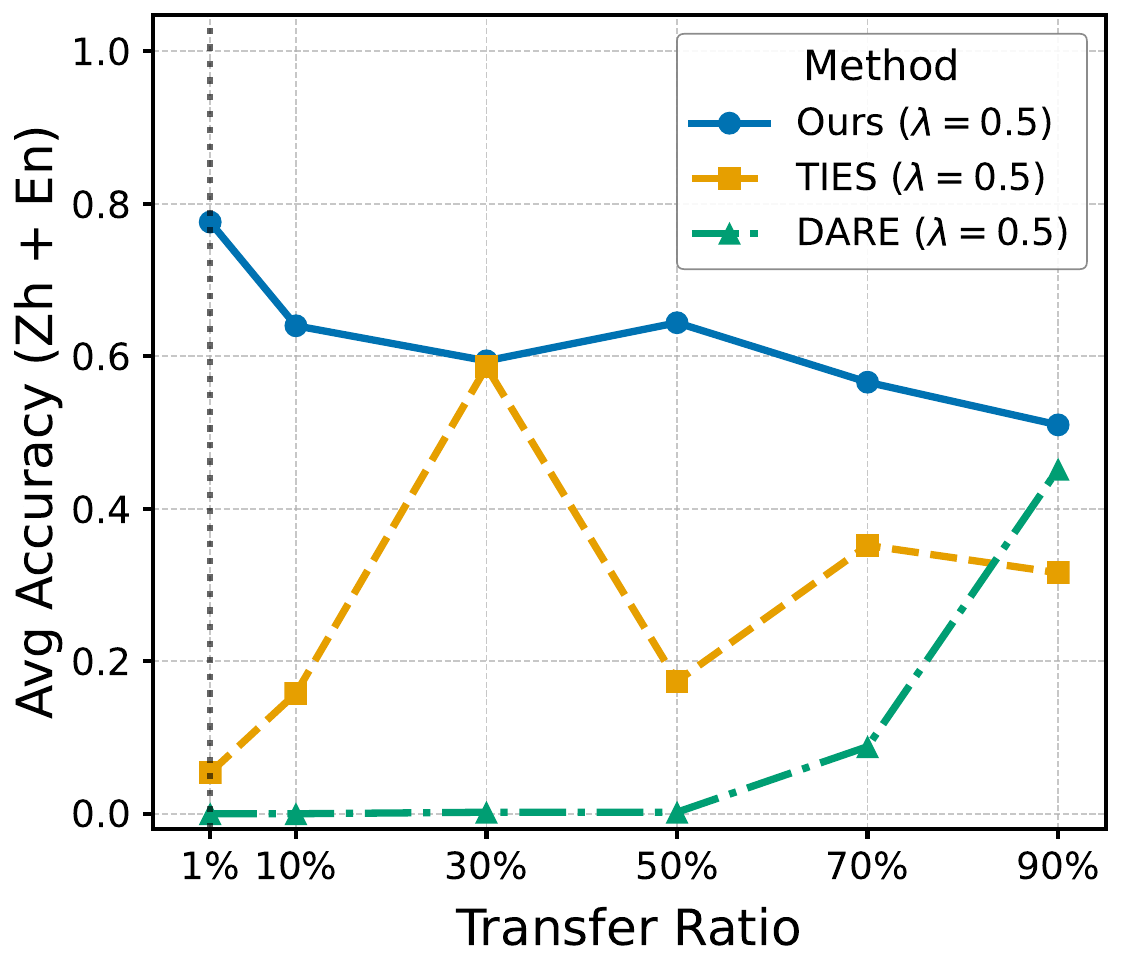}
        \caption{Math (transfer ratio)}
        \label{fig:math_ratio}
    \end{subfigure}
    \hfill
    \begin{subfigure}[b]{0.48\linewidth}
        \centering
        \includegraphics[width=\linewidth]{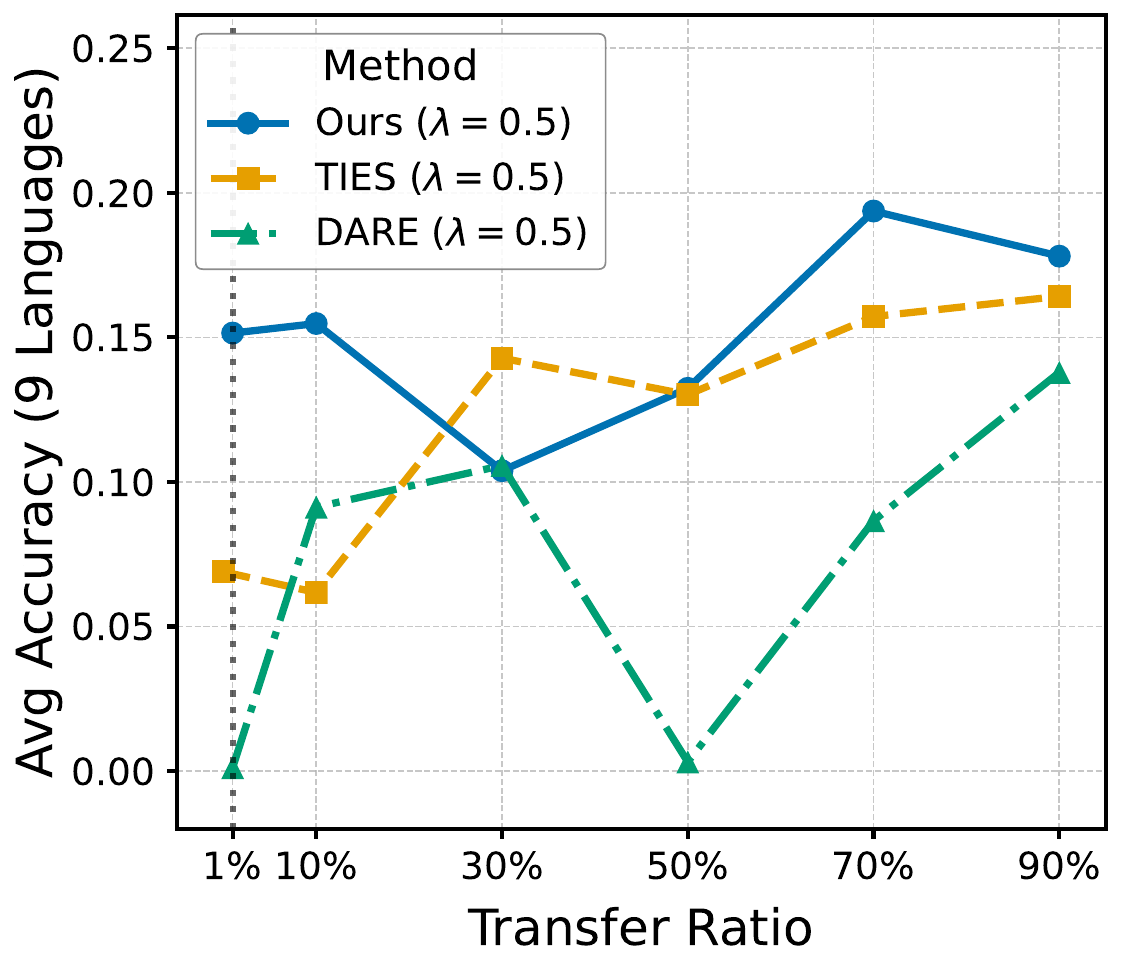}
        \caption{Science (transfer ratio)}
        \label{fig:science_ratio}
    \end{subfigure}
    
    \vspace{0.5cm} % Add vertical space between rows
    
    % --- Row 2: Scaling Factor Comparison ---
    \begin{subfigure}[b]{0.48\linewidth}
        \centering
        \includegraphics[width=\linewidth]{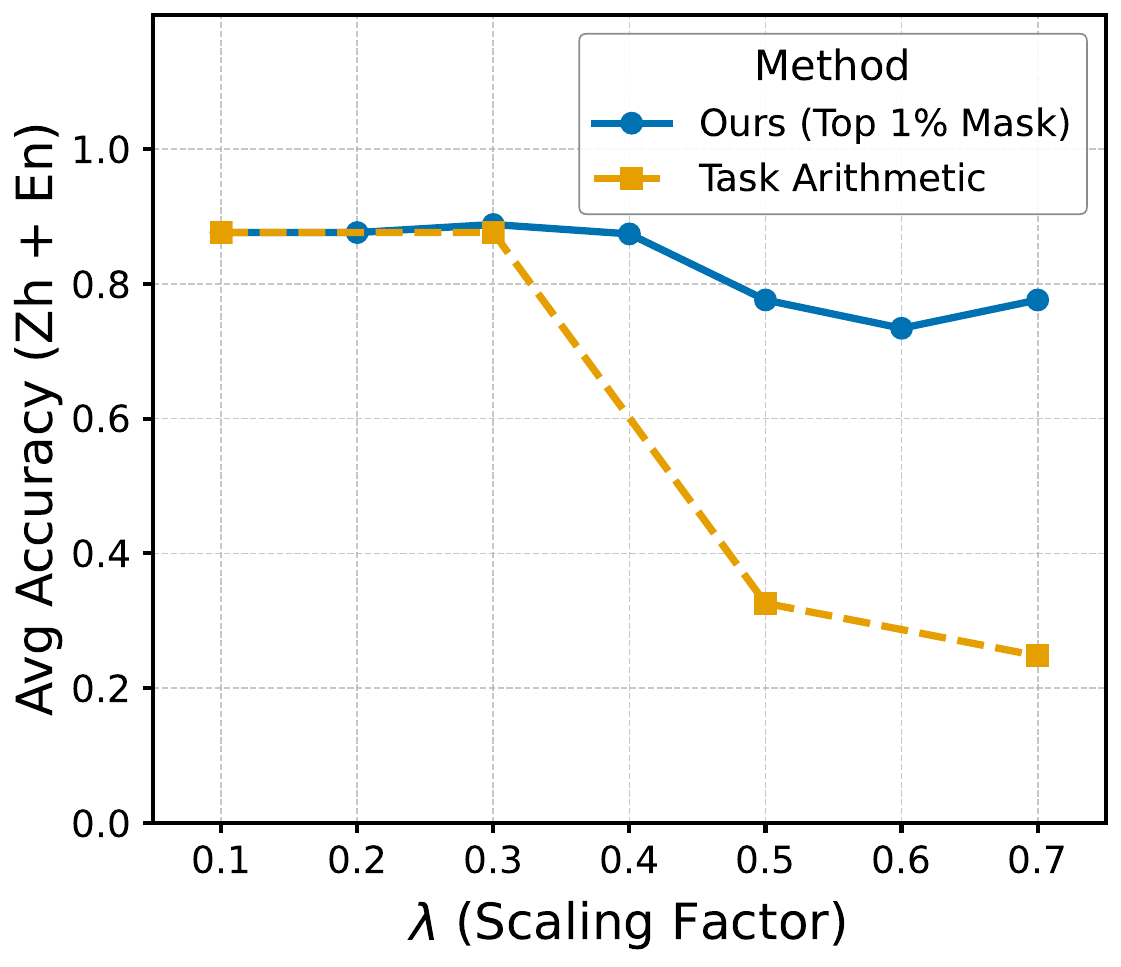}
        \caption{Math ($\lambda$)}
        \label{fig:math_lambda}
    \end{subfigure}
    \hfill
    \begin{subfigure}[b]{0.48\linewidth}
        \centering
        \includegraphics[width=\linewidth]{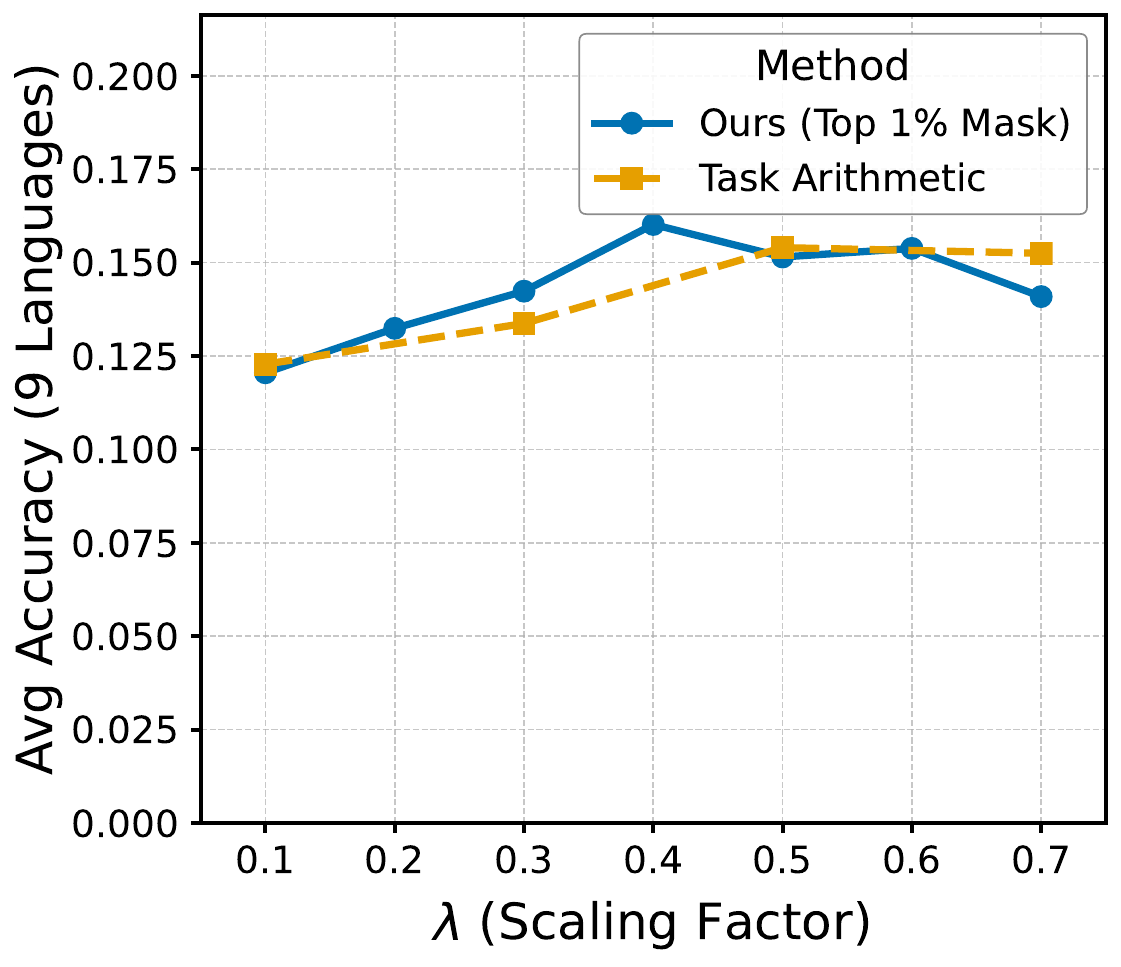}
        \caption{Science ($\lambda$)}
        \label{fig:science_lambda}
    \end{subfigure}
    
    \caption{Performance comparison of transferring from Qwen-2.5-7B-Instruct to Qwen-2.5-Math-7B-Instruct. Top row: Comparison across different transfer ratios with $\lambda=0.5$. Bottom row: Comparison using 1\% transfer ratio with Task Arithmetic across different $\lambda$.}
    \label{fig:overall_comparison}
\end{figure}

\subsubsection{Sensitivity Analysis}

We conduct a sensitivity analysis on transfer-only setting to evaluate robustness under varying per-ability channel transfer ratios $p\%$ and scaling factors $\lambda$.
Figures~\ref{fig:math_ratio} and~\ref{fig:science_ratio} compare ACT with sparse merging baselines under a fixed $\lambda = 0.5$.
ACT preserves averaged math performance across all transfer ratios, while achieving stronger averaged science recovery in both highly sparse ($\leq 10\%$ transferred parameters) and dense regimes ($\geq 50\%$ transferred parameters).
At extreme sparsity (1\% transfer), DARE exhibits severe performance degradation due to its rescaling factor $1/(1-p)$, which amplifies transferred weights by $100\times$ and destabilizes the model.
By contrast, ACT remains stable under this setting.
Figures~\ref{fig:math_lambda} and~\ref{fig:science_lambda} further compare ACT with Task Arithmetic across different $\lambda$ values using a fixed 1\% transfer ratio.
While both methods maintain math accuracy for small $\lambda$, ACT consistently achieves higher science performance and remains stable as $\lambda$ increases.
Task Arithmetic, in contrast, fails to recover the Instruct model’s performance for relatively large $\lambda$, likely due to configuration and parameter mismatches between the Math-Instruct and Instruct models.
Overall, ACT demonstrates greater robustness to hyperparameter variations than baselines.

% \subsubsection{Sensitivity Analysis}
% We evaluate our performance against baselines under different hyperparameter settings on Qwen-2.5-7B-based models in Figure~\ref{fig:overall_comparison}. 
% Figures~\ref{fig:math_ratio} and \ref{fig:science_ratio} compare our method against sparse merging baselines with fixed $\lambda=0.5$. 
% Our method retains the highest math capability across all ratios and achieves superior science ability in sparse settings ($\le 10\%$ transfer ratio). 
% Notably, at extreme sparsity setting (1\% transfer ratio, i.e. sparsity ratio $p=0.99$), DARE's rescaling factor ($\frac{1}{1-p}$) amplifies weights by $100\times$, thus significantly degrading the base model's language capabilities. Figures~\ref{fig:math_lambda} and \ref{fig:science_lambda} contrast our approach with classic Task Arithmetic across varying $\lambda$. 
% While both methods preserve math proficiency for $0.1 \le \lambda \le 0.3$, our model yields comparable or even superior science performance using only \textbf{1\%} of the weights. 
% % This efficiency supports the Lottery Ticket Hypothesis~\citep{frankle2019lotterytickethypothesisfinding}, suggesting that new capabilities rely on sparse, subnet-level updates.  
% Furthermore, our method demonstrates greater stability as $\lambda$ increases; in contrast, TA fails to recover the original Qwen-2.5-7B-Instruct performance as $\lambda \to 1$ due to its configuration discrepancies with Qwen math model family.  Overall, our approach exhibits superior robustness to hyperparameter variations compared to all baselines.

\subsection{Multi-Abilities Merging Test}
\label{sec:exp_llama2}

We next evaluate ACT on merging multiple abilities from different fine-tuned models.% within the LLaMA-2-13B family.

\paragraph{Models and Abilities.}
We consider three widely used LLaMA-2-13B-based fine-tuned models, each specializing in a distinct English-language reasoning ability:
WizardMath-13B for mathematical reasoning,
Tulu-2-DPO-13B for scientific reasoning,
and WizardLM-13B-V1.2 for code generation.
Each model performs well on its target ability while being less competitive on the others.
Our goal is to integrate these abilities into a single model without degrading any individual strength. Detailed implementations are in Appendix~\ref{app:implementation_llama2}.

\paragraph{Results.}
Table~\ref{tab:llama2_ability_transfer} reports English math, science, and code performance for individual ability models and merged models.
ACT achieves the best overall performance, with an average score of $41.8$, substantially exceeding any single specialized model.
This demonstrates that our method effectively integrates multiple abilities when ability masks originate from different source models.
Notably, all baseline merging methods also achieve competitive results, only slightly below our method.
This contrasts with the Qwen2.5 results and may be attributed to the fact that the specialized LLaMA-2 models are obtained via standard fine-tuning with relatively limited data, rather than continuous pretraining.
As a result, parameter and activation differences between models are substantially smaller.
As shown in Appendix~\ref{app:finding1}, activation differences for LLaMA-2 model pairs are generally around $10^{-1}$ (with maxima around $10^{1}$), compared to Qwen2.5 model pairs where differences are typically around $1$ and can reach up to $10^{3}$.
Although heavy-tailed activation differences still exist, the reduced magnitude leads to weaker interference and makes merging more effective.
This experiment highlights that ACT is advantageous in regimes with large representation shifts, while remaining competitive in fine-tuning-level merging settings.

\begin{table}[t]
\centering
\resizebox{1\linewidth}{!}{%
\begin{tabular}{lcccc} 
\toprule
\textbf{Methods}             & \textbf{Math} & \textbf{Science} & \textbf{Code} & \textbf{Avg.}  \\ 
\midrule
\multicolumn{5}{c}{\textit{LLaMA 2 based models (English)}}                                      \\
\textbf{WizardMath-13B}      & \textbf{63.6} & 24.1             & 12.2          & 33.3           \\
\textbf{Tulu-2-DPO-13B}      & 34.0          & \textbf{28.1}    & 32.3          & 31.5           \\
\textbf{WizardLM-13B-V1.2}   & 55.2          & 23.4             & \textbf{36.6} & 38.4           \\ 
\hdashline
\multicolumn{5}{c}{\textit{Ability Transfer Approaches}}                                         \\
+~Task Arithmetic            & 68.4          & 24.6             & 31.7          & 41.6           \\
+ TIES Merging               & 66.4          & 25.0             & 31.7          & 41.0           \\
+ DARE Merging               & 68.4          & 25.4             & 31.1          & 41.6           \\
+ \textbf{Ours (transfer-only)} & 66.8          & 24.3             & 34.2          & \textbf{41.8}  \\
\bottomrule
\end{tabular}
}
\caption{
Performance of merging WizardMath (English math), Tulu-2 (science), and WizardLM (code).
}
\label{tab:llama2_ability_transfer}
\vspace{-15pt}
\end{table}

\section{Conclusion}
In this paper, we investigated how different abilities are stored in LLMs and how they can be efficiently recovered or transferred after specialization. Activation analysis across layers and modules shows that ability-related signals are concentrated in a small fraction of channels and remain disentangled, sufficient, and stable under fine-tuning, suggesting modularized ability-specific parameters.
Building on these findings, we proposed ACT, which localizes ability-relevant channels via cross-model activation differences and selectively transfers only the corresponding parameters through masked task-vector merging, with lightweight post-transfer fine-tuning for compatibility. Experiments demonstrate that ACT recovers forgotten multilingual reasoning abilities while preserving retained skills, and enables multi-ability integration with minimal interference while modifying only a small fraction of parameters per language–domain setting.

%Looking forward, ACT opens several promising directions: extending activation-guided localization to broader ability types beyond reasoning, and exploring compositional transfer for many abilities at larger scales. More broadly, our findings highlight channel-wise modularity as a practical handle for controllable continual adaptation, offering a step toward LLM specialization that is both efficient and less prone to forgetting.

\section{Limitations}

While our results demonstrate the effectiveness of activation-guided ability transfer, several limitations remain.
First, our experiments primarily consider model pairs or small groups of models that share the same pretrained backbone.
Extending the analysis to models with substantially different architectures or pretraining corpora is an interesting direction for future work.
Second, ability-specific channel localization is based on curated datasets for each ability.
As a result, the quality of the identified channel masks may depend on how well these datasets capture the underlying ability.
Finally, our study focuses on reasoning-oriented domains such as mathematics, science, and code.
It remains to be explored whether similar activation-based structures arise for other categories of abilities.

\bibliography{custom}

\clearpage
% \newpage

\appendix

\section{Implementation Details for Recovering Forgetten Abilities}
\label{app:implementation_recovery}
Following Section~\ref{sec:analysis_setup}, we use 1{,}500 samples for each language--domain combination.
For ability localization, we construct activation-based channel masks for scientific reasoning in the remaining 9 languages except English and Chinese, and take their union to guide parameter transfer.
Post-transfer fine-tuning is then performed on the full multilingual dataset, including both mathematics and science data across all 11 languages, to ensure balanced adaptation. 
We employ grid search to identify optimal hyperparameters. The merging scaling factor is evaluated over $\lambda \in \{0.1, 0.2, \dots, 0.9\}$. For the transfer ratio, we apply a fine-grained search with a 1\% step size for ratios $\le 10\%$ and a 5\% step size for ratios $> 10\%$. For our results reported in Table~\ref{tab:math_keep_science_recover}, we use top-$1\%$ activation-based channel masks per ability in the transfer-only setting; after taking the union across all selected abilities, this corresponds to transferring $4.73\%$ of the model parameters, with a scaling factor $\lambda = 0.4$.
When post-transfer fine-tuning is applied, we instead use top-$9\%$ masks per ability, whose union covers $11.35\%$ of the model parameters, with a scaling factor $\lambda = 0.7$.
Fine-tuning is conducted using the Adam optimizer~\citep{kingma2017adammethodstochasticoptimization} with a learning rate of $2\times10^{-5}$ for one epoch, and gradient updates every 16 samples. All fine-tuning is done on NVIDIA-H100 and NVIDIA-A100. Each training process takes approximately 2 hours on both devices.  We use NVIDIA-H100, NVIDIA-A100 and NVIDIA-A6000 for all evaluation. 

\section{Implementation Details for Multi-abilities Merging}
\label{app:implementation_llama2}
We use LLaMA-2-13B as the target model and transfer abilities from the three specialized models.
Ability-specific channel masks are extracted using 1{,}500 English samples for each domain:
math (MetaMathQA~\citealp{yu2024metamathbootstrapmathematicalquestions}),
science (MegaScience~\citealp{fan2025megasciencepushingfrontiersposttraining}),
and code (OpenCodeInstruct~\citealp{ahmad2025opencodeinstructlargescaleinstructiontuning}).
In this setting, we apply transfer-only ability transfer without post-transfer fine-tuning, as the merged models already achieve strong performance. 
For our results reported in Table~\ref{tab:llama2_ability_transfer}, we use top-90\% masks per ability, with the same scaling factor $\lambda = 0.6$ for all models.
We compare ACT against same representative model merging baselines used previously.

\section{AI Assistance Disclosure}
We used large language models to assist with manuscript editing and code refinement. Specifically, GPT-based models were used to polish the writing, and Claude Sonnet 4.5 was used to help revise and debug code. All scientific ideas, experimental design, results, and interpretations are solely those of the authors.

\section{Additional Empirical Analysis}
\label{app:additional_empirical_analysis}

\subsection{Additional Results for Finding~1}
\label{app:finding1}

In the main text (Finding~1), we show that cross-model activation differences are sparse and heavy-tailed: most output channels exhibit small deviations, while a small fraction accounts for the largest discrepancies.
Here, we provide additional evidence that this pattern is robust across model families (Qwen2.5 and LLaMA-2), model scales (1.5B/7B/13B), and ability domains (math/science/code in 11 languages).

\begin{figure}[htbp]
    \centering
    \includegraphics[width=\linewidth]{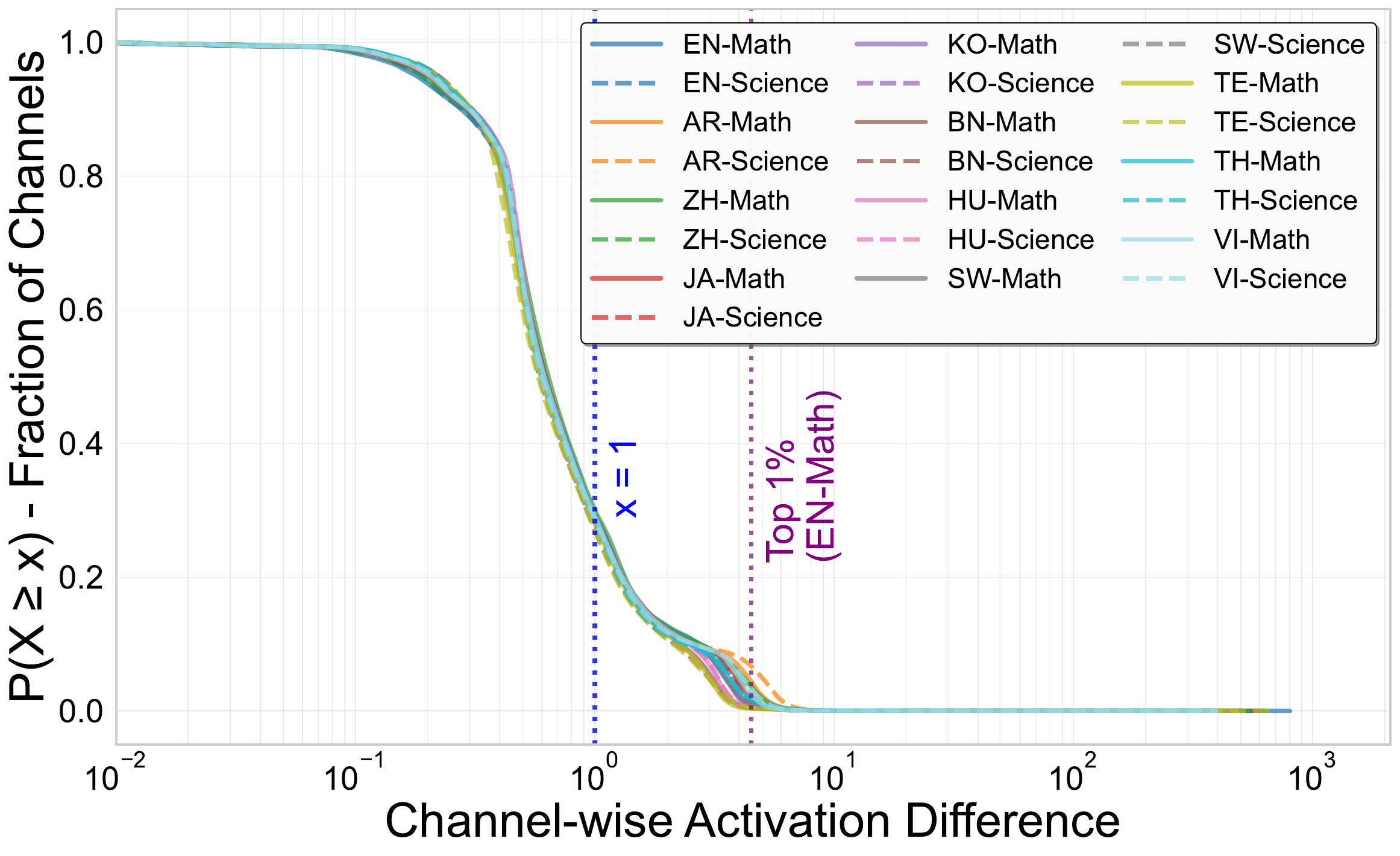}
    \caption{
    \textbf{Qwen2.5-Math-7B-Instruct vs.\ Qwen2.5-7B-Instruct.}
    CCDFs of channel-wise activation differences across 11 languages for math (solid) and science (dashed).
    Large activation differences concentrate in a small fraction of channels across all language-domain combinations.
    }
    \label{fig:app_ccdf_qwen_math_7b_all_lang}
\end{figure}

\begin{figure}[htbp]
    \centering
    \includegraphics[width=\linewidth]{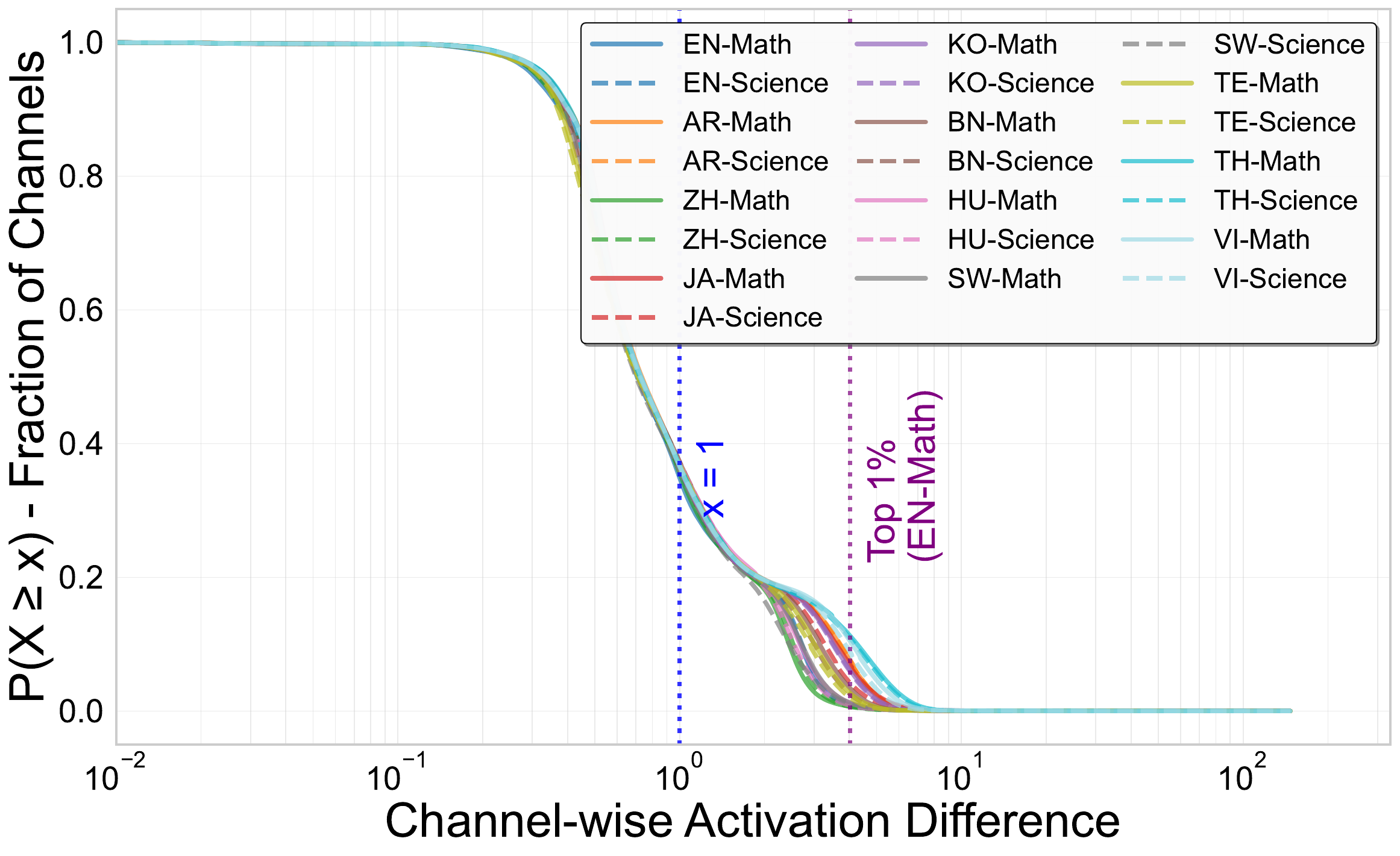}
    \caption{
    \textbf{Qwen2.5-Math-1.5B-Instruct vs.\ Qwen2.5-1.5B-Instruct.}
    CCDFs of channel-wise activation differences across 11 languages for math (solid) and science (dashed).
    The same sparse, heavy-tailed pattern persists at a smaller scale.
    }
    \label{fig:app_ccdf_qwen_math_15b_all_lang}
\end{figure}

\paragraph{Qwen2.5 Math Models.}
Figures~\ref{fig:app_ccdf_qwen_math_7b_all_lang} and~\ref{fig:app_ccdf_qwen_math_15b_all_lang} report CCDFs of channel-wise activation differences between Qwen2.5-Math-Instruct and Qwen2.5-Instruct at 7B and 1.5B, respectively.
We evaluate mathematical (solid) and scientific (dashed) reasoning across 11 languages.
Across all language-domain combinations, the CCDFs are highly consistent and heavy-tailed, indicating that large cross-model discrepancies concentrate in a small subset of channels rather than being uniformly distributed.

% --------------------------
% Qwen2.5 Math (7B / 1.5B)
% --------------------------

\begin{figure}[htbp]
    \centering
    \includegraphics[width=\linewidth]{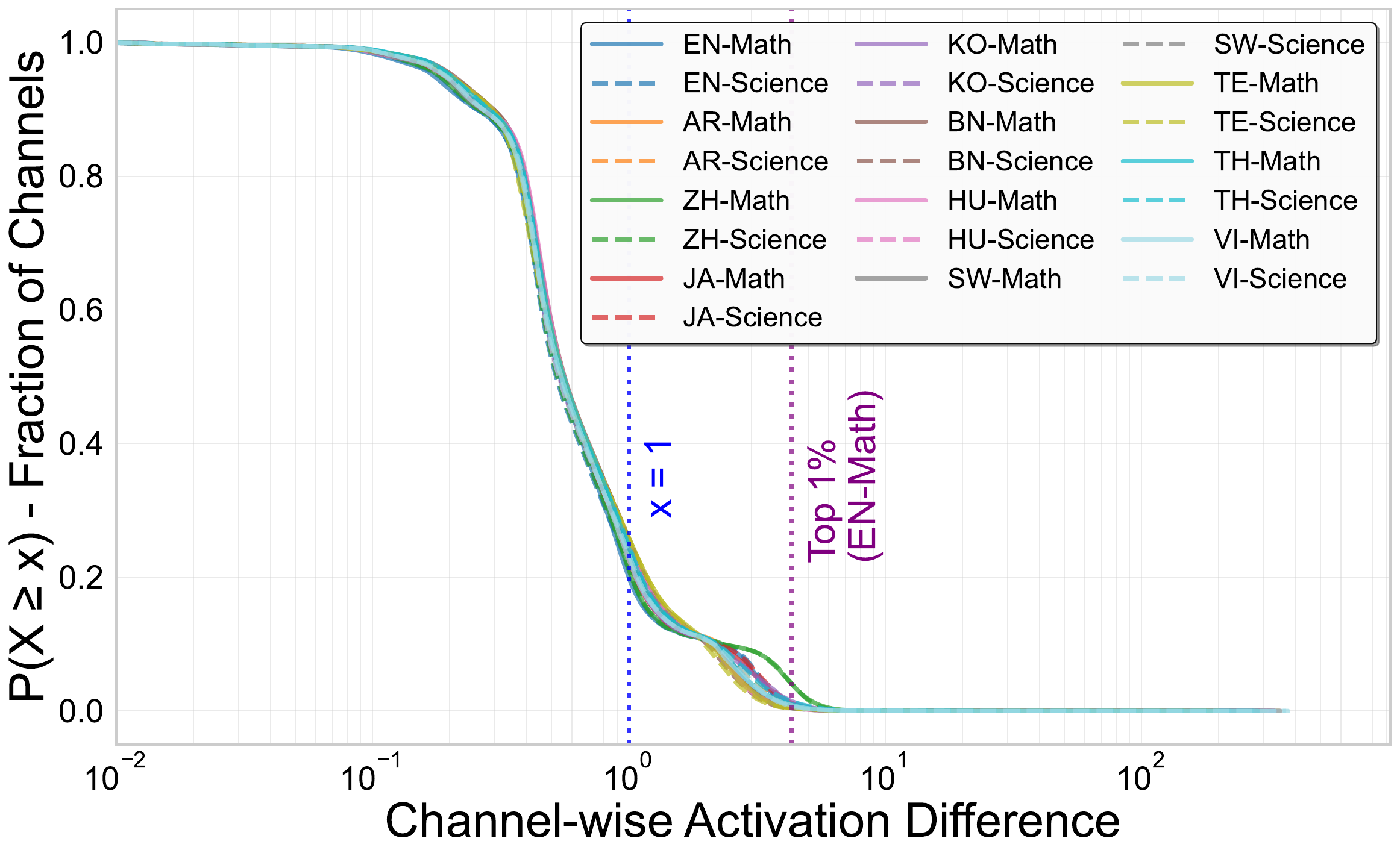}
    \caption{
    \textbf{Qwen2.5-Coder-7B-Instruct vs.\ Qwen2.5-7B-Instruct.}
    CCDFs of channel-wise activation differences across 11 languages for math (solid) and science (dashed).
    Despite coder specialization, activation differences remain sparse and heavy-tailed.
    }
    \label{fig:app_ccdf_qwen_coder_7b_all_lang}
\end{figure}

\begin{figure}[htbp]
    \centering
    \includegraphics[width=\linewidth]{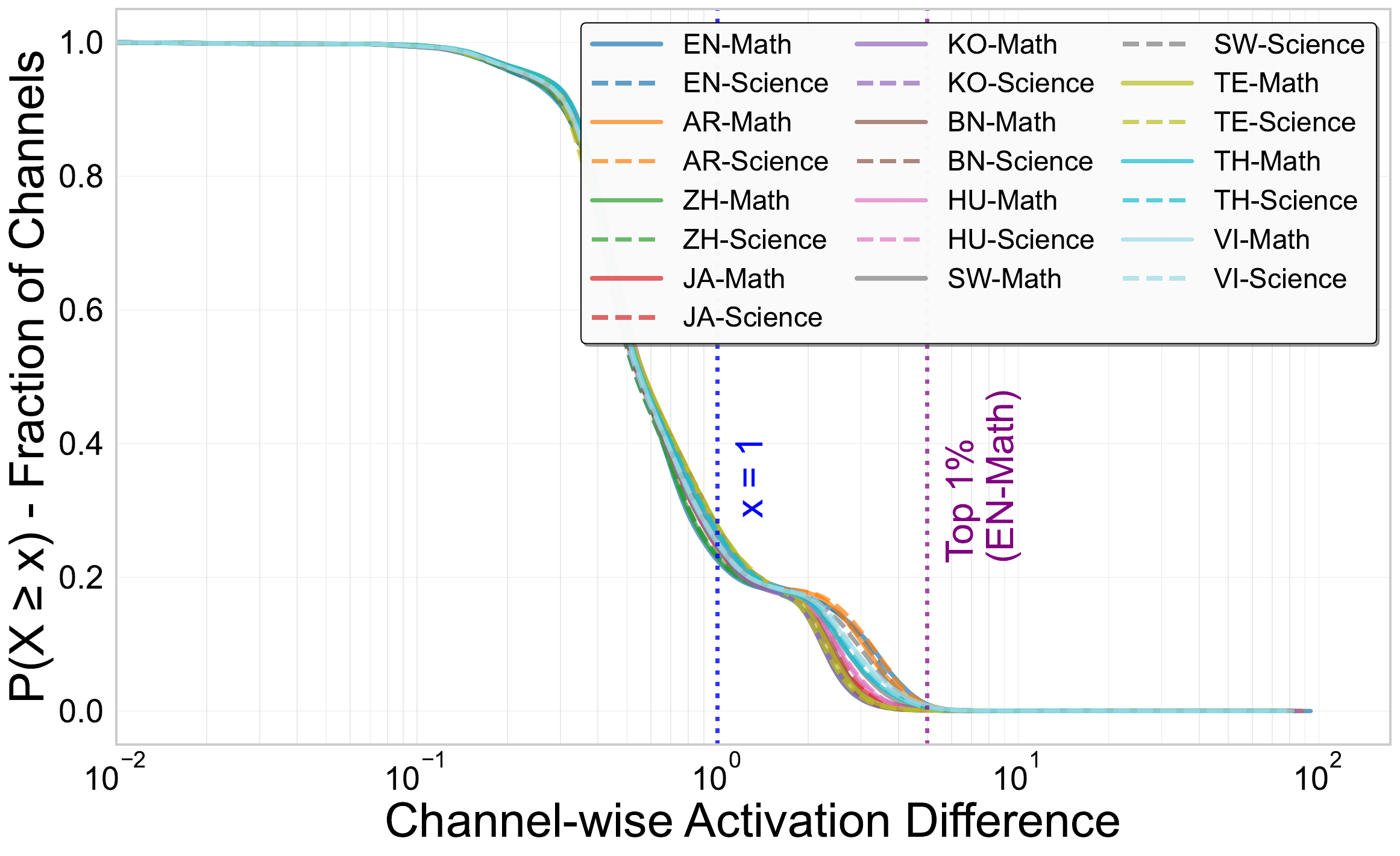}
    \caption{
    \textbf{Qwen2.5-Coder-1.5B-Instruct vs.\ Qwen2.5-1.5B-Instruct.}
    CCDFs of channel-wise activation differences across 11 languages for math (solid) and science (dashed).
    The heavy-tailed structure is consistent at 1.5B scale.
    }
    \label{fig:app_ccdf_qwen_coder_15b_all_lang}
\end{figure}

\paragraph{Qwen2.5 Coder Models.}
We repeat the same multilingual math/science analysis for code-specialized models.
Figures~\ref{fig:app_ccdf_qwen_coder_7b_all_lang} and~\ref{fig:app_ccdf_qwen_coder_15b_all_lang} show CCDFs for Qwen2.5-Coder-Instruct versus Qwen2.5-Instruct at 7B and 1.5B.
Despite different fine-tuning objectives, activation differences for these abilities remain sparse and heavy-tailed.

% --------------------------
% Qwen2.5 Coder (7B / 1.5B)
% Replace filenames if needed.
% --------------------------

\begin{figure}[htbp]
    \centering
    \includegraphics[width=\linewidth]{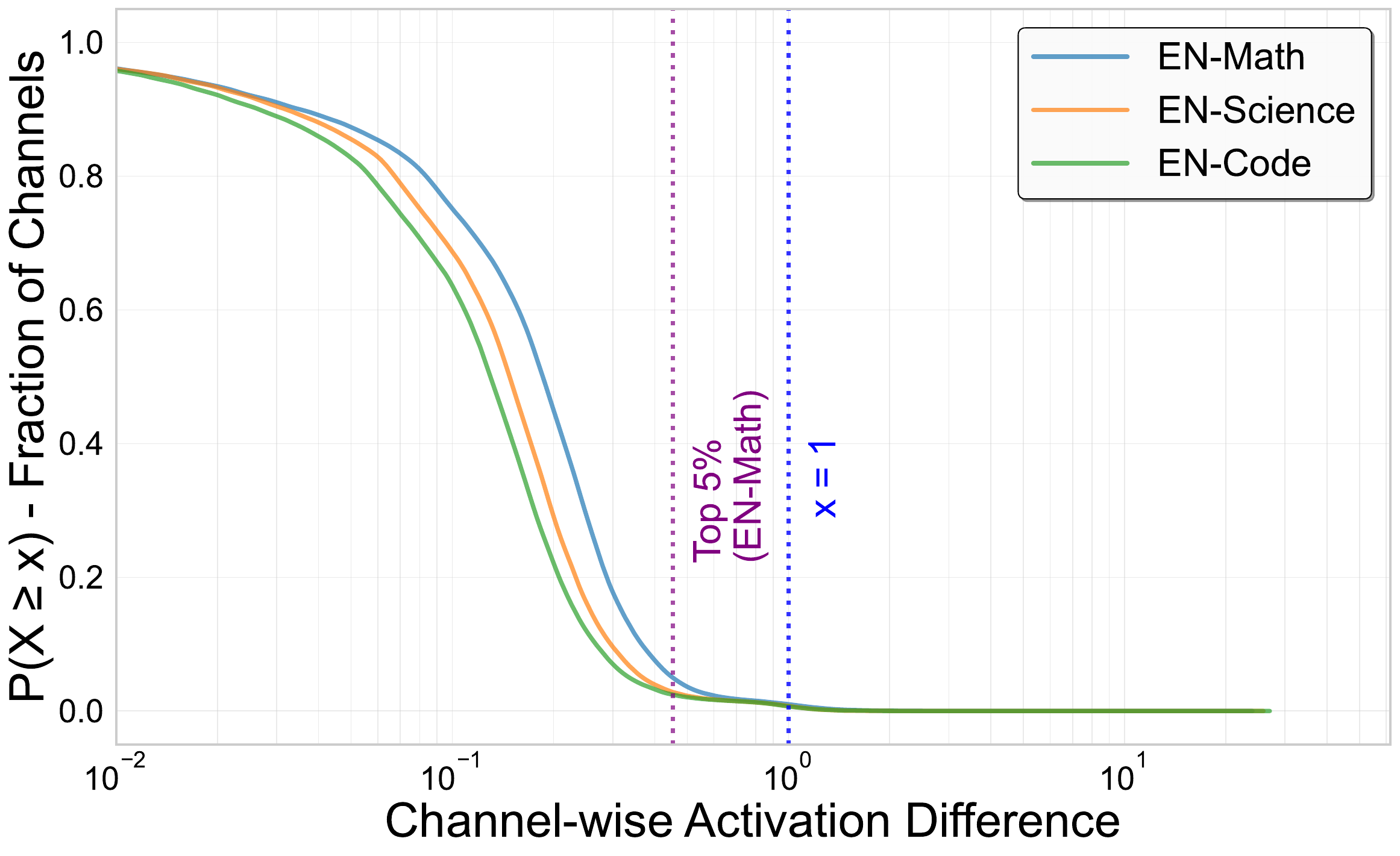}
    \caption{
    \textbf{WizardMath-13B vs.\ LLaMA-2-13B.}
    CCDFs of channel-wise activation differences on English math, science, and code inputs.
    Activation differences are heavy-tailed, with large deviations concentrated in a small subset of channels.
    }
    \label{fig:app_ccdf_llama_wizardmath}
\end{figure}

\begin{figure}[htbp]
    \centering
    \includegraphics[width=\linewidth]{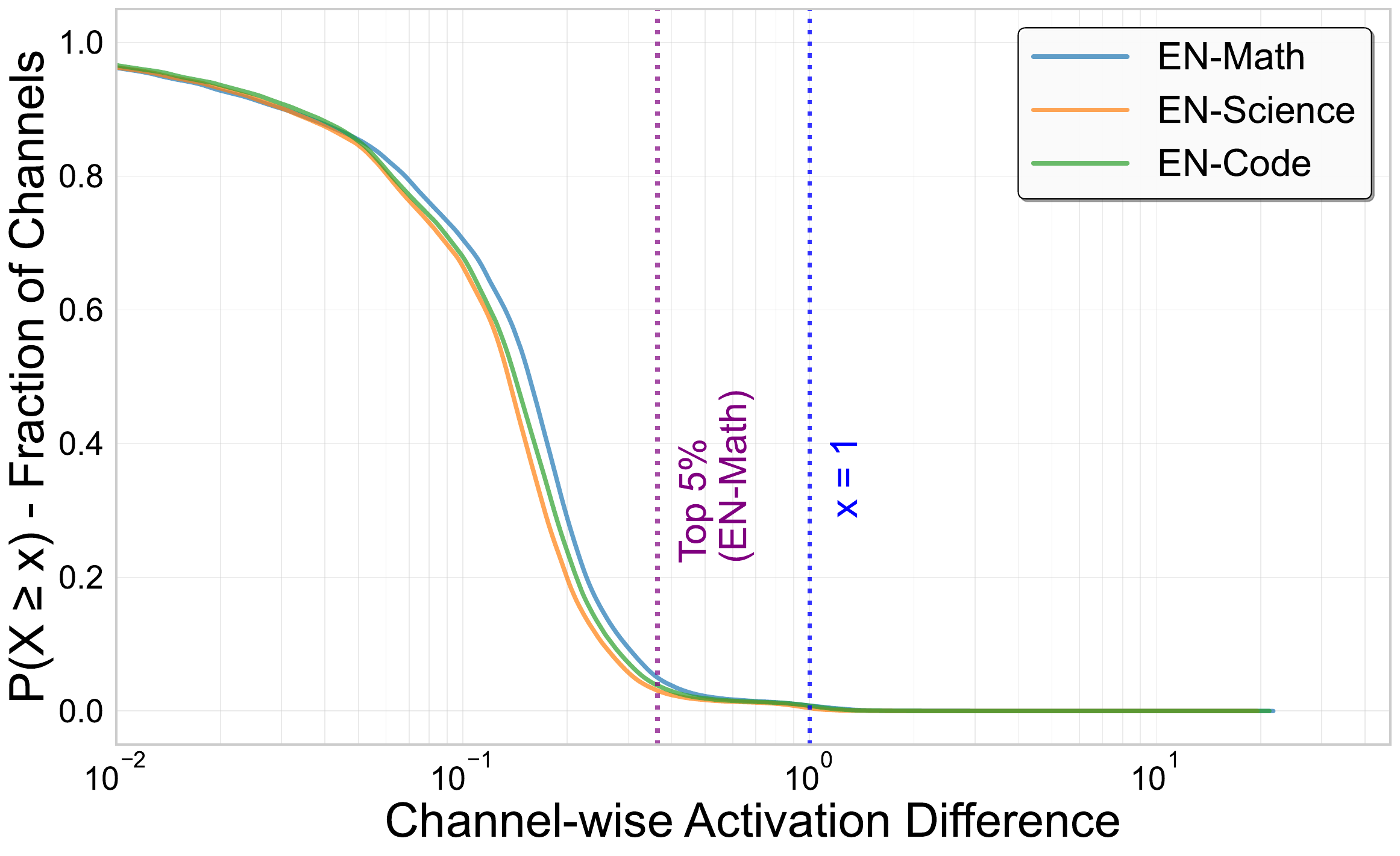}
    \caption{
    \textbf{Tulu-2-13B vs.\ LLaMA-2-13B.}
    CCDFs of channel-wise activation differences on English math, science, and code inputs.
    Activation differences remain heavy-tailed across ability domains.
    % The sparse, heavy-tailed pattern persists under science-oriented specialization.
    }
    \label{fig:app_ccdf_llama_tulu}
\end{figure}

\begin{figure}[htbp]
    \centering
    \includegraphics[width=\linewidth]{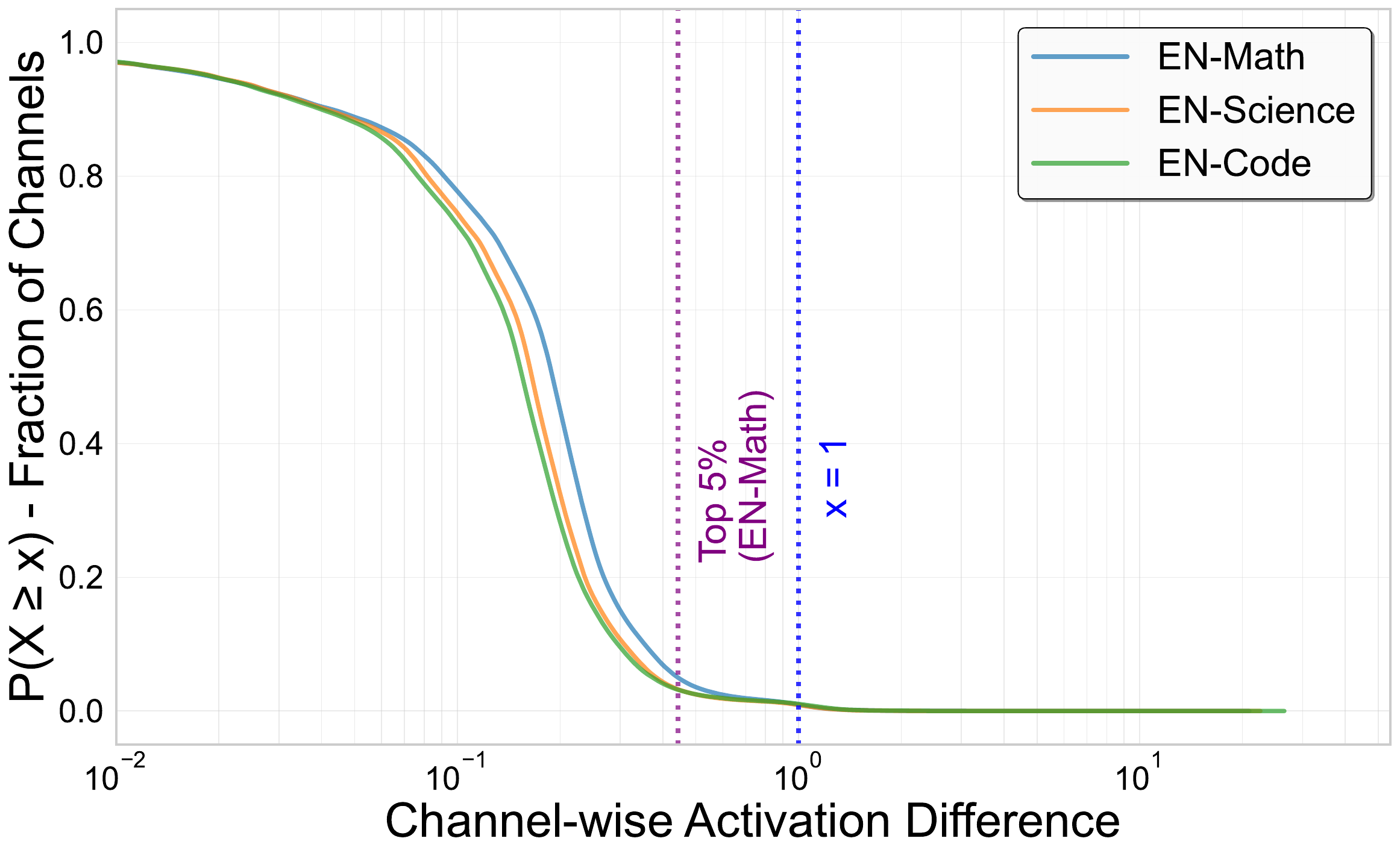}
    \caption{
    \textbf{WizardLM-13B vs.\ LLaMA-2-13B.}
    CCDFs of channel-wise activation differences on English math, science, and code inputs.
    Activation differences remain heavy-tailed across ability domains.
    }
    \label{fig:app_ccdf_llama_wizardlm}
\end{figure}

\paragraph{LLaMA-2 Family Models.}
To assess generalization beyond Qwen, we further analyze model pairs derived from LLaMA-2-13B.
We compare the base model with WizardMath-13B, WizardLM-13B, and Tulu-2-13B, which are specialized for mathematics, code, and scientific reasoning, respectively.
Due to limited multilingual coverage, these experiments focus on English only.
We compute activation differences using 1{,}500 samples each for English math and science (the same datasets as in the main analysis), and 1{,}500 English code samples from OpenCodeInstruct~\citep{ahmad2025opencodeinstructlargescaleinstructiontuning}.
Figures~\ref{fig:app_ccdf_llama_wizardmath}, \ref{fig:app_ccdf_llama_tulu}, and \ref{fig:app_ccdf_llama_wizardlm} report CCDFs for English math, science, and code.
All three comparisons again exhibit heavy-tailed behavior, with a small subset of channels accounting for most cross-model discrepancies.
Interestingly, while the distribution shapes are similar, the overall activation-difference magnitudes are generally smaller than those observed in the Qwen2.5 family.

% --------------------------
% LLaMA-2 family (13B)
% --------------------------

\paragraph{Summary.}
Overall, these results reinforce Finding~1: sparse, heavy-tailed cross-model activation differences are a robust phenomenon across scales, model families, and ability domains.
This supports using activation differences as a principled signal for localizing and selectively transferring ability-relevant channels.

\subsection{Additional Results for Finding~2}
\label{app:finding2}

In the main text (Finding~2), we show that large cross-model activation differences are sparse yet broadly distributed across decoder layers and module types, exhibiting consistent heavy-tailed patterns with moderate structural concentration.
Here, we provide additional results to verify that this structural behavior is robust across model families, scales, abilities, and languages.

Rather than exhaustively enumerating all possible model--language-domain combinations, we report a representative but diverse set of configurations that jointly vary model family, model scale, specialization objective, language, and domain.
This design allows us to validate the generality of Finding~2 without redundant repetition.

\begin{figure*}[htbp]
    \centering

    % ---------- EN-Math ----------
    \begin{subfigure}[t]{0.45\textwidth}
        \centering
        \includegraphics[width=\linewidth]{images/Distribution_of_Activation_Differences/Qwen_Qwen2.5-7B-Instruct+Qwen_Qwen2.5-Math-7B-Instruct_metamathqa_en_tokendiff_assistant_only_ccdf_by_layer.pdf}
        \caption{EN-Math (per layer)}
        \label{fig:app_f2_qwen_math_7b_en_math_layer}
    \end{subfigure}
    \hspace{10pt}
    % \hfill
    \begin{subfigure}[t]{0.45\textwidth}
        \centering
        \includegraphics[width=\linewidth]{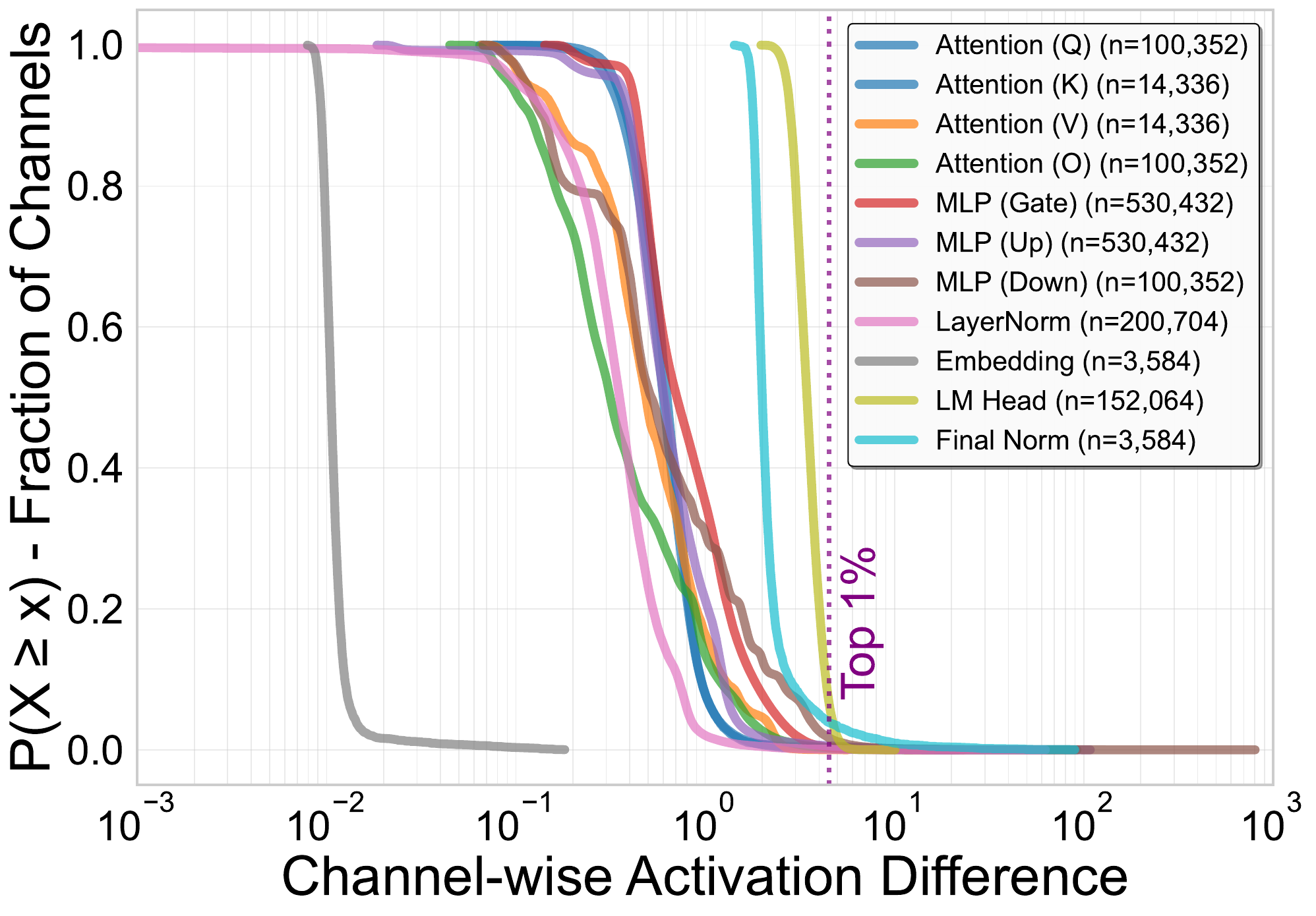}
        \caption{EN-Math (per module)}
        \label{fig:app_f2_qwen_math_7b_en_math_module}
    \end{subfigure}

    \vspace{0.5em}

    % ---------- EN-Science ----------
    \begin{subfigure}[t]{0.45\textwidth}
        \centering
        \includegraphics[width=\linewidth]{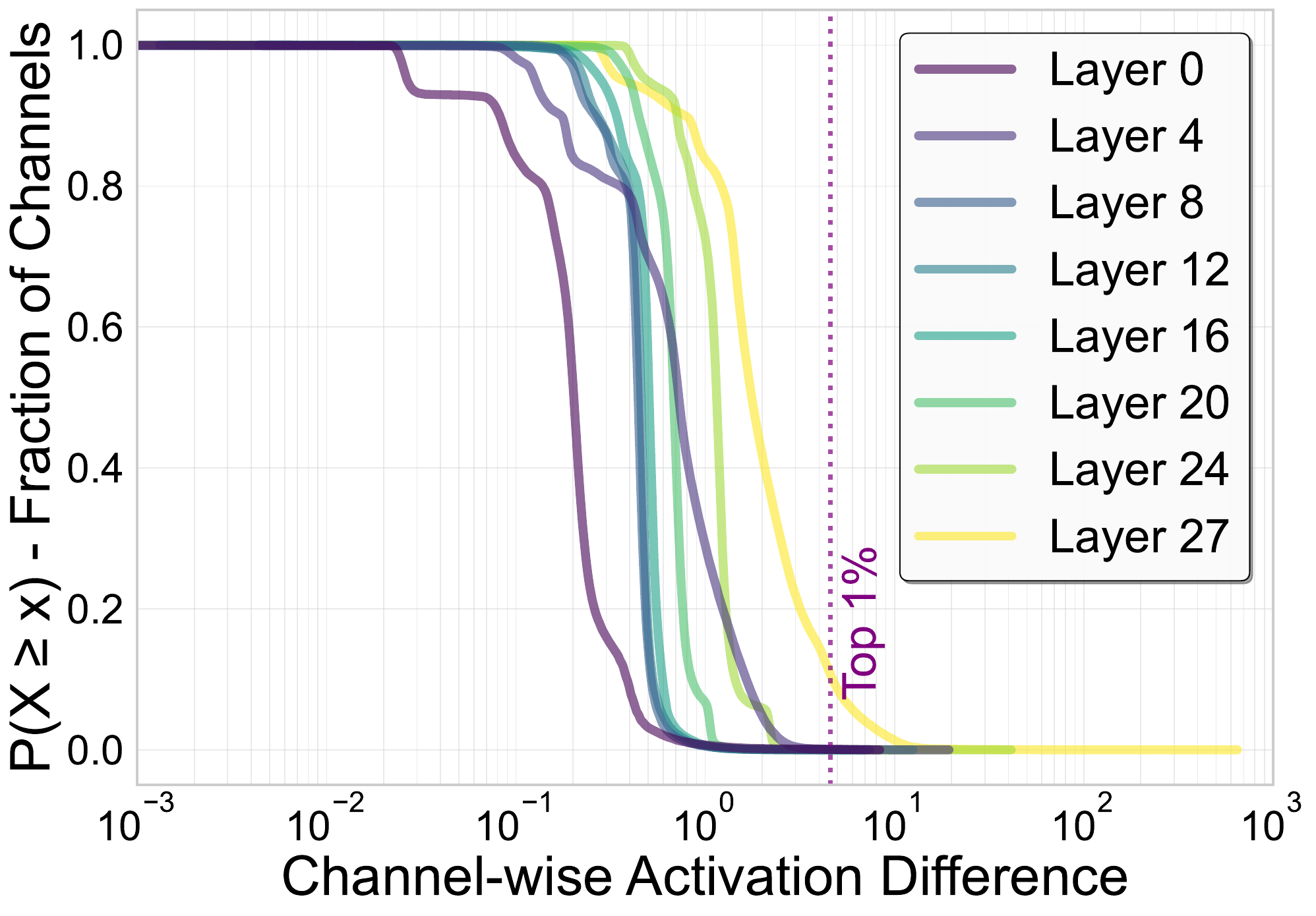}
        \caption{EN-Science (per layer)}
        \label{fig:app_f2_qwen_math_7b_en_sci_layer}
    \end{subfigure}
    \hspace{10pt}
    % \hfill
    \begin{subfigure}[t]{0.45\textwidth}
        \centering
        \includegraphics[width=\linewidth]{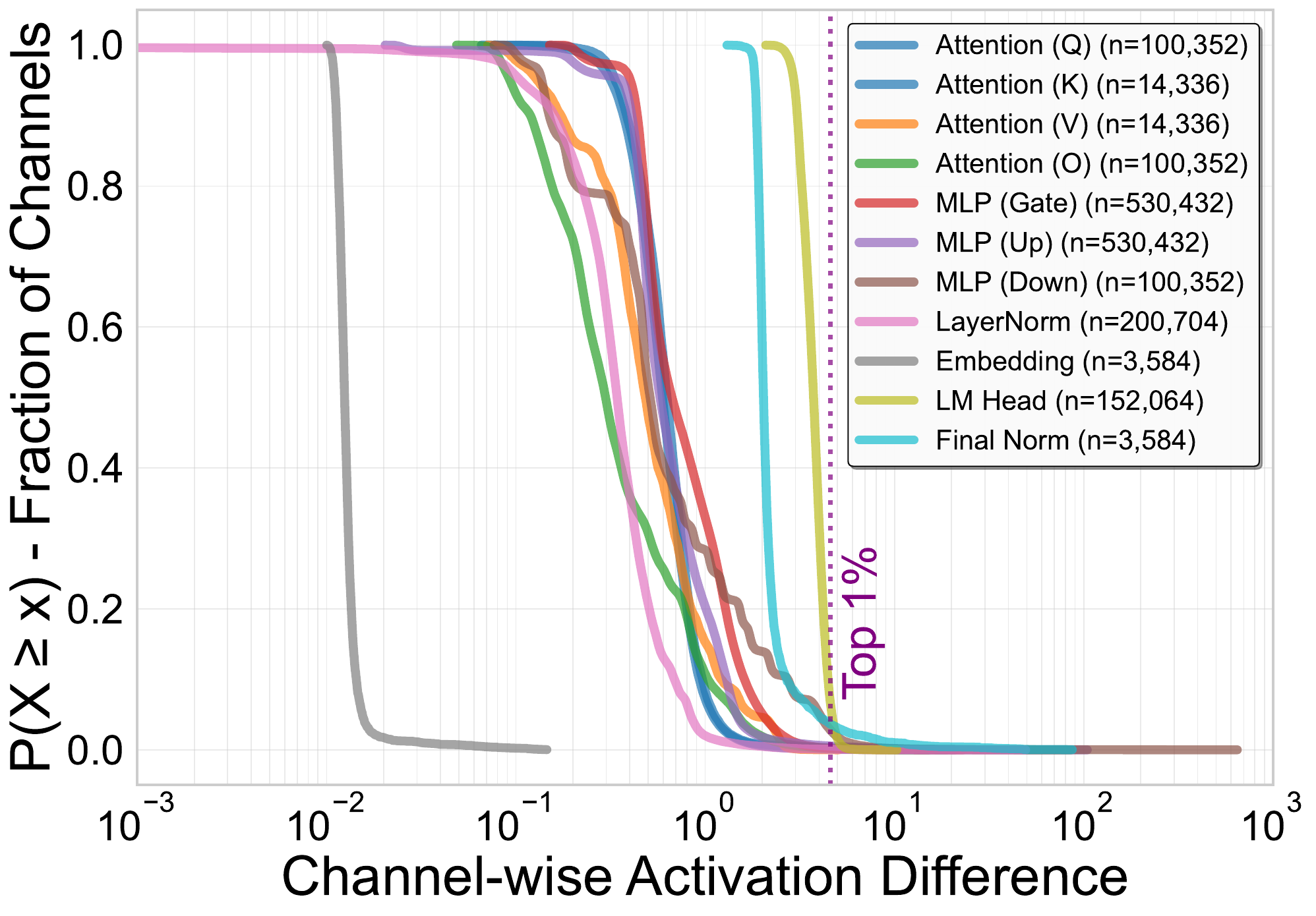}
        \caption{EN-Science (per module)}
        \label{fig:app_f2_qwen_math_7b_en_sci_module}
    \end{subfigure}

    \vspace{0.5em}

    % ---------- AR-Math ----------
    \begin{subfigure}[t]{0.45\textwidth}
        \centering
        \includegraphics[width=\linewidth]{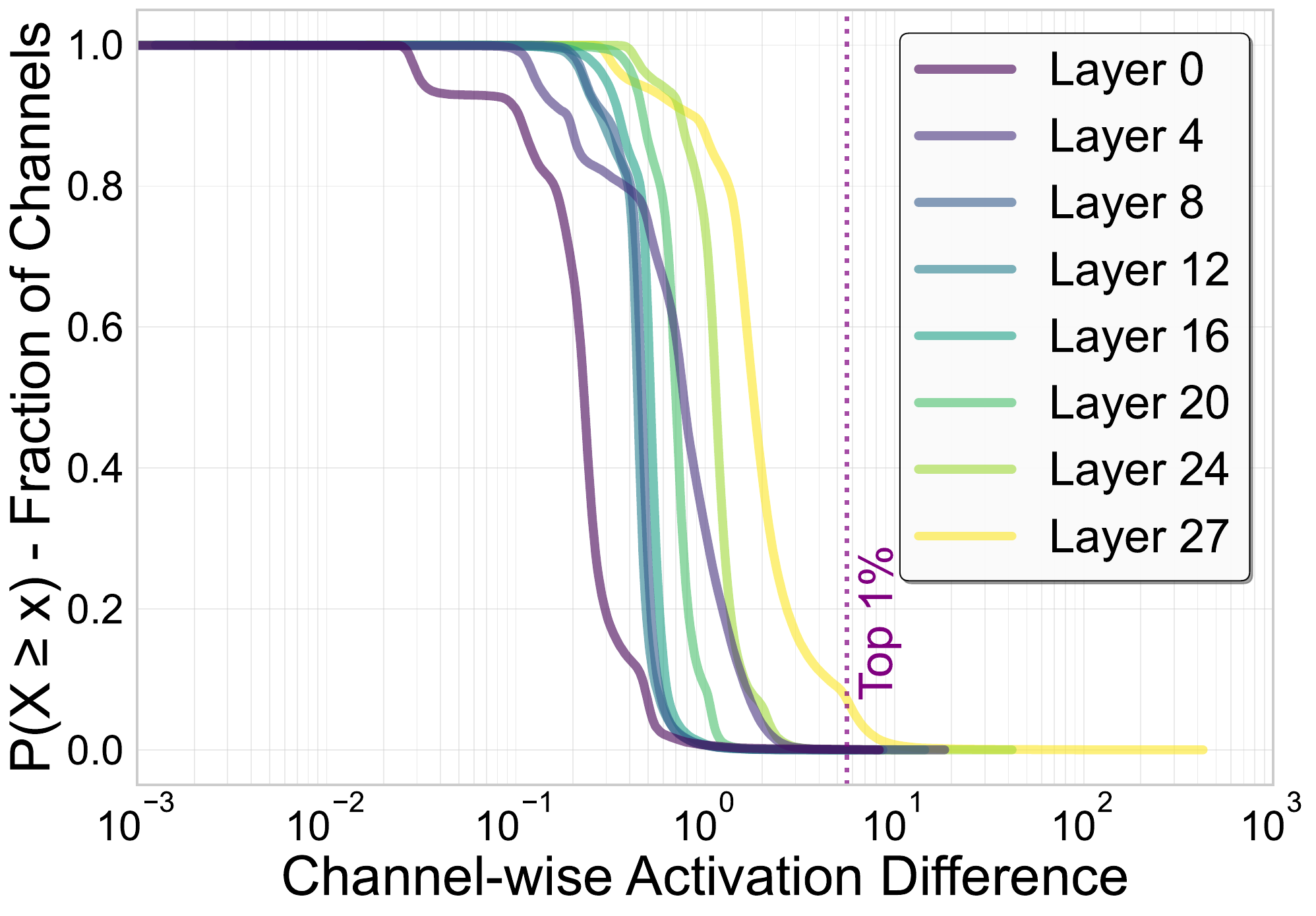}
        \caption{AR-Math (per layer)}
        \label{fig:app_f2_qwen_math_7b_ar_math_layer}
    \end{subfigure}
    \hspace{10pt}
    % \hfill
    \begin{subfigure}[t]{0.45\textwidth}
        \centering
        \includegraphics[width=\linewidth]{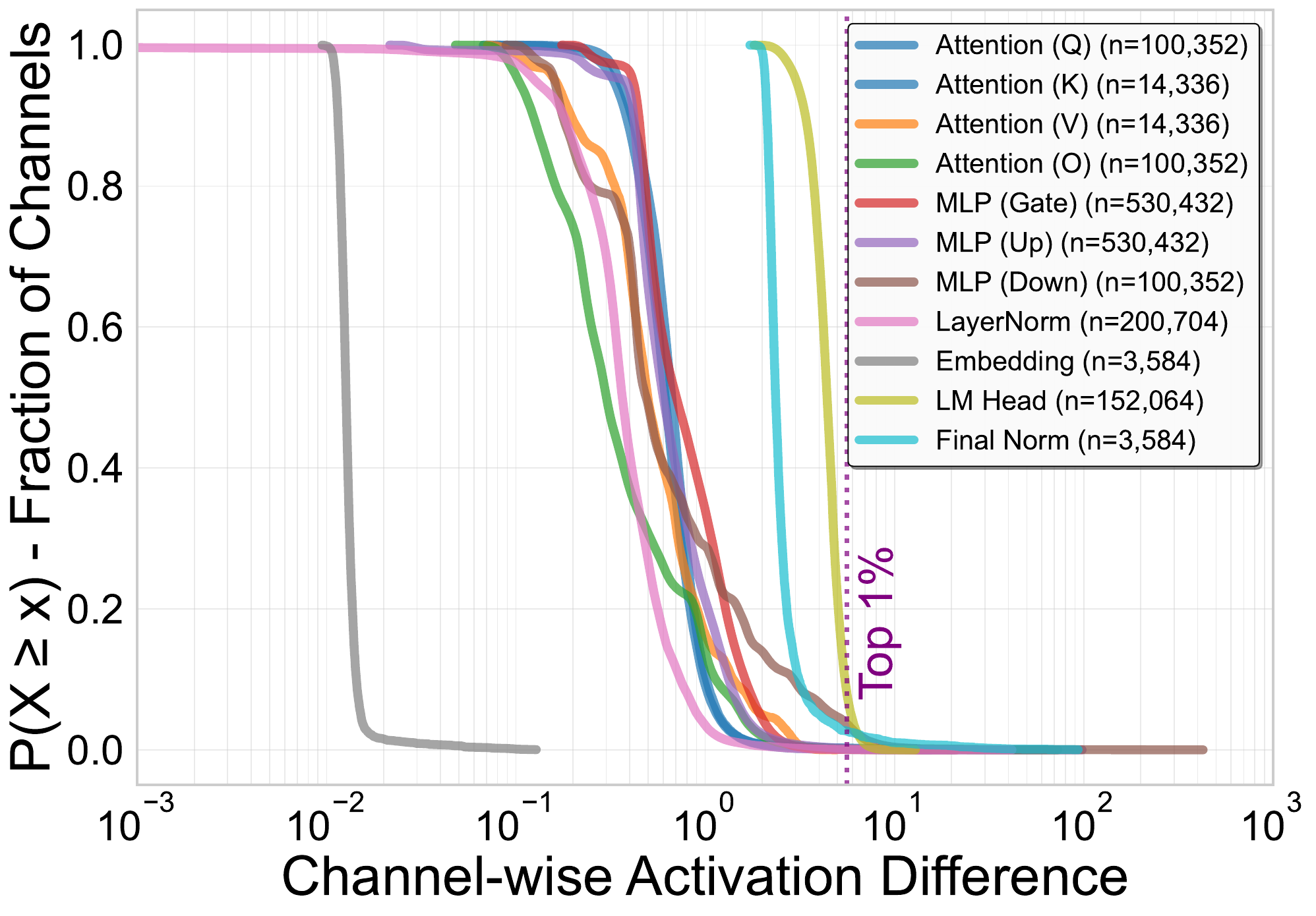}
        \caption{AR-Math (per module)}
        \label{fig:app_f2_qwen_math_7b_ar_math_module}
    \end{subfigure}

    \vspace{0.5em}

    % ---------- AR-Science ----------
    \begin{subfigure}[t]{0.45\textwidth}
        \centering
        \includegraphics[width=\linewidth]{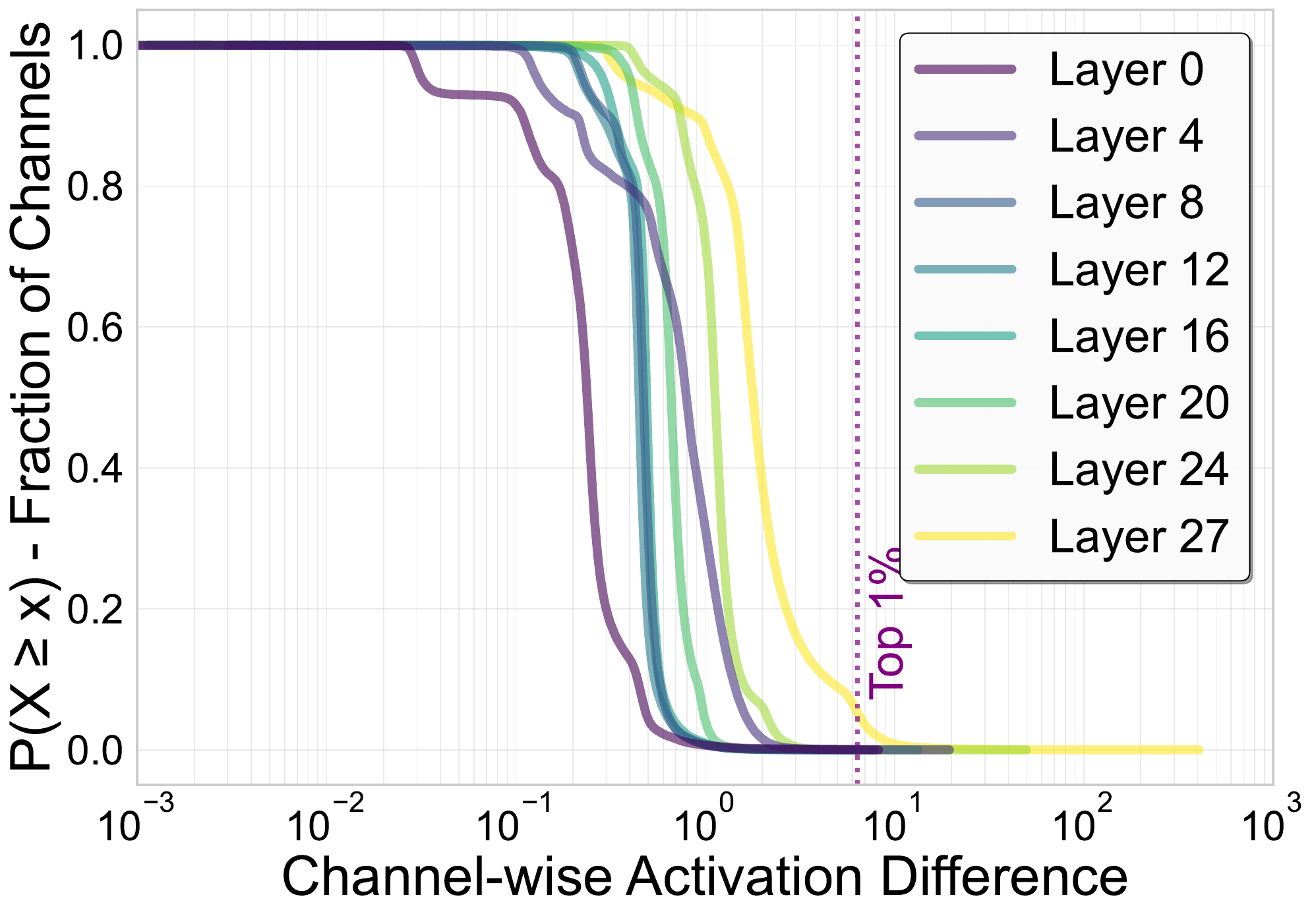}
        \caption{AR-Science (per layer)}
        \label{fig:app_f2_qwen_math_7b_ar_sci_layer}
    \end{subfigure}
    \hspace{10pt}
    % \hfill
    \begin{subfigure}[t]{0.45\textwidth}
        \centering
        \includegraphics[width=\linewidth]{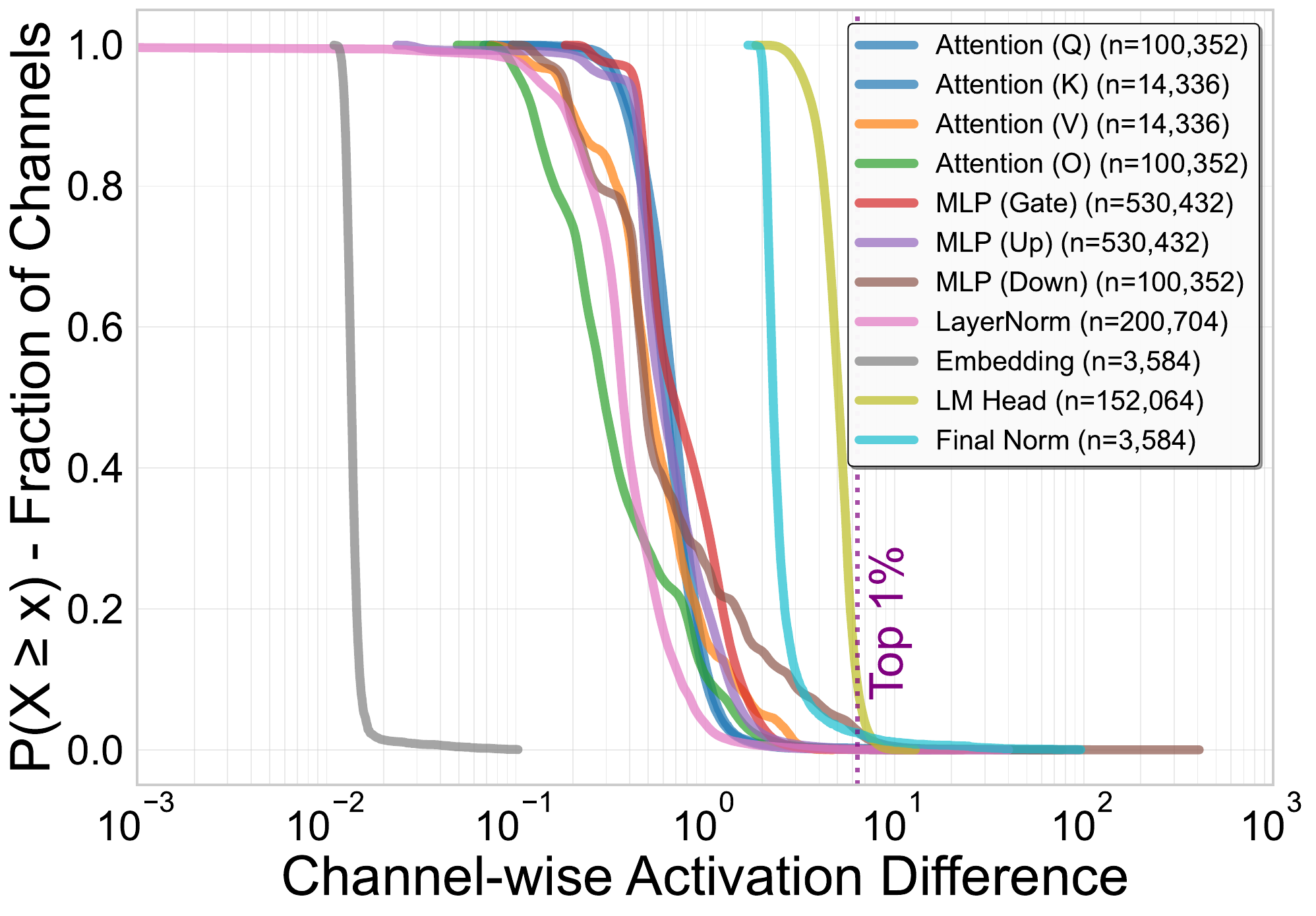}
        \caption{AR-Science (per module)}
        \label{fig:app_f2_qwen_math_7b_ar_sci_module}
    \end{subfigure}

    \caption{
    Additional results for Finding~2 on \textbf{Qwen2.5-Math-7B-Instruct} vs.\ \textbf{Qwen2.5-7B-Instruct}.
    Each row corresponds to a language-domain ability, and columns show per-layer (left) and per-module (right) CCDFs of channel-wise activation differences.
    Across all abilities, activation differences exhibit consistent heavy-tailed distributions with moderate structural concentration.
    }
    \label{fig:app_f2_qwen_math_7b}
\end{figure*}

\begin{figure*}[htbp]
    \centering

    % ---------- EN-Math ----------
    \begin{subfigure}[t]{0.45\textwidth}
        \centering
        \includegraphics[width=\linewidth]{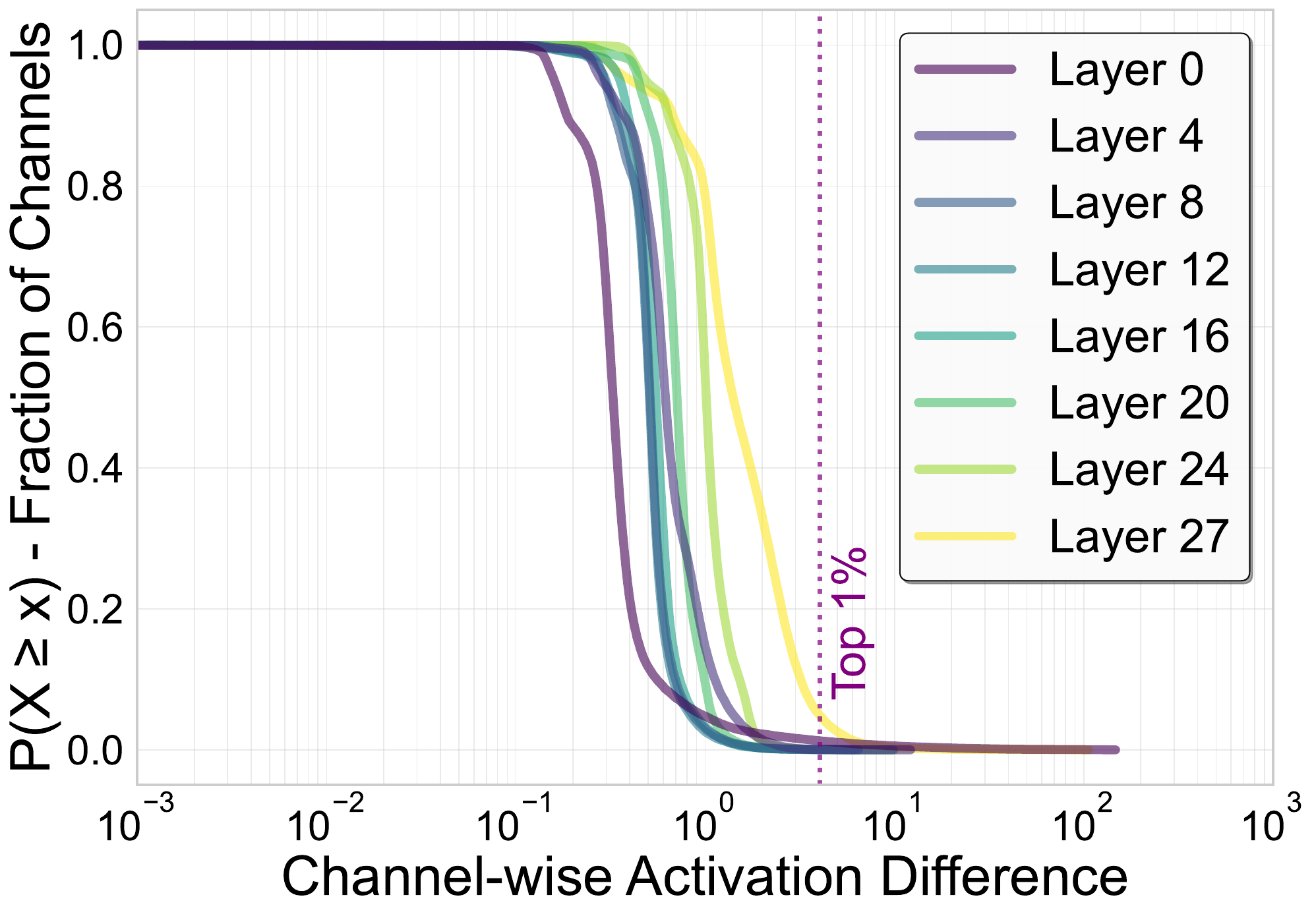}
        \caption{EN-Math (per layer)}
        \label{fig:app_f2_qwen_math_15b_en_math_layer}
    \end{subfigure}
    \hspace{10pt}
    % \hfill
    \begin{subfigure}[t]{0.45\textwidth}
        \centering
        \includegraphics[width=\linewidth]{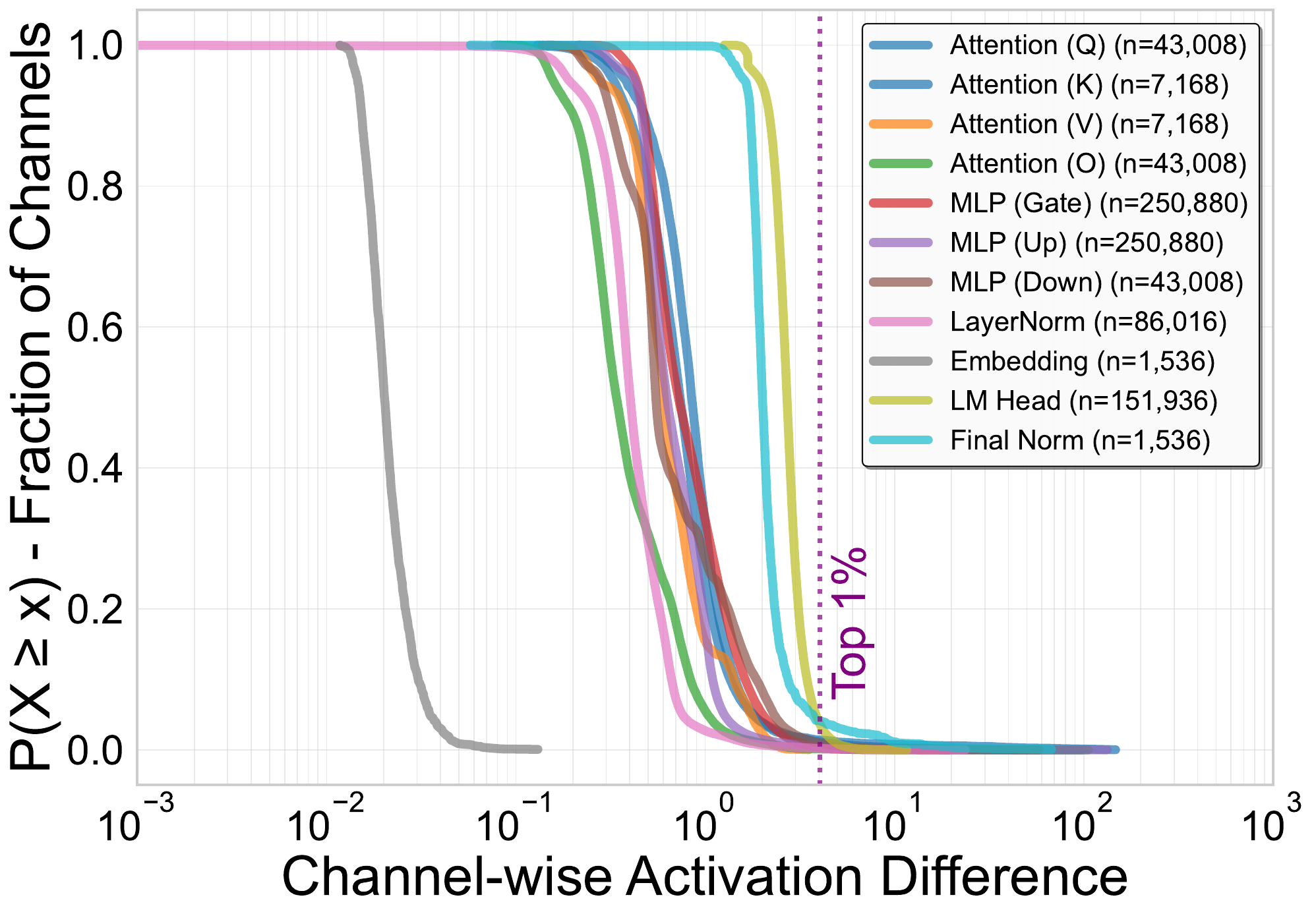}
        \caption{EN-Math (per module)}
        \label{fig:app_f2_qwen_math_15b_en_math_module}
    \end{subfigure}

    \vspace{0.5em}

    % ---------- EN-Science ----------
    \begin{subfigure}[t]{0.45\textwidth}
        \centering
        \includegraphics[width=\linewidth]{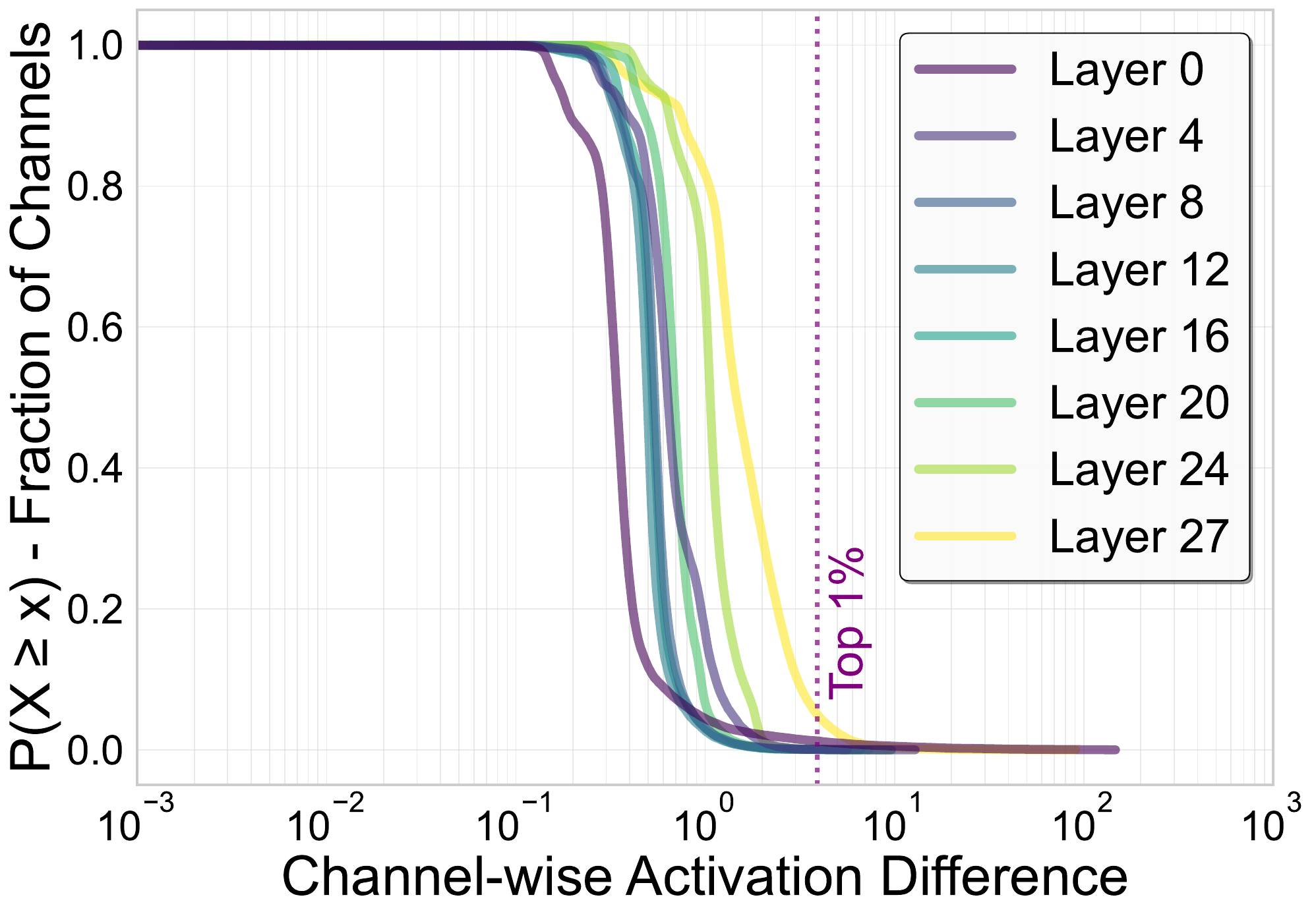}
        \caption{EN-Science (per layer)}
        \label{fig:app_f2_qwen_math_15b_en_sci_layer}
    \end{subfigure}
    \hspace{10pt}
    % \hfill
    \begin{subfigure}[t]{0.45\textwidth}
        \centering
        \includegraphics[width=\linewidth]{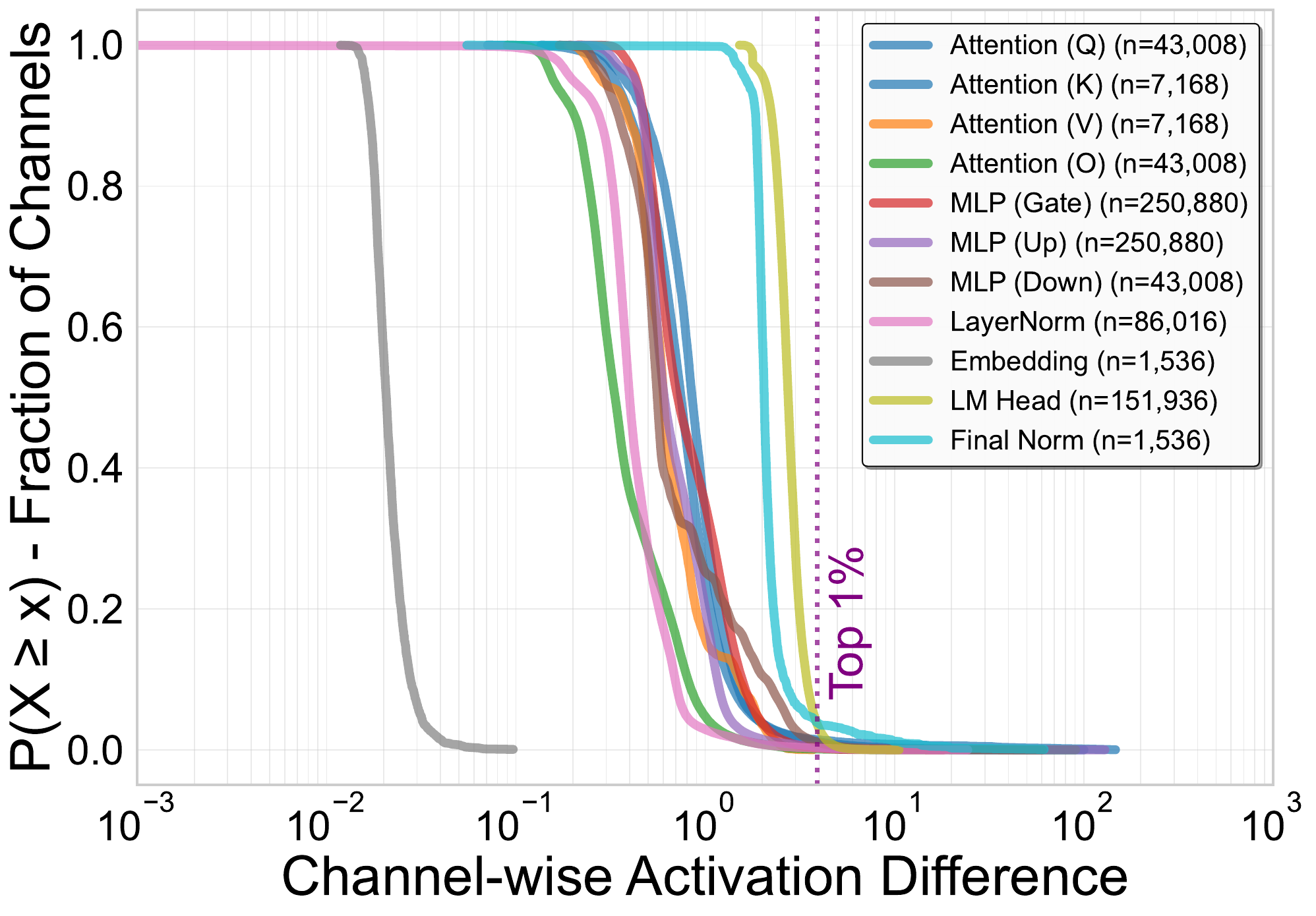}
        \caption{EN-Science (per module)}
        \label{fig:app_f2_qwen_math_15b_en_sci_module}
    \end{subfigure}

    \vspace{0.5em}

    % ---------- AR-Math ----------
    \begin{subfigure}[t]{0.45\textwidth}
        \centering
        \includegraphics[width=\linewidth]{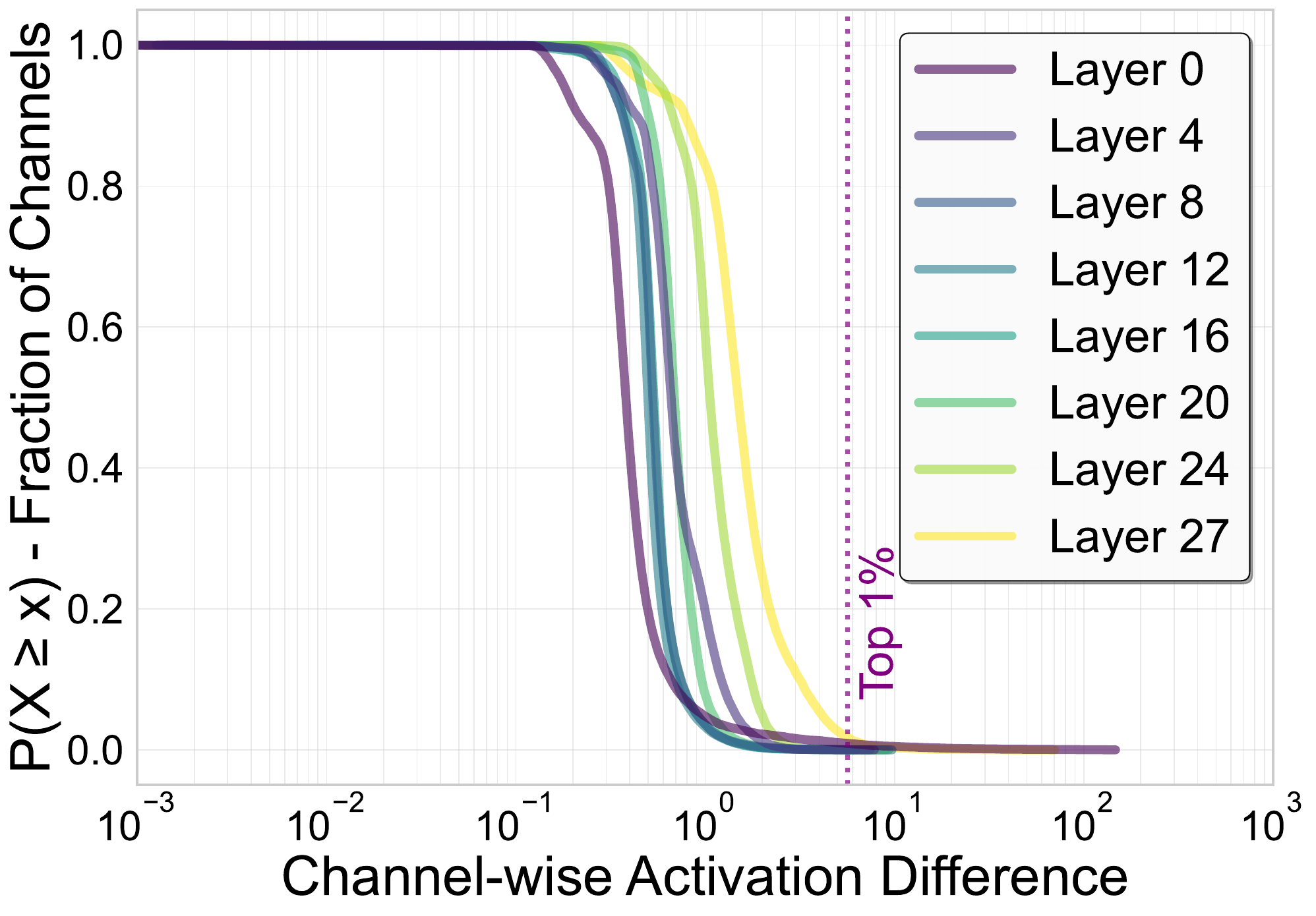}
        \caption{AR-Math (per layer)}
        \label{fig:app_f2_qwen_math_15b_ar_math_layer}
    \end{subfigure}
    \hspace{10pt}
    % \hfill
    \begin{subfigure}[t]{0.45\textwidth}
        \centering
        \includegraphics[width=\linewidth]{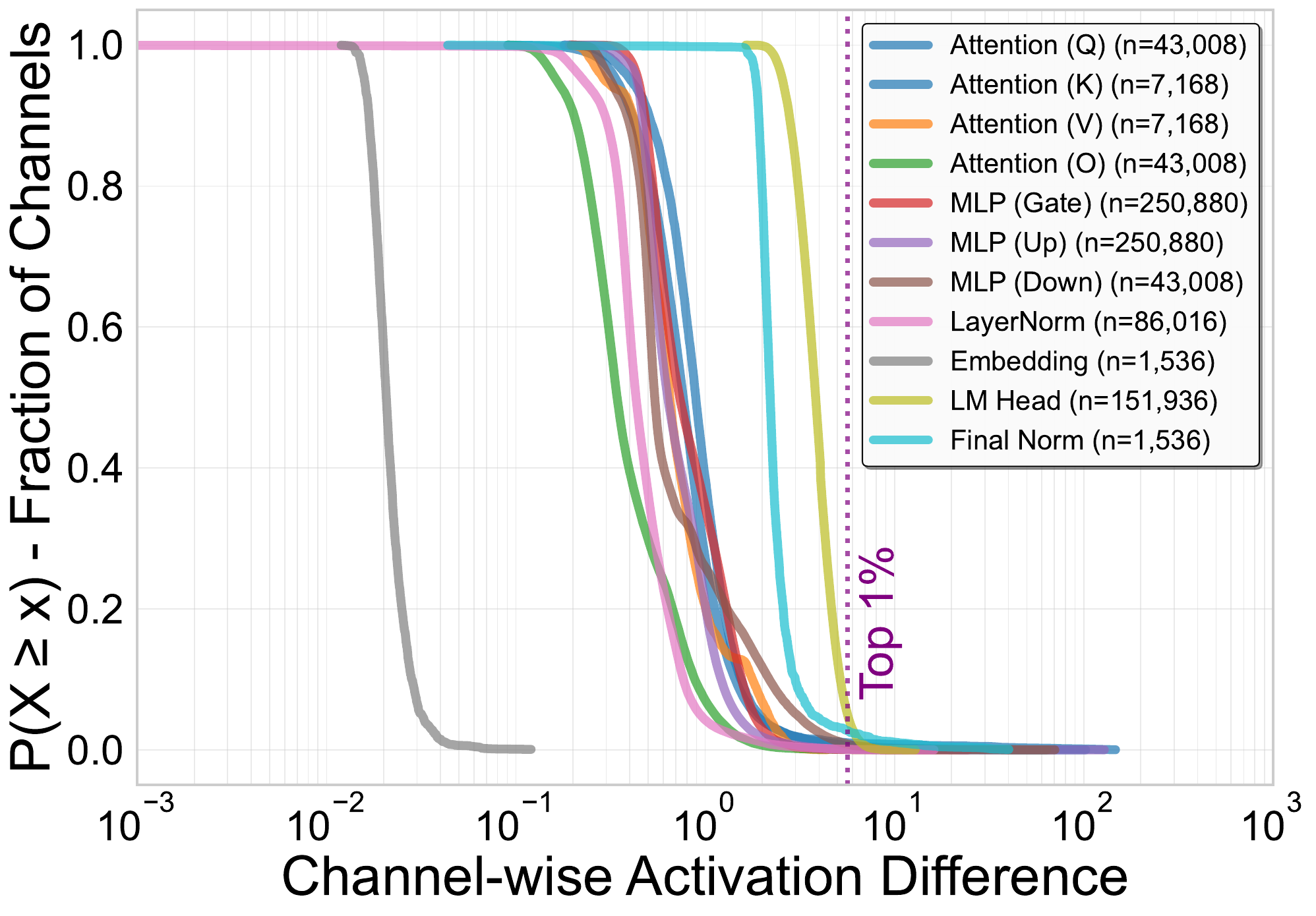}
        \caption{AR-Math (per module)}
        \label{fig:app_f2_qwen_math_15b_ar_math_module}
    \end{subfigure}

    \vspace{0.5em}

    % ---------- AR-Science ----------
    \begin{subfigure}[t]{0.45\textwidth}
        \centering
        \includegraphics[width=\linewidth]{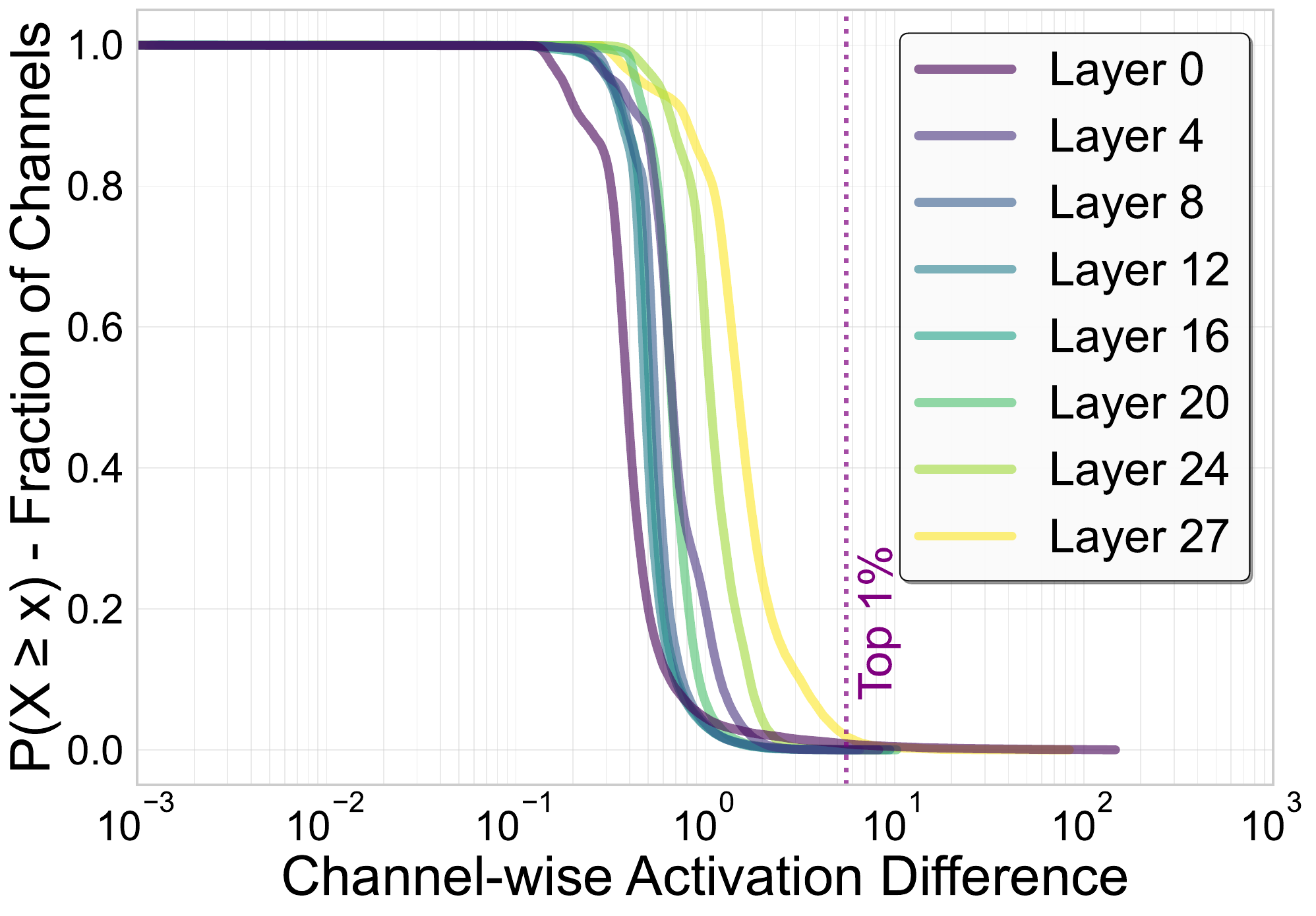}
        \caption{AR-Science (per layer)}
        \label{fig:app_f2_qwen_math_15b_ar_sci_layer}
    \end{subfigure}
    \hspace{10pt}
    % \hfill
    \begin{subfigure}[t]{0.45\textwidth}
        \centering
        \includegraphics[width=\linewidth]{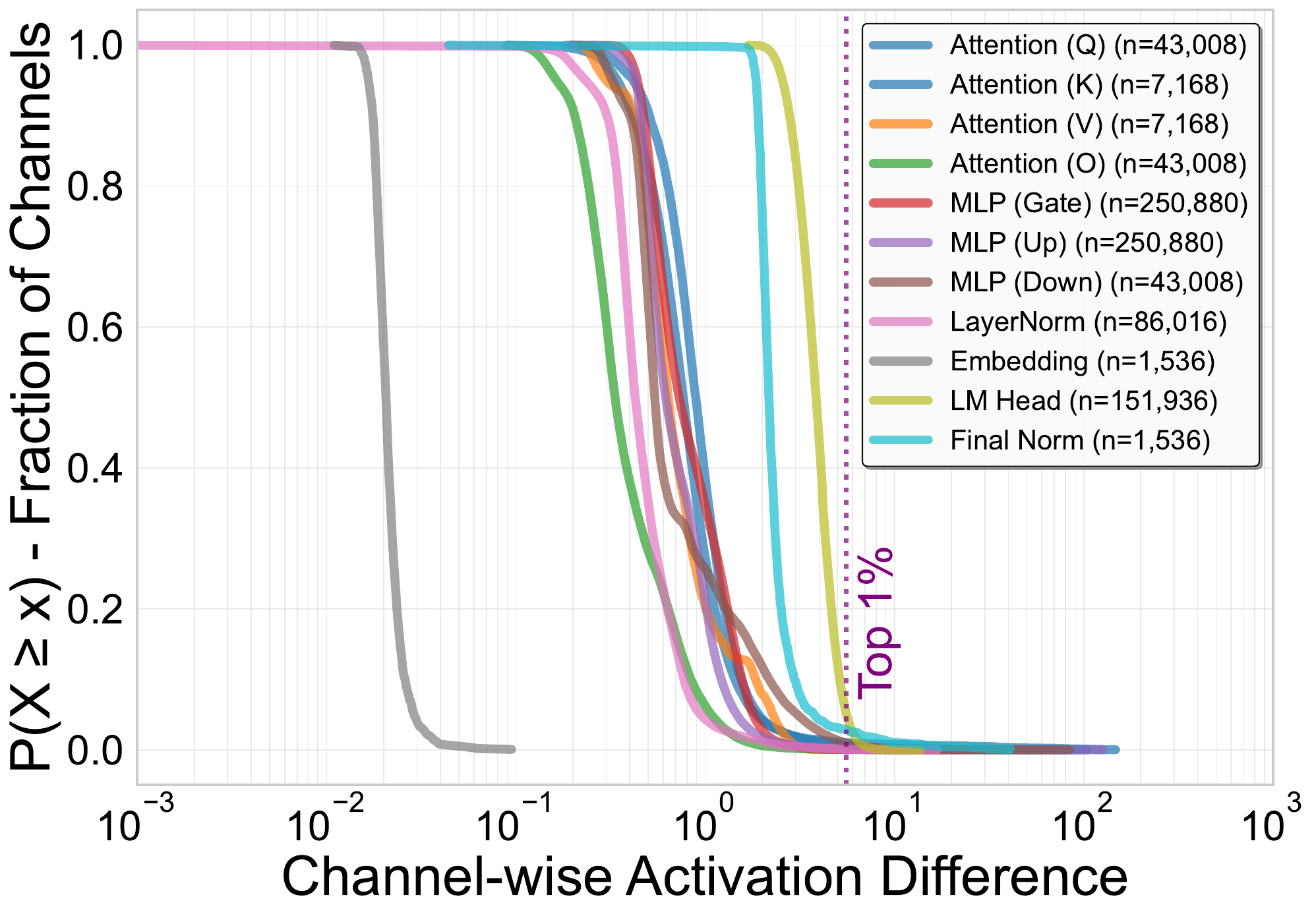}
        \caption{AR-Science (per module)}
        \label{fig:app_f2_qwen_math_15b_ar_sci_module}
    \end{subfigure}

    \caption{
    Additional results for Finding~2 on \textbf{Qwen2.5-Math-1.5B-Instruct} vs.\ \textbf{Qwen2.5-1.5B-Instruct}.
    The layout follows Figure~\ref{fig:app_f2_qwen_math_7b}.
    Across all language-domain abilities, activation differences exhibit consistent heavy-tailed distributions with moderate structural concentration.
    }
    \label{fig:app_f2_qwen_math_15b}
\end{figure*}

\begin{figure*}[htbp]
    \centering
    \begin{subfigure}[t]{0.45\linewidth}
        \centering
        \includegraphics[width=\linewidth]{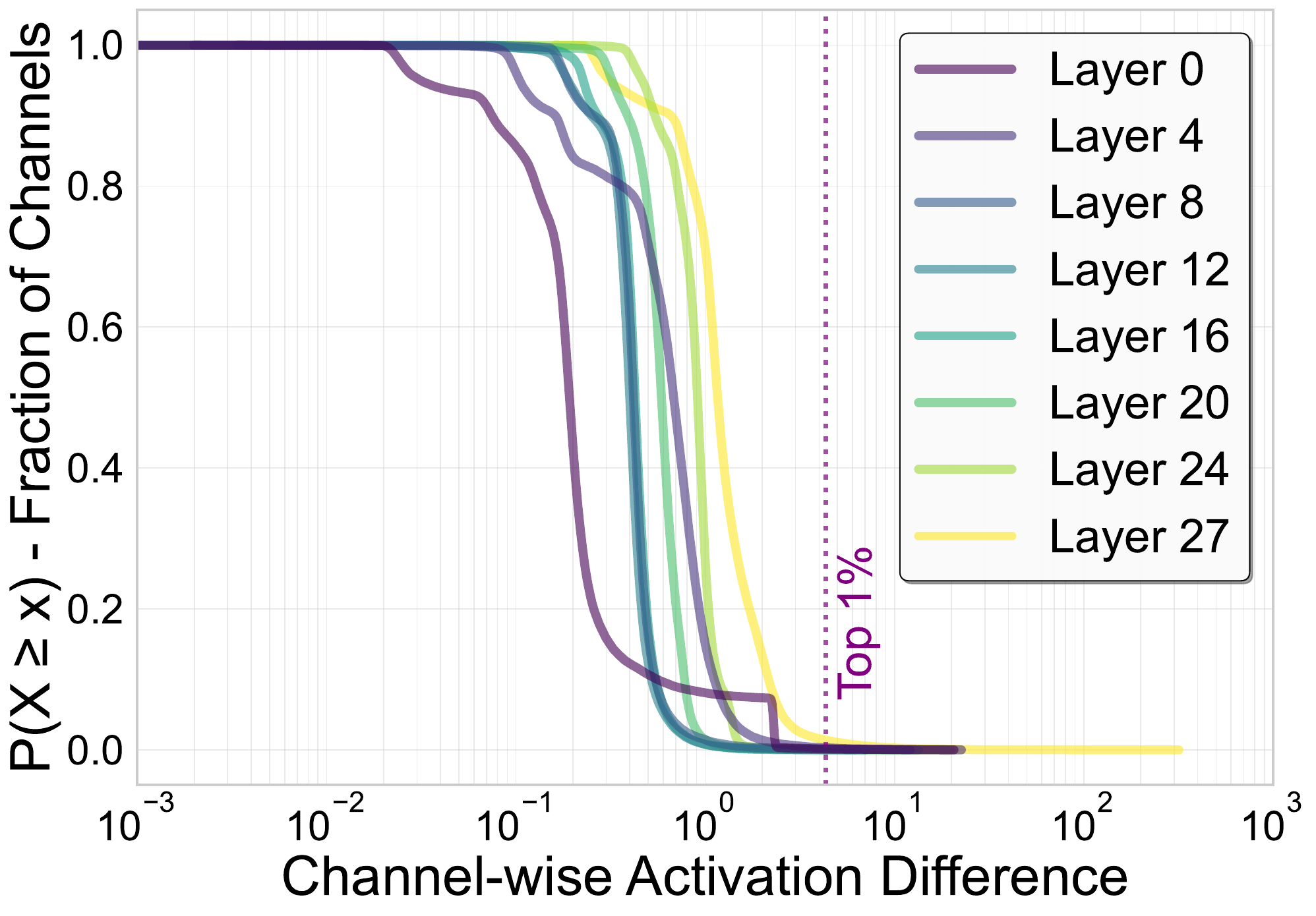}
        \caption{EN-Math (per layer)}
        \label{fig:app_f2_qwen_coder_7b_en_math_layer}
    \end{subfigure}
    \hspace{10pt}
    % \hfill
    \begin{subfigure}[t]{0.45\linewidth}
        \centering
        \includegraphics[width=\linewidth]{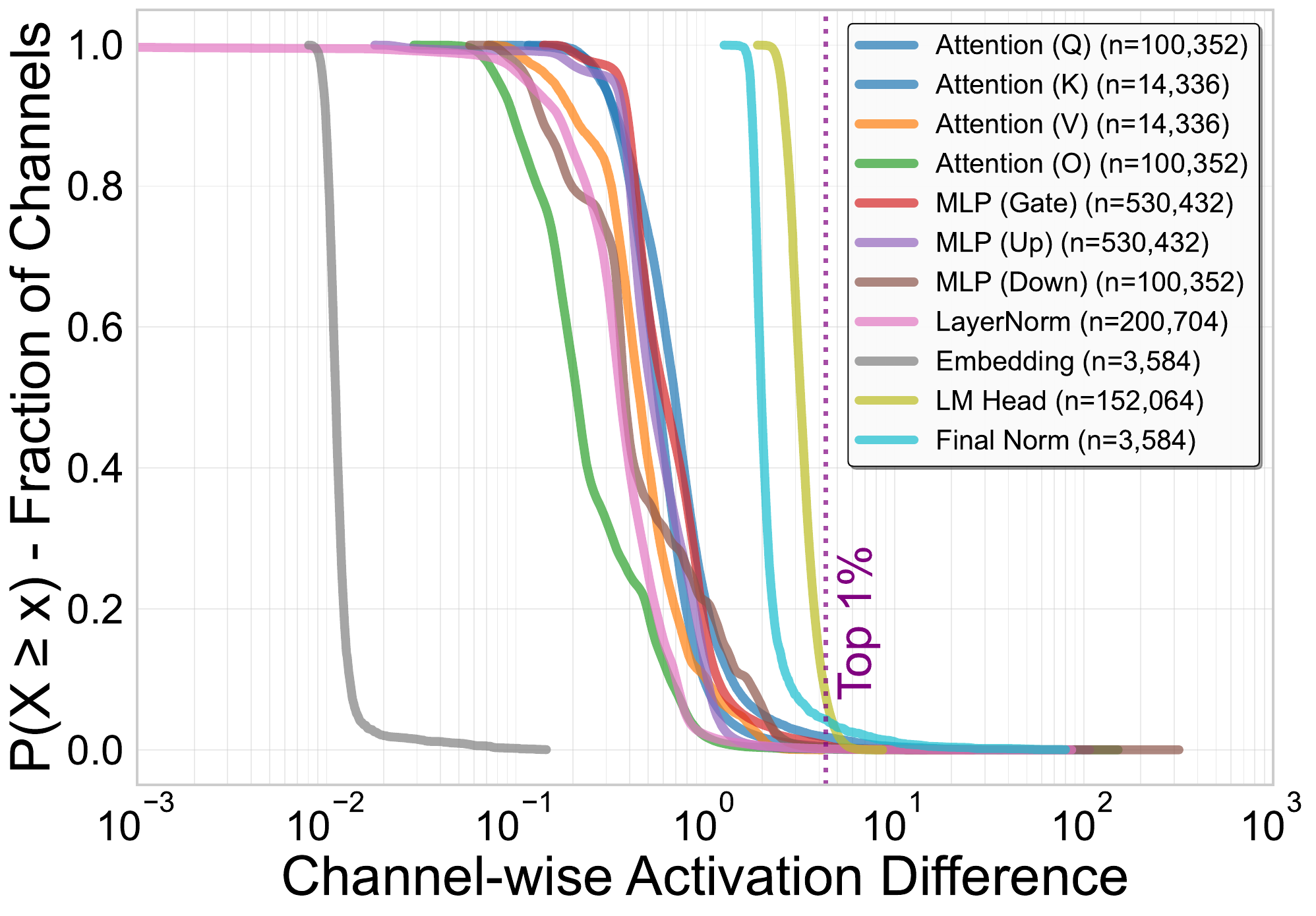}
        \caption{EN-Math (per module)}
        \label{fig:app_f2_qwen_coder_7b_en_math_module}
    \end{subfigure}

    \caption{
    Additional results for Finding~2 on \textbf{Qwen2.5-Coder-7B-Instruct} vs.\ \textbf{Qwen2.5-7B-Instruct}.
    Per-layer (left) and per-module (right) CCDFs of channel-wise activation differences for EN-Math.
    }
    \label{fig:app_f2_qwen_coder_7b}
\end{figure*}

\paragraph{Qwen2.5 Math Models.}
We first analyze layer- and module-wise activation difference distributions for Qwen2.5 Math-Instruct models.
Figure~\ref{fig:app_f2_qwen_math_7b} reports results for the 7B models on four representative abilities: English-Math, English-Science, Arabic-Math, and Arabic-Science.
Across all settings, activation differences exhibit consistent heavy-tailed distributions across layers and modules.
While deeper layers and the LM head tend to show mildly more and larger extrema, no single layer or module dominates the distribution. Figure~\ref{fig:app_f2_qwen_math_15b} presents the corresponding results for the 1.5B models.
Despite the reduced model scale, we observe the same qualitative behavior: sparse large deviations distributed broadly across layers and modules.
These results indicate that the structural pattern identified in Finding~2 is stable across both languages and model scales.

\begin{figure*}[htbp]
    \centering
    \begin{subfigure}[t]{0.45\linewidth}
        \centering
        \includegraphics[width=\linewidth]{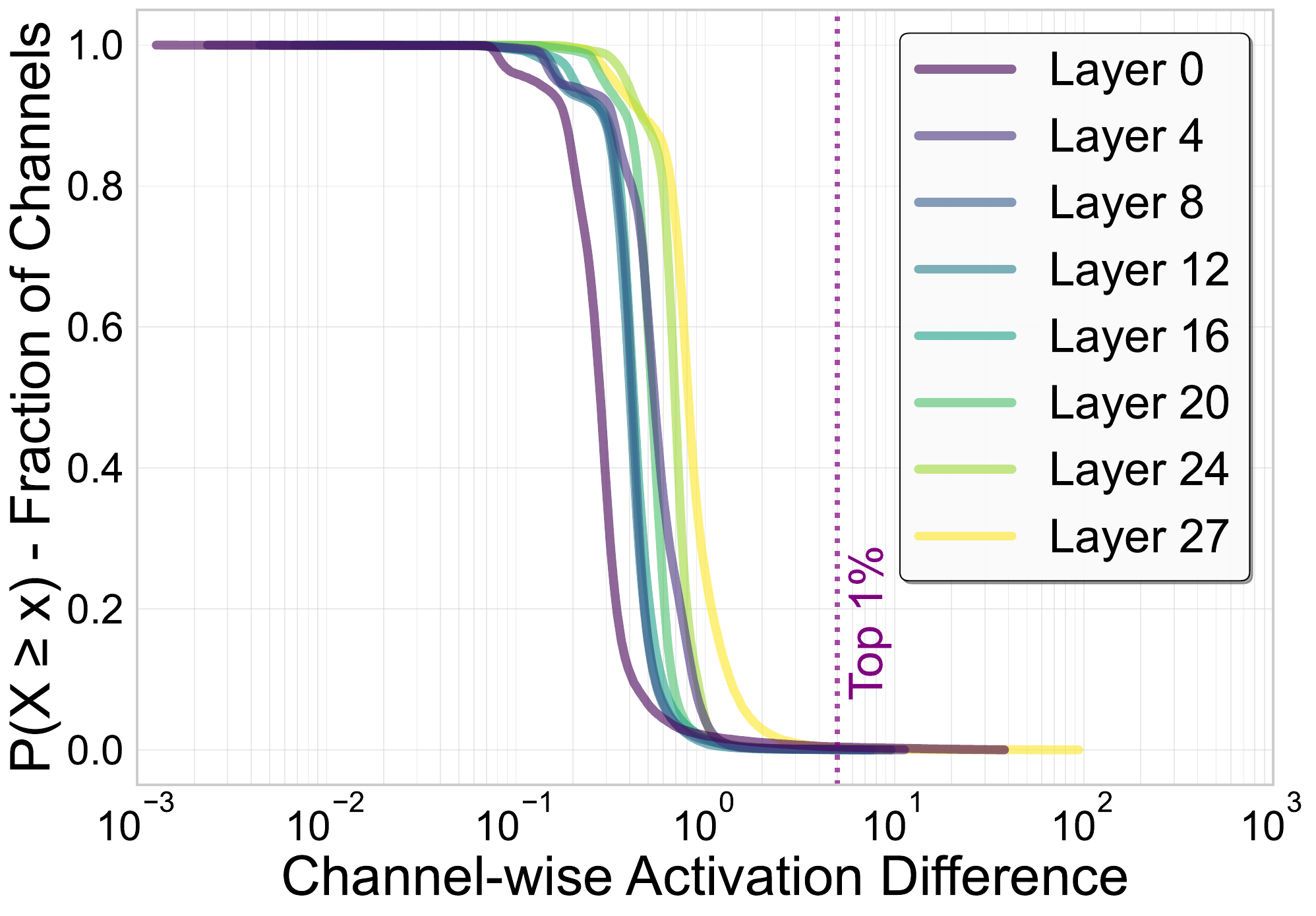}
        \caption{EN-Math (per layer)}
        \label{fig:app_f2_qwen_coder_15b_en_math_layer}
    \end{subfigure}
    \hspace{10pt}
    % \hfill
    \begin{subfigure}[t]{0.45\linewidth}
        \centering
        \includegraphics[width=\linewidth]{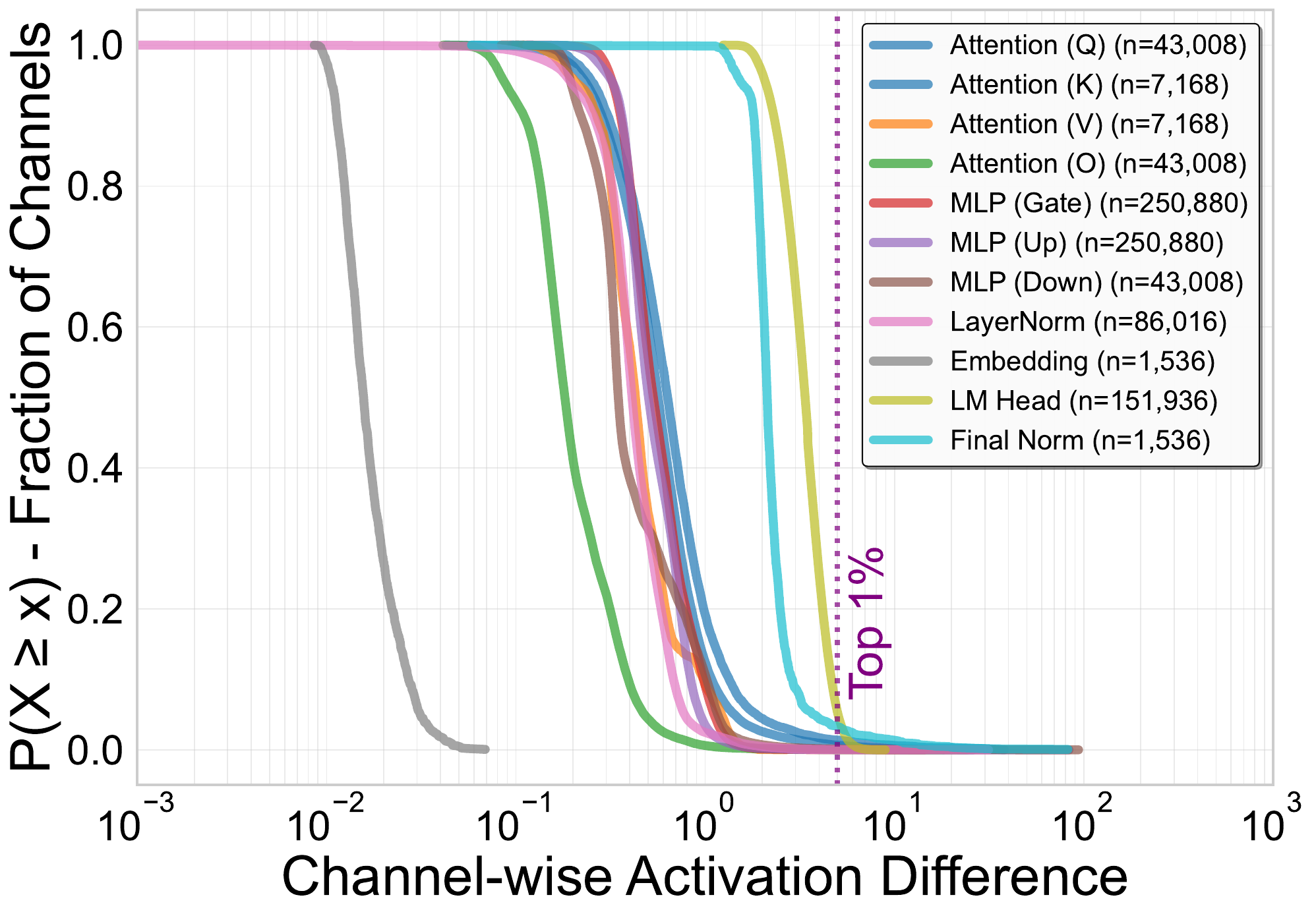}
        \caption{EN-Math (per module)}
        \label{fig:app_f2_qwen_coder_15b_en_math_module}
    \end{subfigure}

    \caption{
    Additional results for Finding~2 on \textbf{Qwen2.5-Coder-1.5B-Instruct} vs.\ \textbf{Qwen2.5-1.5B-Instruct}.
    Per-layer (left) and per-module (right) CCDFs of channel-wise activation differences for EN-Math.
    }
    \label{fig:app_f2_qwen_coder_15b}
\end{figure*}

\begin{figure*}[htbp]
    \centering
    \begin{subfigure}[t]{0.45\linewidth}
        \centering
        \includegraphics[width=\linewidth]{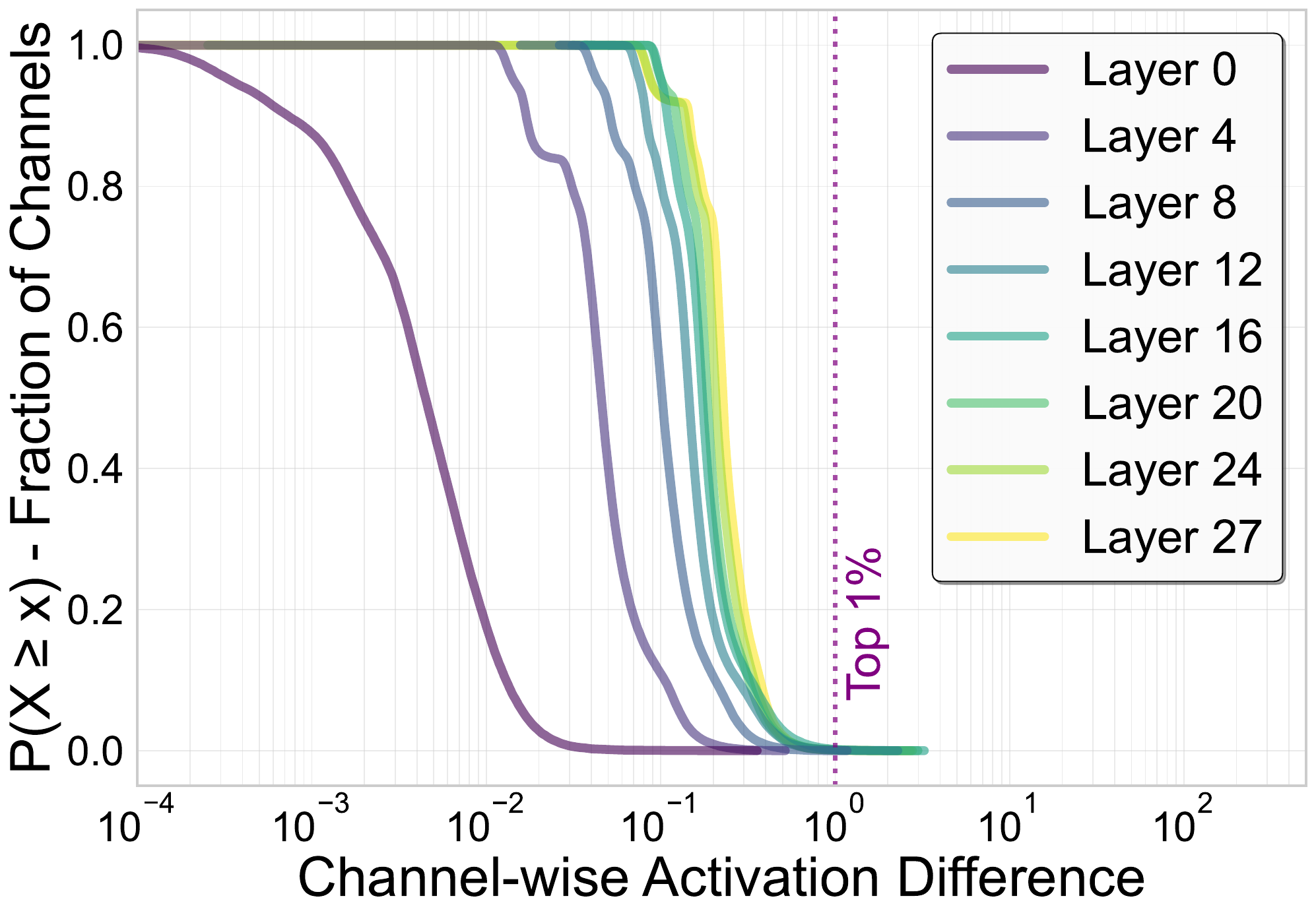}
        \caption{EN-Math (per layer)}
        \label{fig:app_f2_wizardmath_en_math_layer}
    \end{subfigure}
    \hspace{10pt}
    % \hfill
    \begin{subfigure}[t]{0.45\linewidth}
        \centering
        \includegraphics[width=\linewidth]{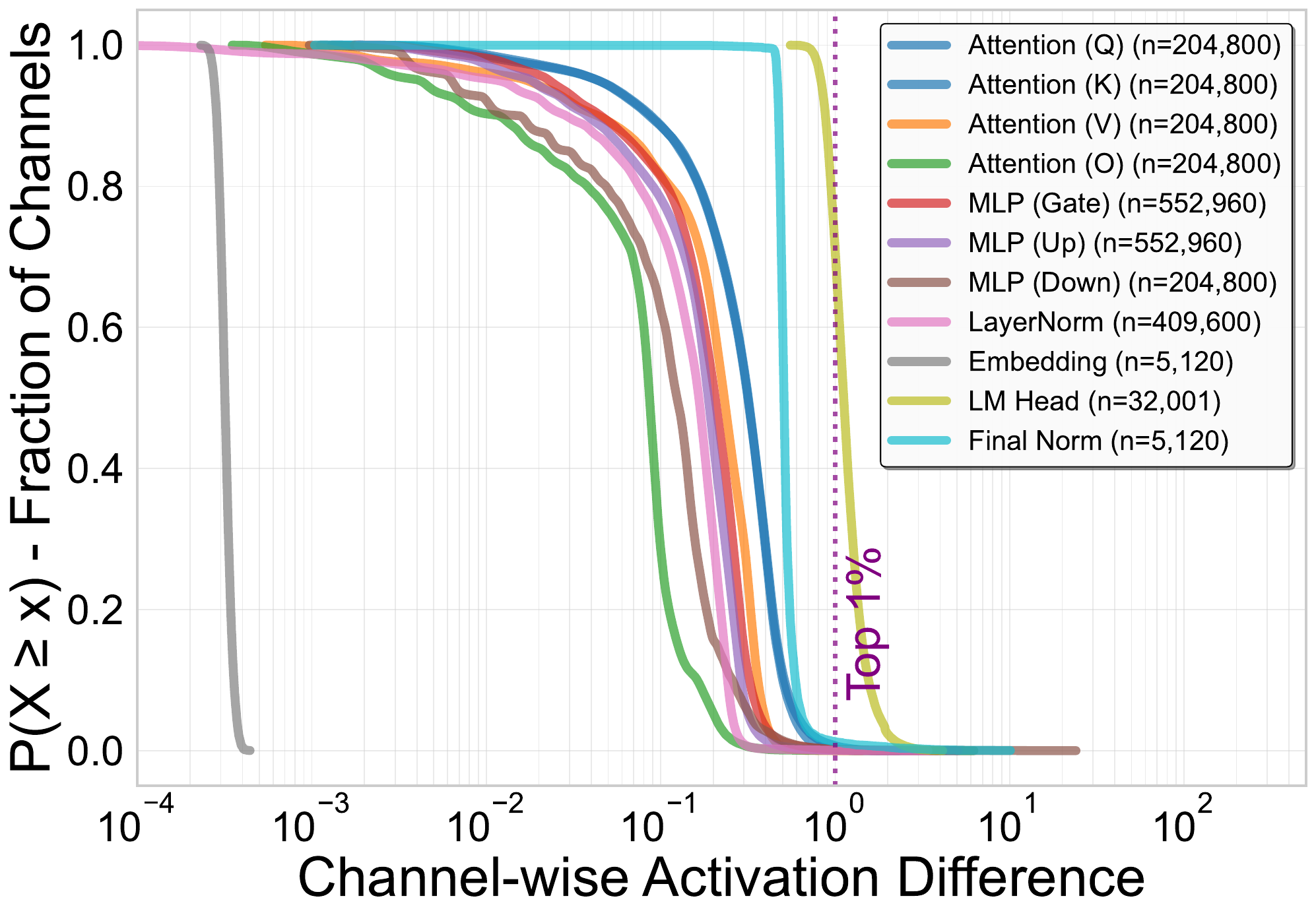}
        \caption{EN-Math (per module)}
        \label{fig:app_f2_wizardmath_en_math_module}
    \end{subfigure}
    \caption{
    Additional results for Finding~2 on \textbf{WizardMath-13B} vs.\ \textbf{LLaMA-2-13B} for English mathematical reasoning.
    }
    \label{fig:app_f2_wizardmath}
\end{figure*}

\begin{figure*}[htbp]
    \centering
    \begin{subfigure}[t]{0.45\linewidth}
        \centering
        \includegraphics[width=\linewidth]{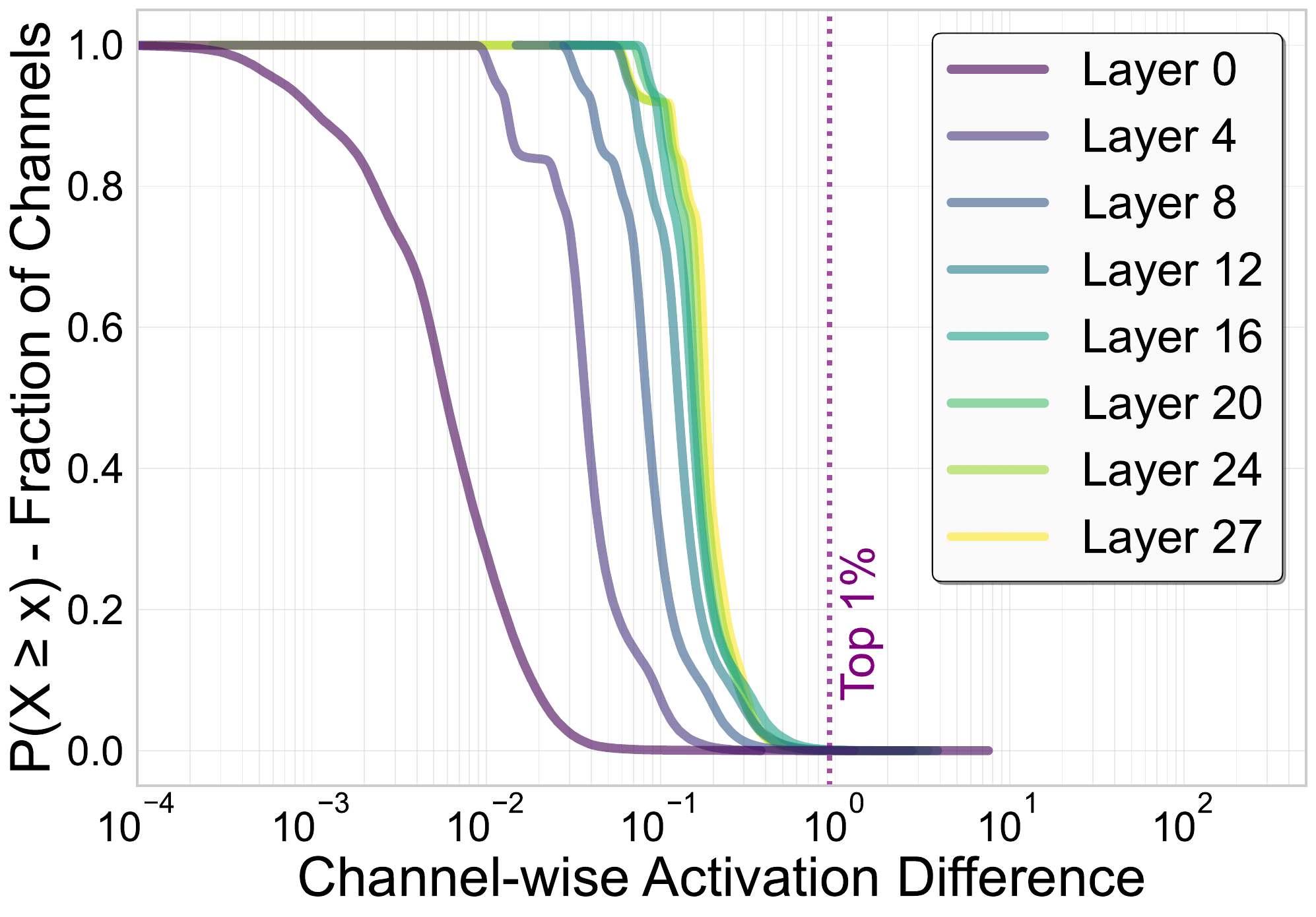}
        \caption{EN-Math (per layer)}
        \label{fig:app_f2_tulu_en_math_layer}
    \end{subfigure}
    \hspace{10pt}
    % \hfill
    \begin{subfigure}[t]{0.45\linewidth}
        \centering
        \includegraphics[width=\linewidth]{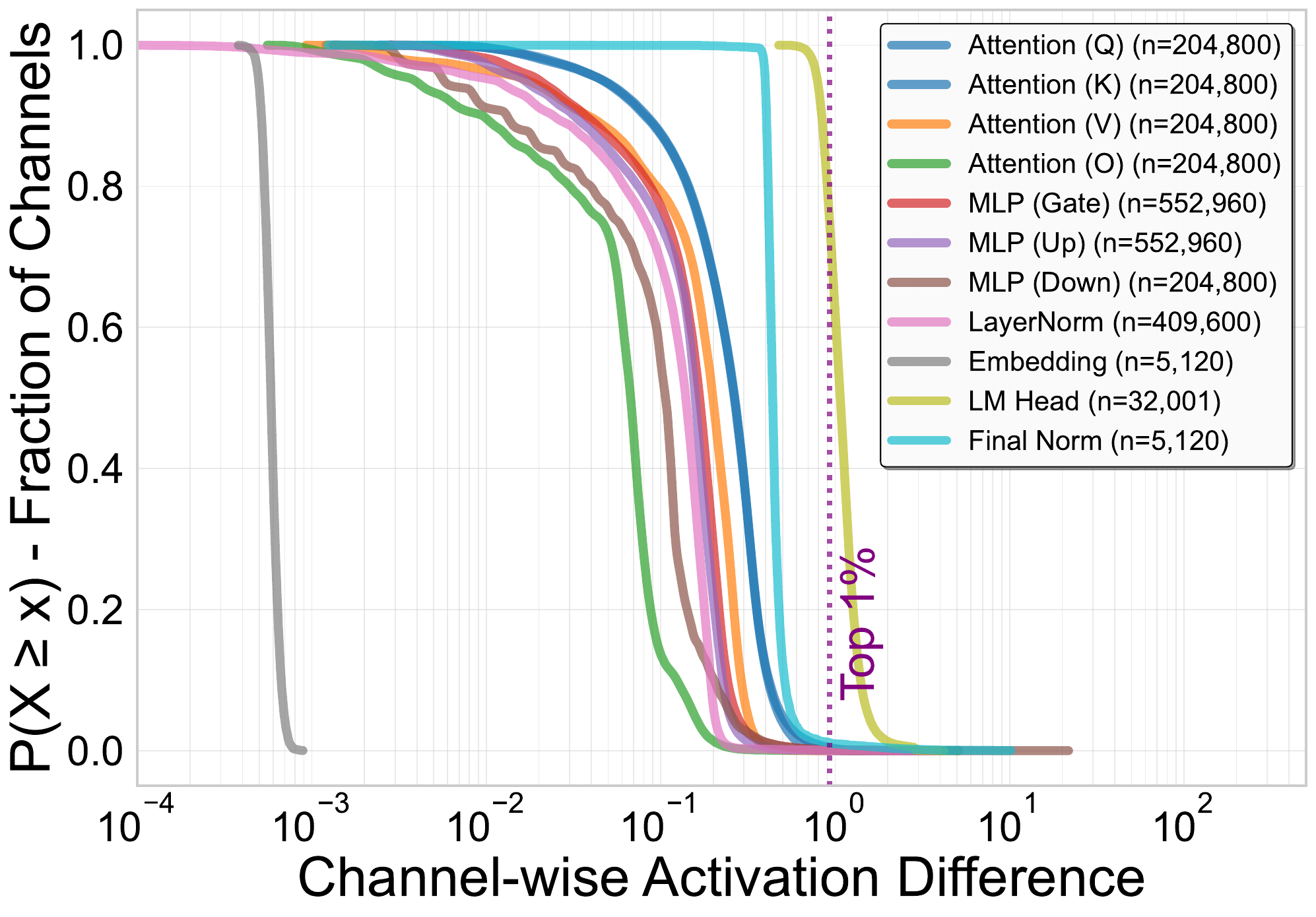}
        \caption{EN-Math (per module)}
        \label{fig:app_f2_tulu_en_math_module}
    \end{subfigure}

    \caption{
    Additional results for Finding~2 on \textbf{Tulu-2-13B} vs.\ \textbf{LLaMA-2-13B} for English mathematical reasoning.
    }
    \label{fig:app_f2_tulu}
\end{figure*}

\begin{figure*}[t]
    \centering

    % ---------- EN-Math ----------
    \begin{subfigure}[t]{0.45\textwidth}
        \centering
        \includegraphics[width=\linewidth]{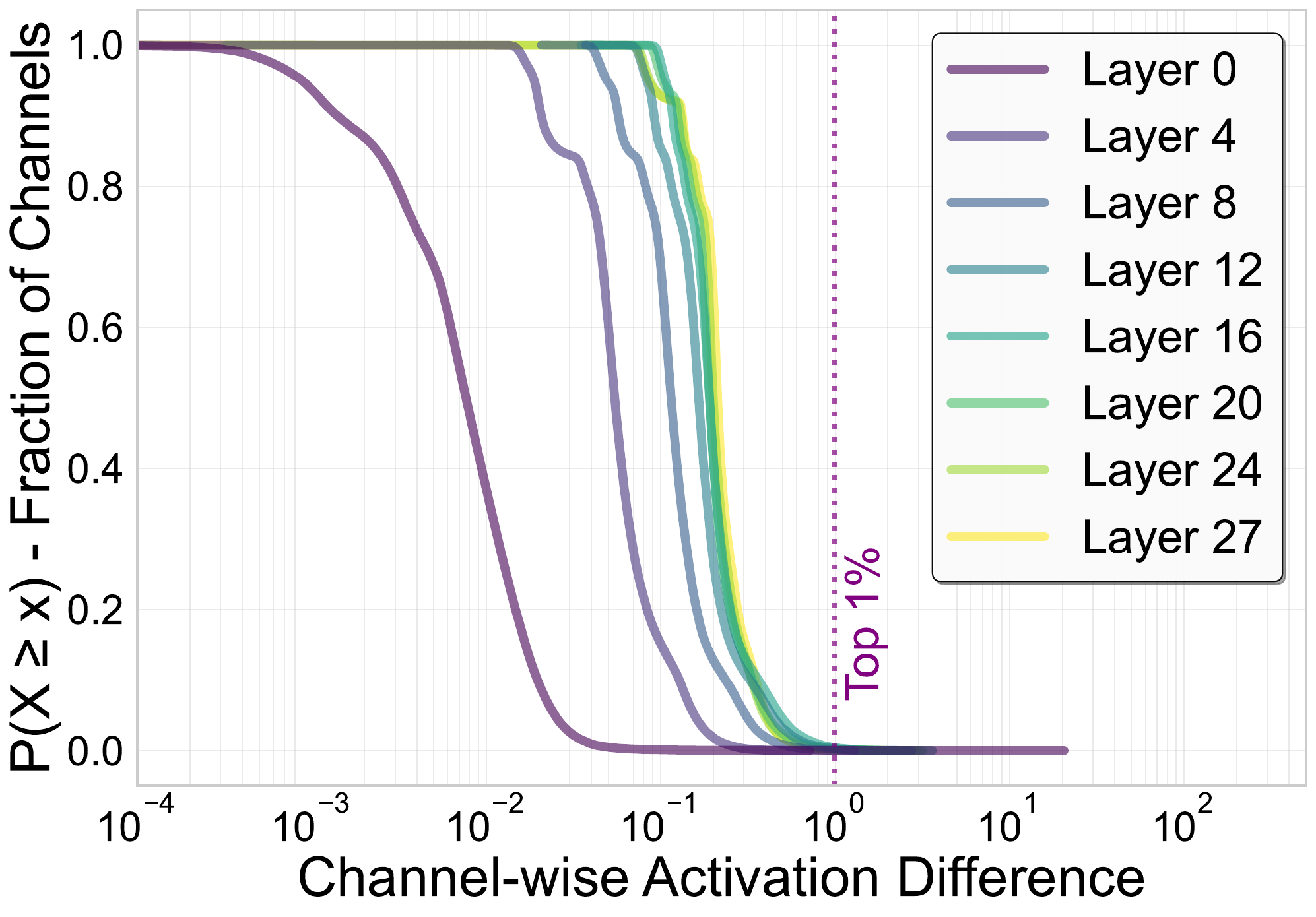}
        \caption{EN-Math (per layer)}
        \label{fig:app_f2_wizardlm_en_math_layer}
    \end{subfigure}
    \hspace{10pt}
    % \hfill
    \begin{subfigure}[t]{0.45\textwidth}
        \centering
        \includegraphics[width=\linewidth]{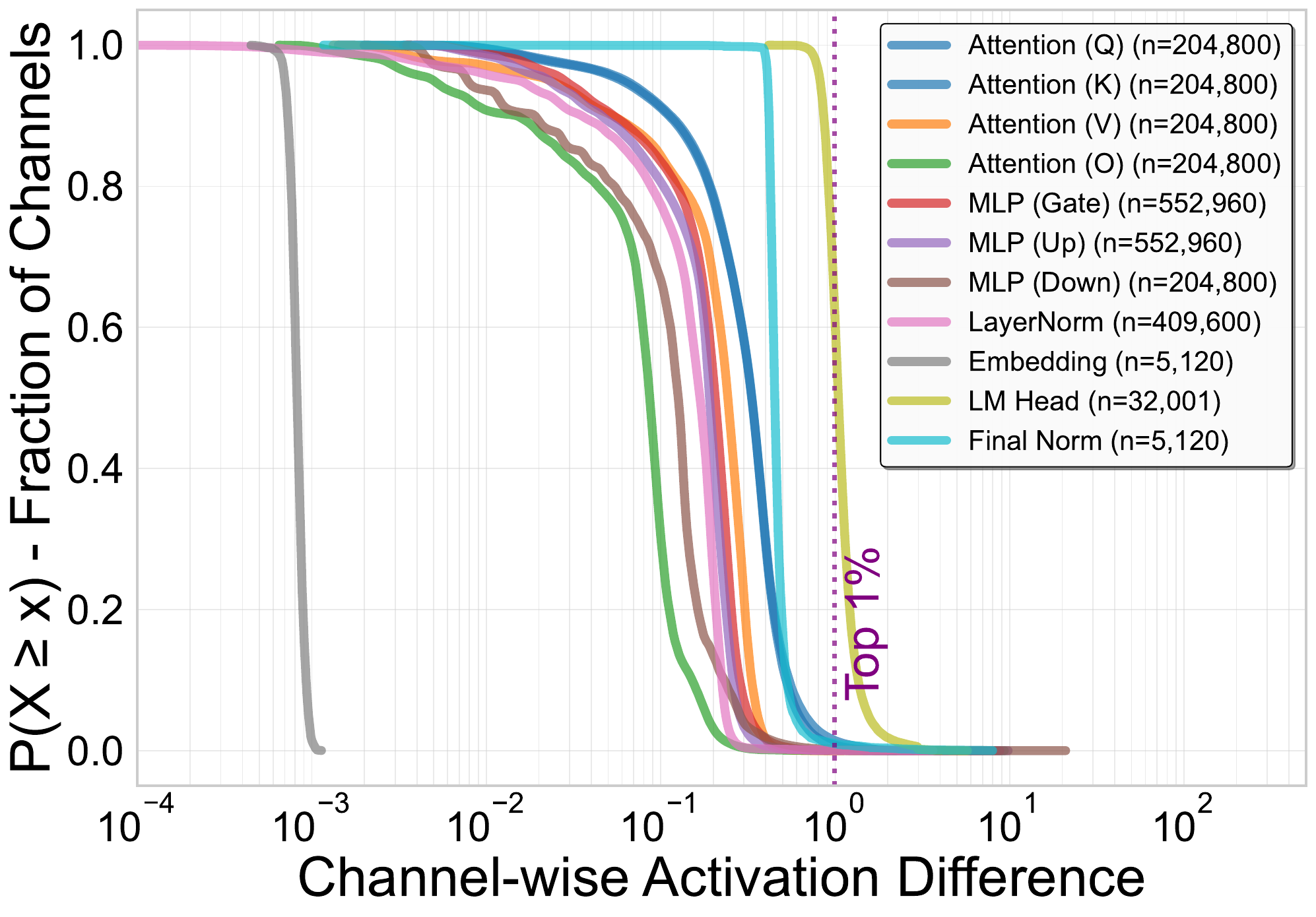}
        \caption{EN-Math (per module)}
        \label{fig:app_f2_wizardlm_en_math_module}
    \end{subfigure}

    \vspace{0.5em}

    % ---------- EN-Science ----------
    \begin{subfigure}[t]{0.45\textwidth}
        \centering
        \includegraphics[width=\linewidth]{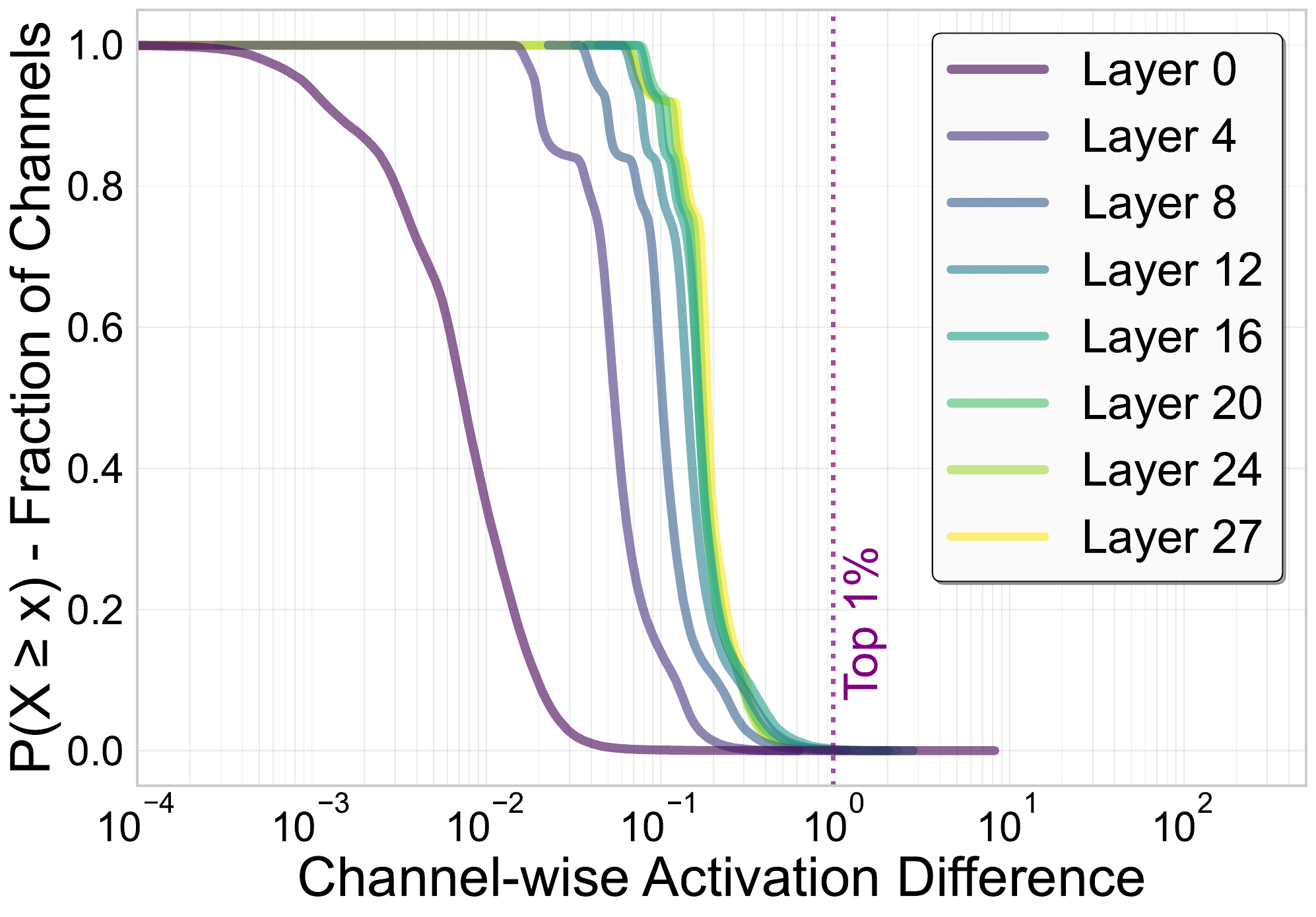}
        \caption{EN-Science (per layer)}
        \label{fig:app_f2_wizardlm_en_sci_layer}
    \end{subfigure}
    \hspace{10pt}
    % \hfill
    \begin{subfigure}[t]{0.45\textwidth}
        \centering
        \includegraphics[width=\linewidth]{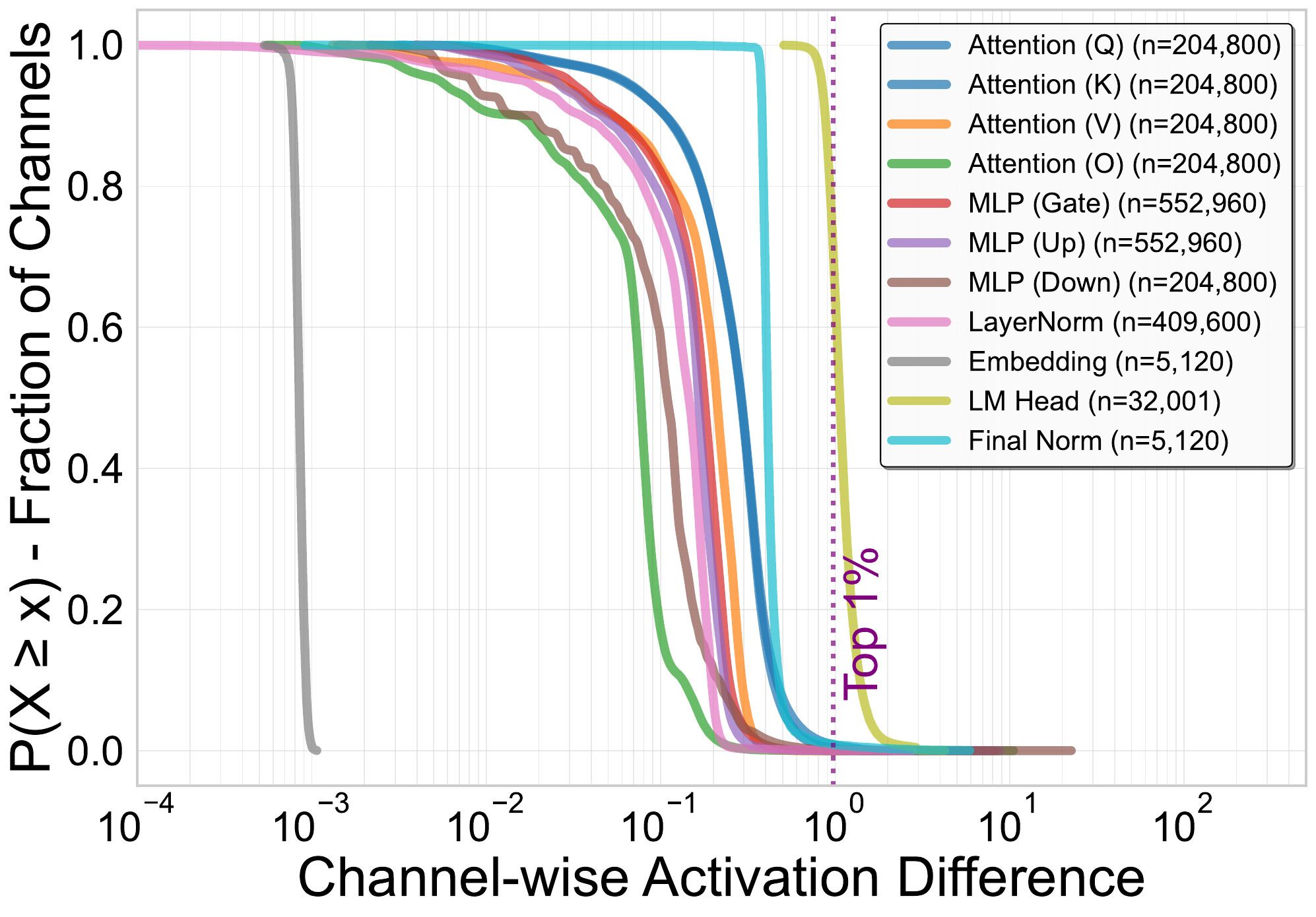}
        \caption{EN-Science (per module)}
        \label{fig:app_f2_wizardlm_en_sci_module}
    \end{subfigure}

    \vspace{0.5em}

    % ---------- EN-Code ----------
    \begin{subfigure}[t]{0.45\textwidth}
        \centering
        \includegraphics[width=\linewidth]{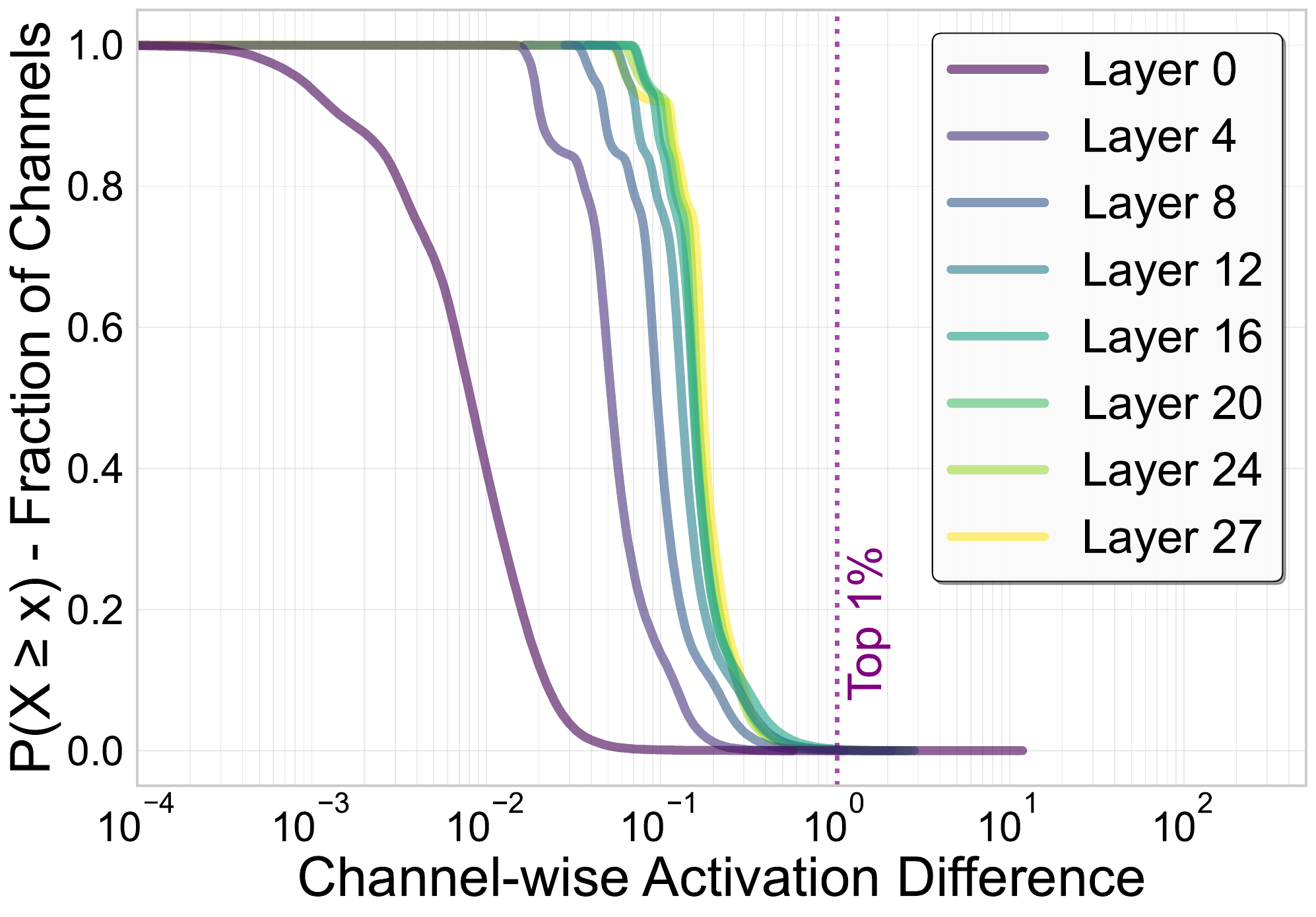}
        \caption{EN-Code (per layer)}
        \label{fig:app_f2_wizardlm_en_code_layer}
    \end{subfigure}
    \hspace{10pt}
    % \hfill
    \begin{subfigure}[t]{0.45\textwidth}
        \centering
        \includegraphics[width=\linewidth]{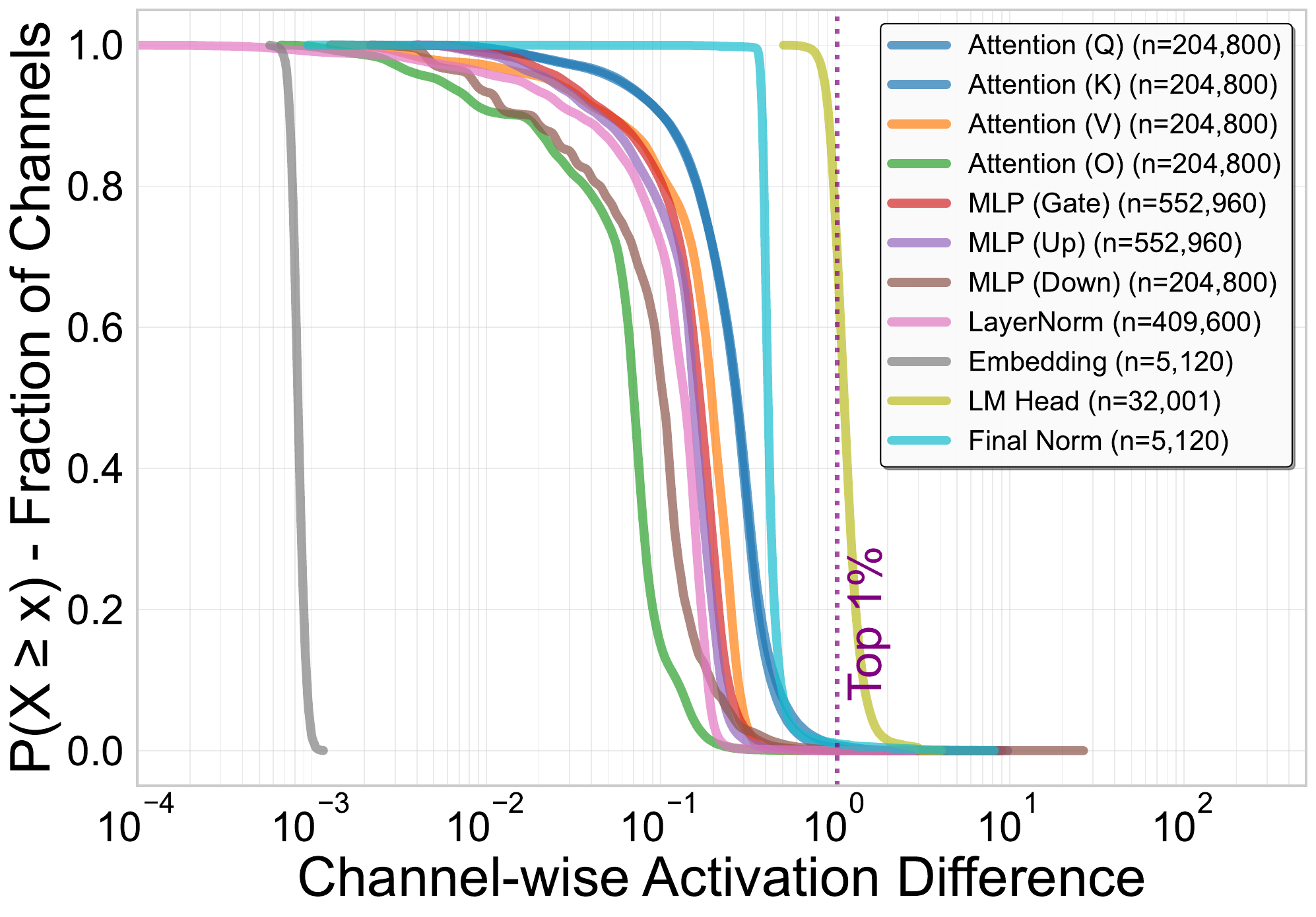}
        \caption{EN-Code (per module)}
        \label{fig:app_f2_wizardlm_en_code_module}
    \end{subfigure}

    \caption{
    Additional results for Finding~2 on \textbf{WizardLM-13B} vs.\ \textbf{LLaMA-2-13B}.
    Each row corresponds to an English ability (math, science, code), and columns show per-layer (left) and per-module (right) CCDFs of channel-wise activation differences.
    }
    \label{fig:app_f2_wizardlm}
\end{figure*}

% \FloatBarrier

\paragraph{Qwen2.5 Coder Models.}
We next consider code-specialized Qwen2.5 models.
Figures~\ref{fig:app_f2_qwen_coder_7b} and~\ref{fig:app_f2_qwen_coder_15b} show layer- and module-wise activation difference distributions for English-Math on the Coder-Instruct vs. Instruct model pairs of 7B and 1.5B, respectively.
Despite being fine-tuned for code generation, the activation difference distributions for mathematical reasoning remain heavy-tailed and broadly distributed across the network.
This suggests that the structural properties identified in Finding~2 generalize beyond math-specialized models and hold across different fine-tuning objectives.

\paragraph{LLaMA-2 Family Models.}
To assess whether Finding~2 generalizes beyond the Qwen family, we further analyze models derived from LLaMA-2-13B.
Specifically, we compare the base model with WizardMath-13B, Tulu-2-13B, and WizardLM-13B, which are specialized for mathematics, scientific reasoning, and code generation, respectively.
Figure~\ref{fig:app_f2_wizardmath} and Figure~\ref{fig:app_f2_tulu} report layer- and module-wise activation difference distributions for English-Math.
Figure~\ref{fig:app_f2_wizardlm} additionally presents results for English-Math, English-Science, and English-Code.
Across all comparisons, activation differences again exhibit consistent heavy-tailed patterns distributed across layers and modules, with mild concentration in later layers and output-related components.
These observations confirm that the structural conclusions of Finding~2 extend to the LLaMA-2 family and across distinct specialization objectives.

\paragraph{Summary.}
Across all examined model families, scales, abilities, and languages, layer- and module-wise activation difference distributions consistently exhibit heavy-tailed behavior with sparse large deviations spread throughout the network.
These additional results further support Finding~2 and motivate the use of channel-wise activation signals for selective and controlled ability transfer.

\subsection{Additional Results for Finding~3}
\label{app:finding3}

In the main text (Finding~3), we show that, within a single model pair, channels associated with large activation differences are not same across language-domain abilities.
Instead, these channels exhibit structured overlap patterns:
overlap across languages is generally low, while overlap between different domains within the same language is higher but still incomplete.
In this appendix, we provide additional results that validate this observation across different model families (Qwen2.5 and LLaMA-2),
model scales (1.5B/7B/13B), and ability domains (math, science, and code in 11 languages), \emph{while restricting analysis to a single model pair at a time}.

\begin{figure*}[t]
    \centering

    % ---------- Qwen2.5 Math 7B ----------
    \begin{subfigure}[t]{0.47\textwidth}
        \centering
        \includegraphics[width=\linewidth]{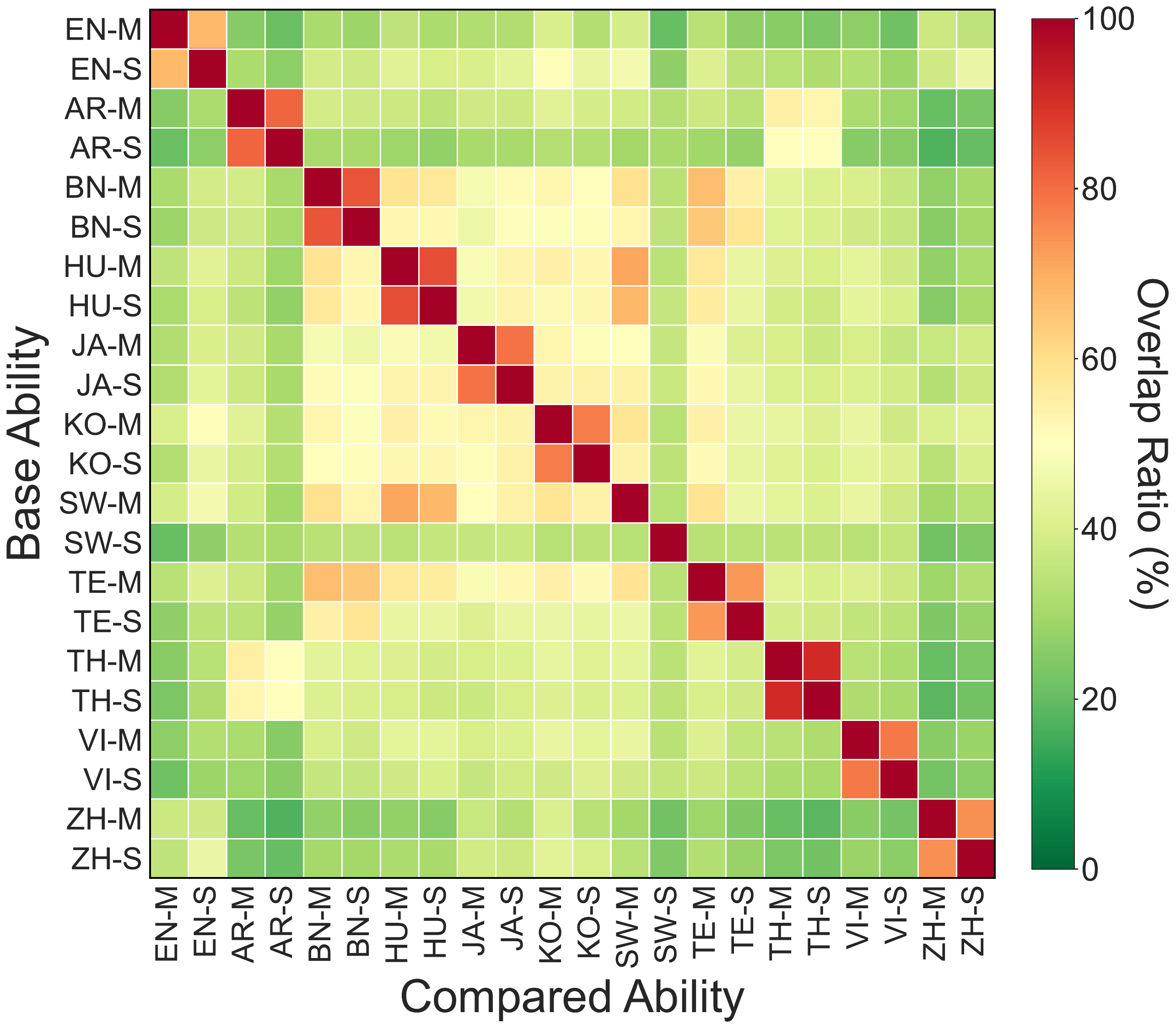}
        \caption{Qwen2.5-Math-7B-Instruct vs.\ Qwen2.5-7B-Instruct}
        \label{fig:app_f3_qwen_math_7b}
    \end{subfigure}
    \hspace{10pt}
    % \hfill
    % ---------- Qwen2.5 Math 1.5B ----------
    \begin{subfigure}[t]{0.47\textwidth}
        \centering
        \includegraphics[width=\linewidth]{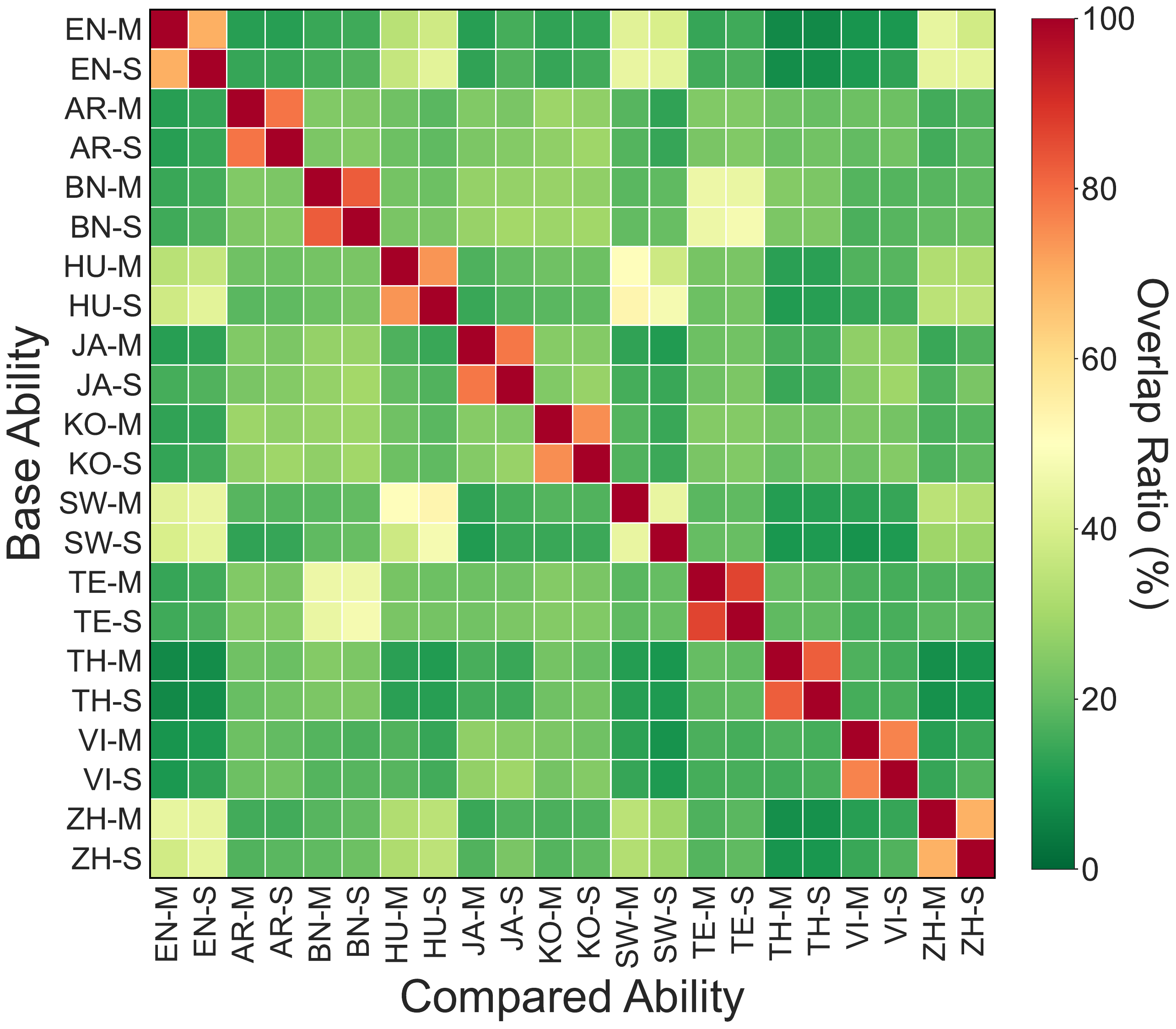}
        \caption{Qwen2.5-Math-1.5B-Instruct vs.\ Qwen2.5-1.5B-Instruct}
        \label{fig:app_f3_qwen_math_15b}
    \end{subfigure}

    \vspace{0.6em}

    % ---------- Qwen2.5 Coder 7B ----------
    \begin{subfigure}[t]{0.47\textwidth}
        \centering
        \includegraphics[width=\linewidth]{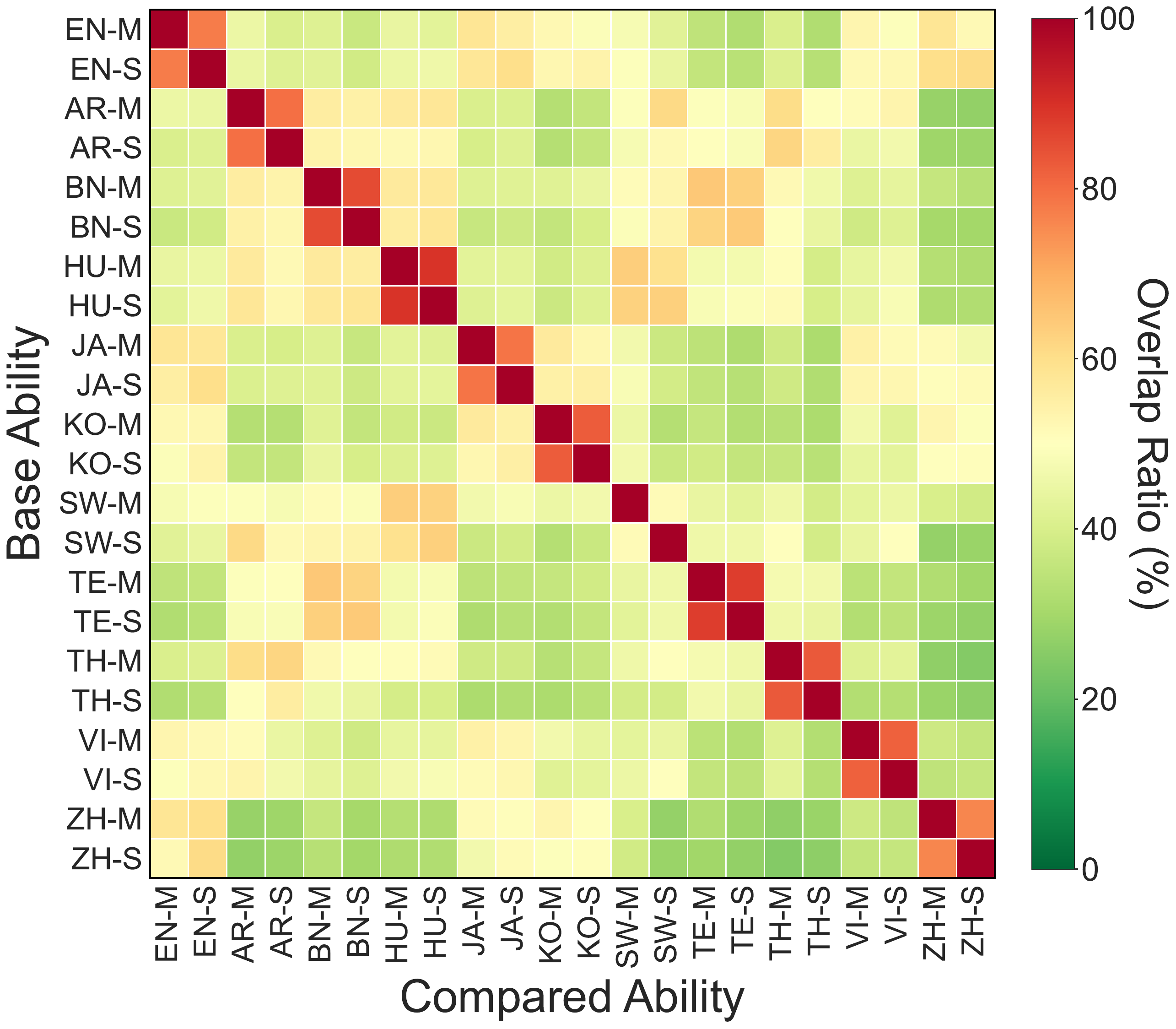}
        \caption{Qwen2.5-Coder-7B-Instruct vs.\ Qwen2.5-7B-Instruct}
        \label{fig:app_f3_qwen_coder_7b}
    \end{subfigure}
    \hspace{10pt}
    % \hfill
    % ---------- Qwen2.5 Coder 1.5B ----------
    \begin{subfigure}[t]{0.47\textwidth}
        \centering
        \includegraphics[width=\linewidth]{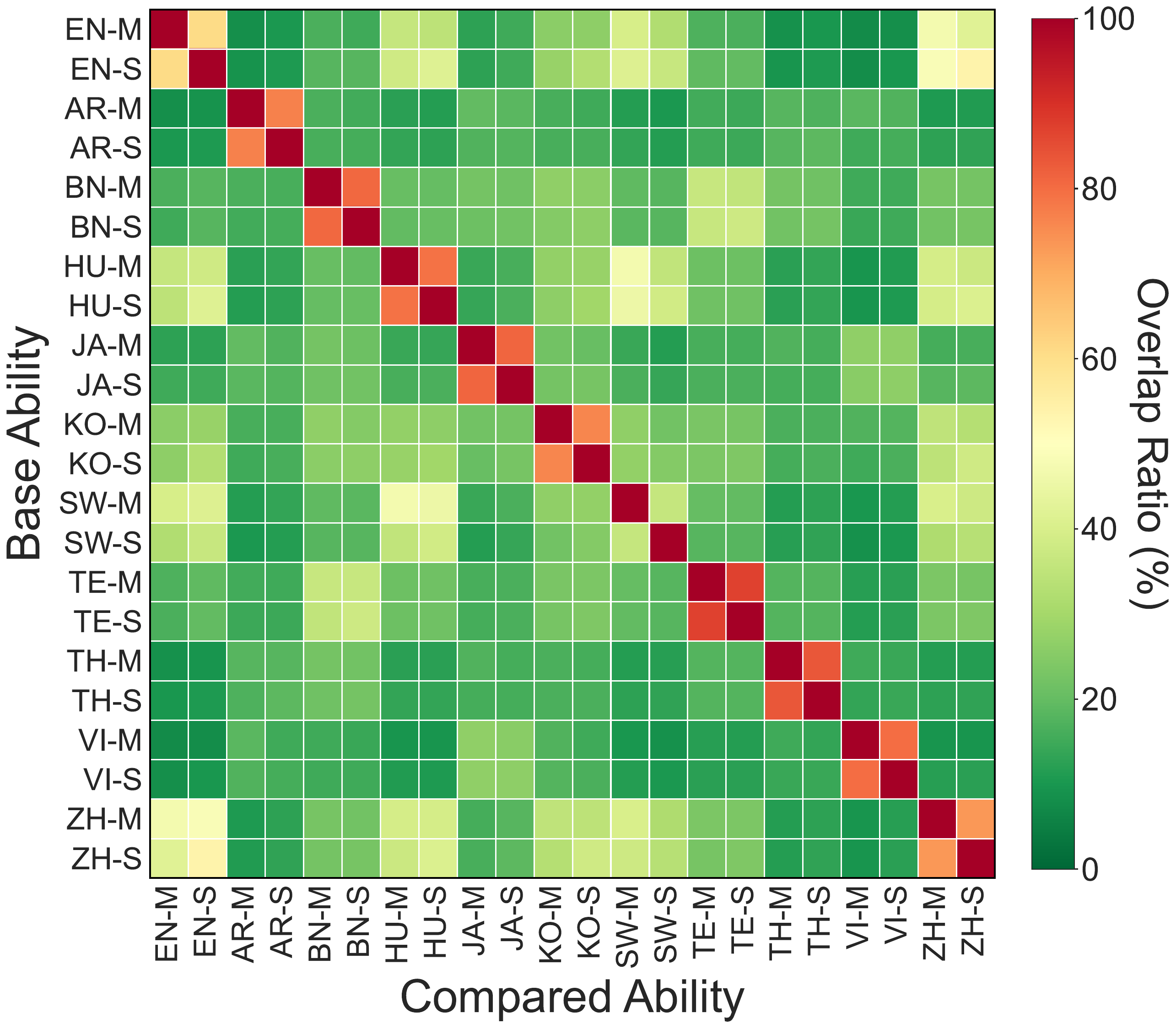}
        \caption{Qwen2.5-Coder-1.5B-Instruct vs.\ Qwen2.5-1.5B-Instruct}
        \label{fig:app_f3_qwen_coder_15b}
    \end{subfigure}

    \caption{
    \textbf{Additional results for Finding~3: overlap of activation-based ability-specific channel masks within model pairs.}
    Each heatmap shows the overlap (\%) between top-$1\%$ activation-difference channel masks across
    $11$ languages and $2$ domains (Math and Science), computed within a fixed model pair.
    Diagonal blocks correspond to same-language comparisons, while off-diagonal blocks reflect cross-language overlap.
    Across all settings, overlaps are generally low across languages, whereas overlaps between Math and Science
    within the same language are consistently higher, indicating partial disentanglement of ability-specific channels
    within a model pair.
    }
    \label{fig:app_f3_qwen_overlap}
\end{figure*}

\begin{table*}[htbp]
\centering
\footnotesize
\setlength{\tabcolsep}{6pt}
\renewcommand{\arraystretch}{1.15}

\begin{subtable}[t]{0.32\textwidth}
\centering
\begin{tabular}{lccc}
\toprule
 & Math & Science & Code \\
\midrule
Math    & 100.0 & 89.1 & 87.5 \\
Science & 89.1  & 100.0 & 89.4 \\
Code    & 87.5  & 89.4 & 100.0 \\
\bottomrule
\end{tabular}
\caption{Tulu-2-13B vs.\ LLaMA-2-13B}
\end{subtable}
\hfill
\begin{subtable}[t]{0.32\textwidth}
\centering
\begin{tabular}{lccc}
\toprule
 & Math & Science & Code \\
\midrule
Math    & 100.0 & 82.4 & 81.8 \\
Science & 82.4  & 100.0 & 86.0 \\
Code    & 81.8  & 86.0 & 100.0 \\
\bottomrule
\end{tabular}
\caption{WizardLM-13B vs.\ LLaMA-2-13B}
\end{subtable}
\hfill
\begin{subtable}[t]{0.32\textwidth}
\centering
\begin{tabular}{lccc}
\toprule
 & Math & Science & Code \\
\midrule
Math    & 100.0 & 89.4 & 85.9 \\
Science & 89.4  & 100.0 & 89.2 \\
Code    & 85.9  & 89.2 & 100.0 \\
\bottomrule
\end{tabular}
\caption{WizardMath-13B vs.\ LLaMA-2-13B}
\end{subtable}

\caption{
Top-1\% ability-mask overlap (\%) within LLaMA-2-13B derived model pairs.
Each entry reports the fraction of channels shared between the base (row) and compared (column) ability masks.
Compared to Qwen2.5 models, overlaps across EN-Math, EN-Science, and EN-Code are generally higher in LLaMA-2, suggesting stronger sharing of large-activation-difference channels within this model family.
This effect is plausibly influenced by the increased concentration of top-1\% channels in shared components such as the LM head, rather than indicating a lack of ability-specific structure.
}

\label{tab:app_f3_llama2_overlap_top1pct}

\end{table*}

\paragraph{Qwen2.5 Models across Languages and Domains.}
We first analyze ability disentanglement within Qwen2.5 model pairs.
For both math-specialized and code-specialized variants, we consider Instruct vs.\ ability-specialized models at the 7B and 1.5B scales. 
For each model pair, we construct activation-based ability-specific channel masks by selecting the top-$1\%$ channels ranked by token-averaged activation differences for each language-domain combination (math or science across 11 languages).
We then compute pairwise overlap ratios between all ability masks, normalized by the size of the base (row) mask.
Figures~\ref{fig:app_f3_qwen_math_7b},
\ref{fig:app_f3_qwen_math_15b},
\ref{fig:app_f3_qwen_coder_7b},
and~\ref{fig:app_f3_qwen_coder_15b}
show the resulting overlap heatmaps.
Across all Qwen2.5 settings, we observe consistent structure.
Overlaps between different languages are generally low, even when the domain is the same (e.g., EN-Math vs.\ AR-Math), indicating strong language-conditioned separation.
In contrast, overlaps between different domains within the same language (e.g., EN-Math vs.\ EN-Science) are noticeably higher, though still far from complete.
This suggests partial channel reuse across domains within a language, alongside substantial domain-specific components.
These patterns are stable across model scales and across math- and code-specialized fine-tuning objectives, demonstrating that ability disentanglement within a model pair is a robust phenomenon.

\paragraph{LLaMA-2 Family Models.}
We next examine ability disentanglement within model pairs derived from \texttt{LLaMA-2-13B}.
Due to limited multilingual coverage, this analysis focuses on English-language abilities.
We consider three model pairs: WizardMath-13B vs.\ LLaMA-2-13B, Tulu-2-13B vs.\ LLaMA-2-13B, and WizardLM-13B vs.\ LLaMA-2-13B, which specialize in math, science, and code reasoning, respectively.
For each pair, we construct activation-based ability-specific masks for English math, science, and code, and compute pairwise overlap ratios between masks.
Table~\ref{tab:app_f3_llama2_overlap_top1pct} reports the resulting $3 \times 3$ overlap matrices.
Compared to Qwen2.5 models, overlaps between different abilities are generally higher in the LLaMA-2 family, indicating stronger sharing of channels with large activation differences across English-language domains.
Nevertheless, overlaps remain structured, with ability-specific masks still exhibiting non-trivial differences across math, science, and code.
This suggests that while the degree of disentanglement depends on architectural and training factors, ability-related activation differences within a single model pair are not fully shared even in LLaMA-2 models.

\paragraph{Summary.}
Across both Qwen2.5 and LLaMA-2 model families, channels exhibiting large activation differences are not uniformly shared across language-domain abilities within a single model pair.
While the overall degree of overlap varies with model architecture and mask composition, the observed overlap patterns remain structured, indicating partial but non-trivial ability separation.
These results support Finding~3 and suggest that activation-based ability disentanglement is a general phenomenon, modulated by architectural and training-specific biases.

\subsection{Additional Results for Finding~4}
\label{app:finding4}

This appendix provides additional evidence supporting Finding~4 by extending the analysis of cross-pair consistency of ability-specific channels to smaller model scales and to more models in LLaMA-2 family.
Across all experiments, consistency is measured as the overlap between top-1\% activation-based channel masks corresponding to the same language-domain ability, extracted from different fine-tuning pairs that share a common pretrained base model.
These results further examine the extent to which ability-relevant channel structure is preserved after fine-tuning.

\begin{figure*}[t]
    \centering
    \begin{subfigure}[t]{0.40\textwidth}
        \centering
        \includegraphics[width=\linewidth]{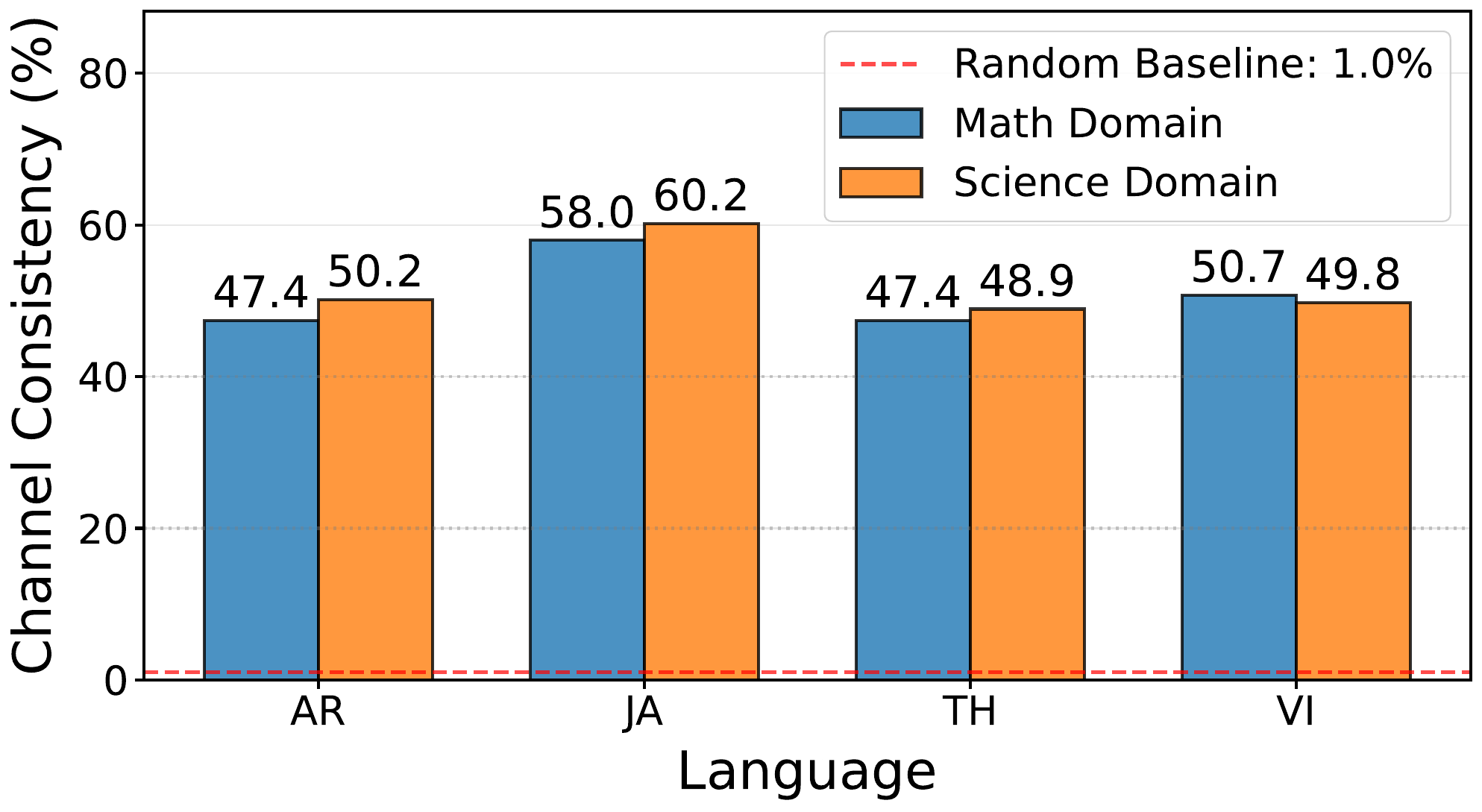}
        \caption{}
        \label{fig:app_f4_qwen15b_filtered}
    \end{subfigure}
    \hfill
    \begin{subfigure}[t]{0.29\textwidth}
        \centering
        \includegraphics[width=\linewidth]{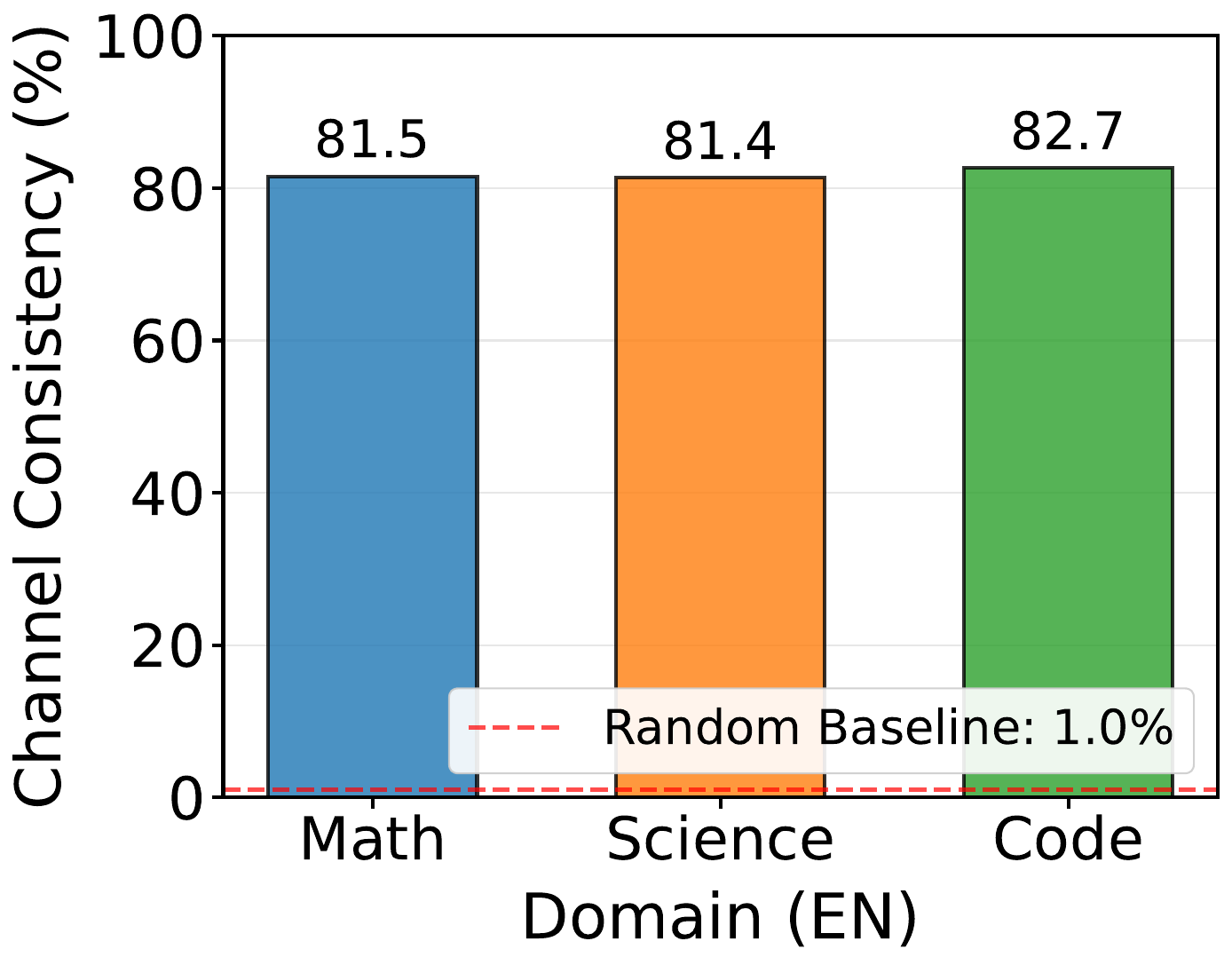}
        \caption{}
        \label{fig:app_f4_llama2_tulu_wizardmath}
    \end{subfigure}
    \hfill
    \begin{subfigure}[t]{0.29\textwidth}
        \centering
        \includegraphics[width=\linewidth]{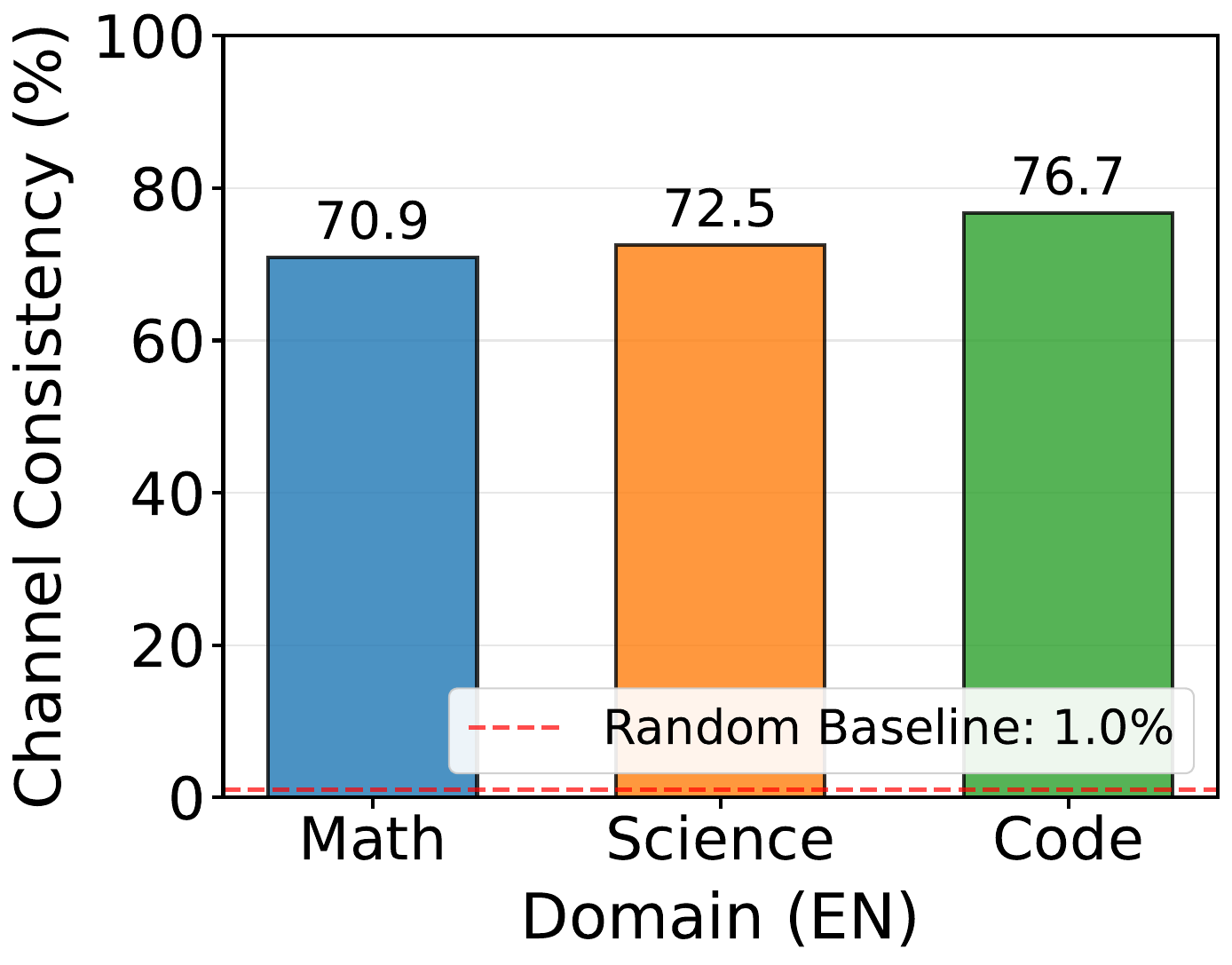}
        \caption{}
        \label{fig:app_f4_llama2_wizardlm_wizardmath}
    \end{subfigure}

    % \vspace{-15pt}
    \caption{
    \textbf{Additional cross-pair consistency results for Finding~4.}
    \textbf{Left:} Qwen2.5-1.5B cross-pair overlap between top-1\% activation-based masks from (Instruct vs.\ Math-Instruct) and (Instruct vs.\ Coder-Instruct), shown for a subset of representative languages; bars report math and science results separately.
    \textbf{Middle, Right:} LLaMA-2-13B cross-pair overlap between ability-specific channel masks extracted from different fine-tuned vs.\ base model pairs, between (Tulu-2-DPO vs.\ LLaMA-2) and (WizardMath vs.\ LLaMA-2), and between (WizardLM vs.\ LLaMA-2) and (WizardMath vs.\ LLaMA-2), evaluated on English math, science, and code abilities.
    Overall, these results extend the main-text findings by showing that cross-pair consistency persists at smaller model scales and remains uniformly high within the LLaMA-2 family, while Qwen models exhibit mild language-dependent variation.
    }
    \label{fig:app_f4_additional}
\end{figure*}

\paragraph{Qwen2.5 Models at Smaller Scale.}
Figure~\ref{fig:app_f4_qwen15b_filtered} reports cross-pair consistency results on Qwen2.5-1.5B.
We compare ability-specific channel masks extracted from the Instruct vs.\ Math-Instruct and Instruct vs.\ Coder-Instruct model pairs.
For each selected language, overlap ratios are shown separately for math and science abilities.
Consistent with the 7B-scale results in the main text, we observe that ability-specific channel masks exhibit substantial overlap across fine-tuning objectives, remaining well above the random baseline.
Although the magnitude of overlap varies across languages, the presence of non-trivial cross-pair consistency at this smaller scale indicates that fine-tuning largely reuses a shared set of ability-relevant channels rather than inducing entirely new ones.
This suggests that the core channel structure associated with each ability is preserved under fine-tuning, even when model capacity is reduced.

\paragraph{LLaMA-2 Family Models.}
We further examine cross-pair consistency within the LLaMA-2-13B family by complementing the results presented in the main text.
For each fine-tuned model (WizardMath-13B, WizardLM-13B, and Tulu-2-DPO-13B), ability-specific channel masks are extracted by comparing its activations against the shared base model LLaMA-2-13B.
Each mask therefore captures activation differences induced by a specific fine-tuning objective relative to the same pretrained representation.
While the main text reports cross-pair consistency between the Tulu-2-DPO-13B and WizardLM-13B model pairs, the appendix focuses on the remaining comparisons involving WizardMath-13B.
Specifically, we report overlap between ability-specific masks derived from
(i) Tulu-2-DPO-13B vs.\ LLaMA-2-13B and WizardMath-13B vs.\ LLaMA-2-13B, and
(ii) WizardLM-13B vs.\ LLaMA-2-13B and WizardMath-13B vs.\ LLaMA-2-13B.
Due to limited multilingual coverage, this analysis is restricted to English math, science, and code abilities.
As shown in Figures~\ref{fig:app_f4_llama2_tulu_wizardmath} and \ref{fig:app_f4_llama2_wizardlm_wizardmath}, both comparisons exhibit consistently high overlap across all abilities.
Together with the results in the main text, these findings indicate that, within the LLaMA-2 family, channels associated with large activation differences for a given ability are highly consistent across different fine-tuning objectives.
This further supports the stability of ability-specific channel structure after fine-tuning.

\paragraph{Summary.}
Across both Qwen2.5 and LLaMA-2 families, ability-specific channels associated with large activation differences exhibit obvious cross-pair consistency after fine-tuning.
In Qwen2.5 models, ability-relevant channel structure is preserved across fine-tuning objectives and model scales, despite moderate variation in overlap magnitude across languages.
Within the LLaMA-2 family, where the analysis is restricted to English abilities, consistently high overlap is observed across all reported model pairs.
Taken together, these results support Finding~4 by showing that fine-tuning predominantly reuses and reweights a stable set of ability-relevant channels anchored in the pretrained representation, rather than inducing fundamentally new channel structures.

\begin{figure*}[htbp]
  \centering

  % -------- (a) Global --------
  \begin{subfigure}[t]{0.32\linewidth}
    \centering
    \includegraphics[width=\linewidth]{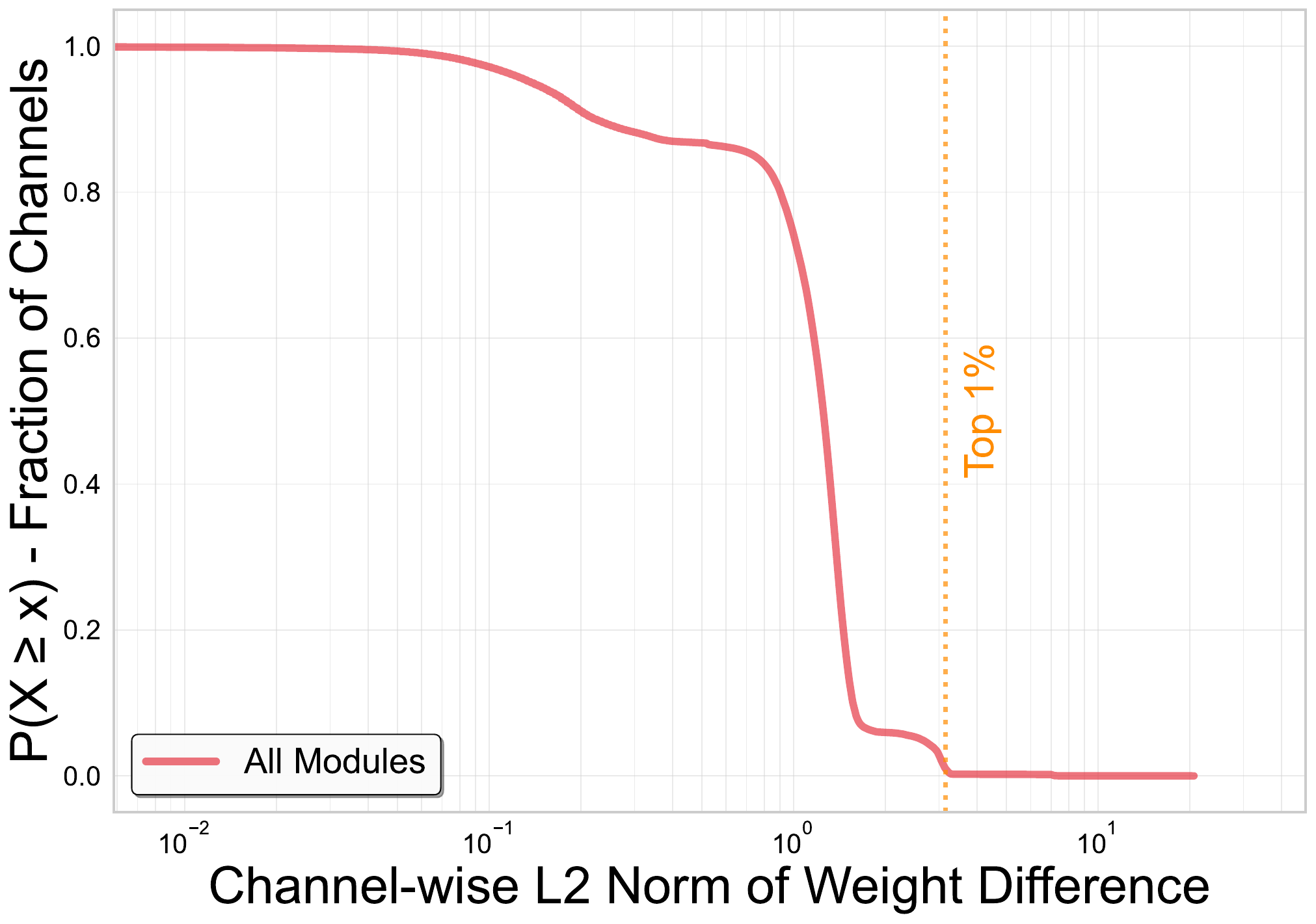}
    \caption{Global CCDF}
    \label{fig:weight_ccdf_global}
  \end{subfigure}
  \hfill
  % -------- (b) Layer-wise --------
  \begin{subfigure}[t]{0.32\linewidth}
    \centering
    \includegraphics[width=\linewidth]{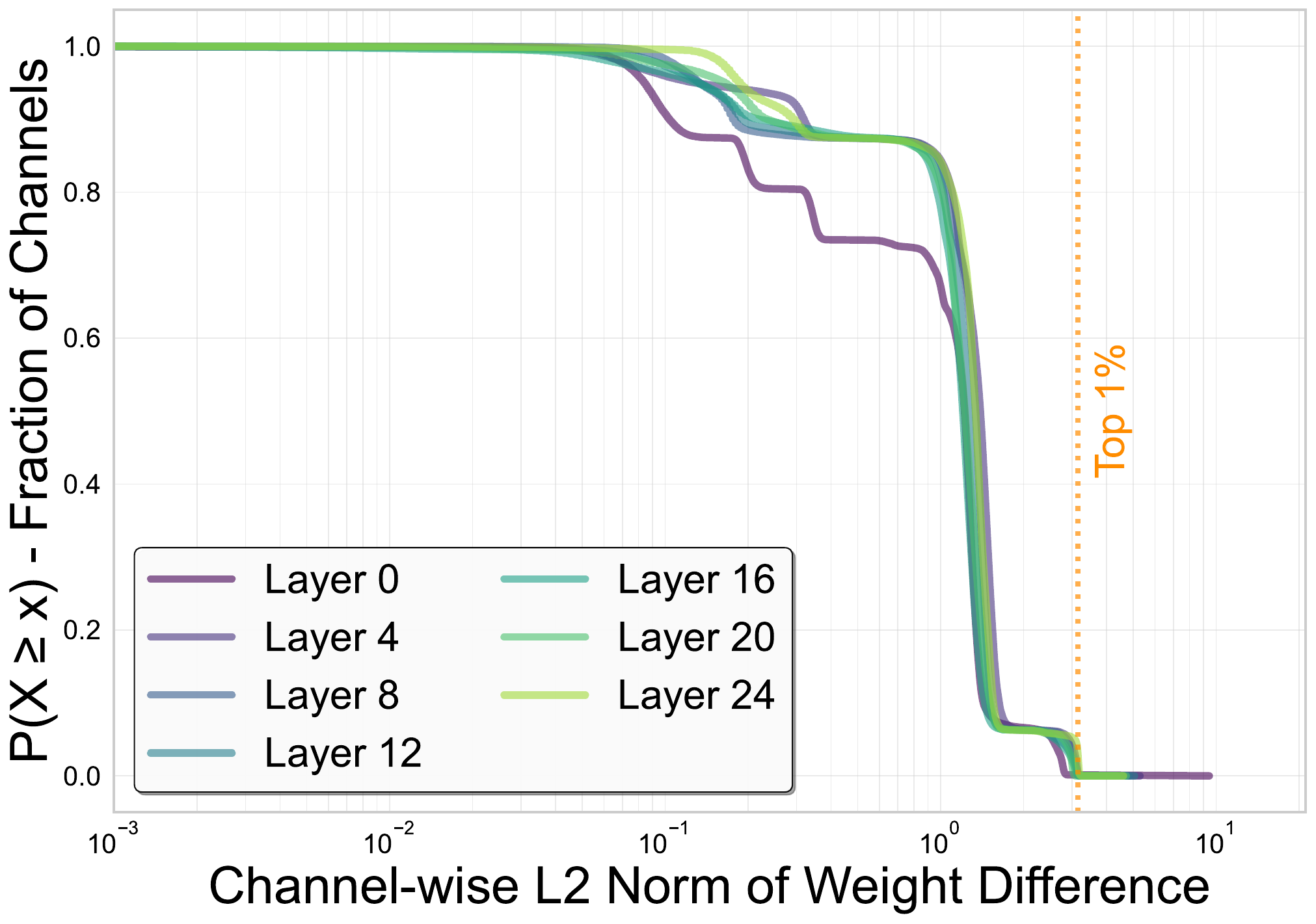}
    \caption{CCDF per layer)}
    \label{fig:weight_ccdf_by_layer}
  \end{subfigure}
  \hfill
  % -------- (c) Module-wise --------
  \begin{subfigure}[t]{0.32\linewidth}
    \centering
    \includegraphics[width=\linewidth]{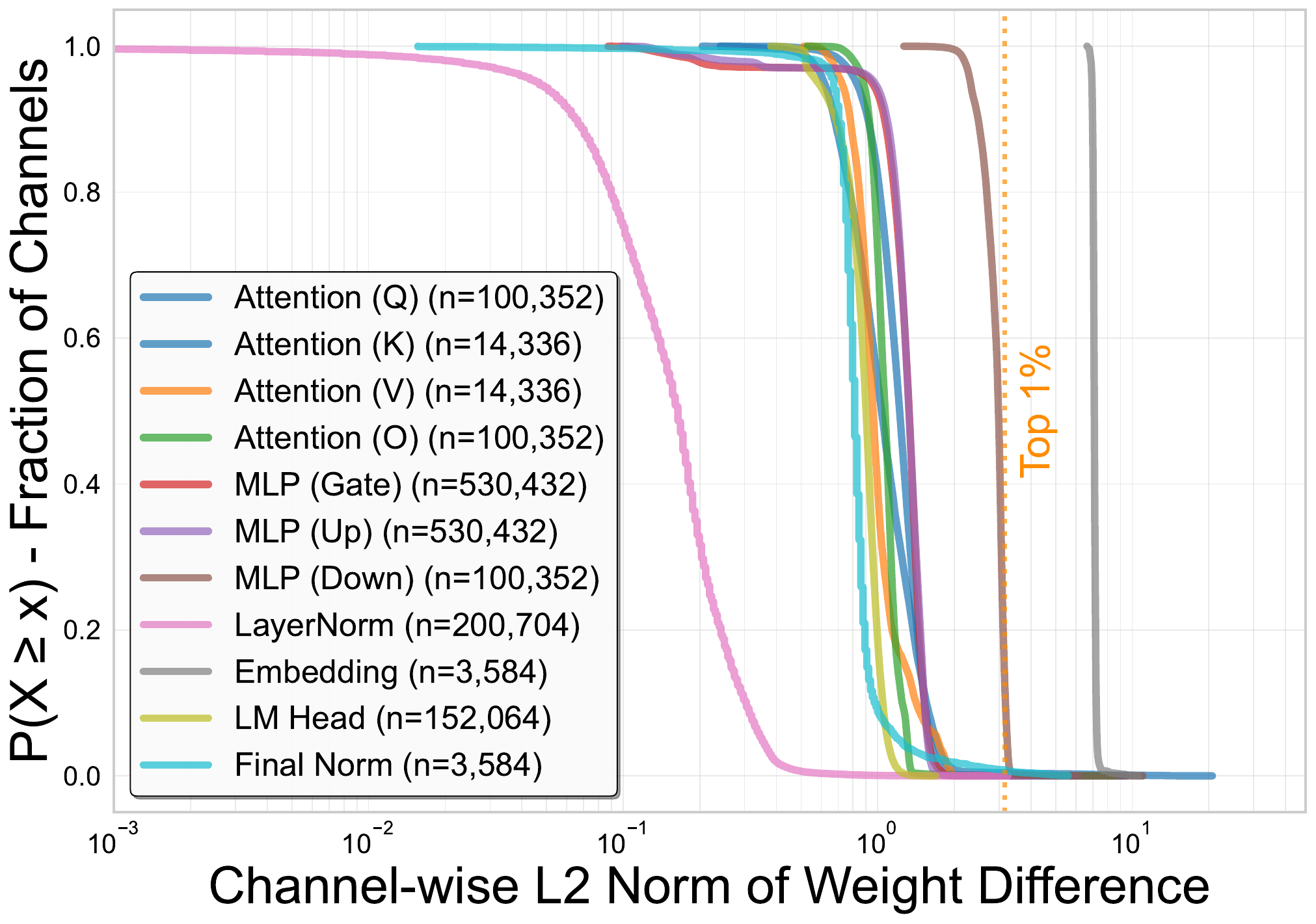}
    \caption{CCDF per module}
    \label{fig:weight_ccdf_by_module}
  \end{subfigure}

\caption{
CCDFs of channel-wise $\ell_2$ weight differences on \textbf{Qwen2.5-Math-7B-Instruct} vs. \textbf{Qwen2.5-7B-Instruct}, shown globally, per layer, and per module. Weight updates exhibit moderately heavy-tailed distributions, though less extreme than activation differences.
}
  \label{fig:weight_ccdf_by_structure}
\end{figure*}

% \begin{figure}[htbp]
%     \centering
%     \includegraphics[width=\linewidth]{latex/images/Activation_vs_Weight/weight_diff_ccdf_by_module.pdf}
%     \caption{CCDFs of channel-wise $\ell_2$ weight differences by module, which exhibit moderately heavy tails.}   
%     % \caption{
%     % Empirical CCDFs of channel-wise $\ell_2$ weight differences between Qwen2.5-Math-7B-Instruct and Qwen2.5-7B-Instruct, grouped by module type.
%     % Weight differences exhibit moderately heavy-tailed distributions across modules.
%     % }
%     \label{fig:weight_ccdf_by_module}
% \end{figure}

\begin{table*}[htbp]
\centering
\small
\setlength{\tabcolsep}{4pt}
\renewcommand{\arraystretch}{1.15}
\resizebox{\textwidth}{!}{
\begin{tabular}{lccccccccccc ccccccccccc}
\toprule
& \multicolumn{11}{c}{\textbf{Math}} & \multicolumn{11}{c}{\textbf{Science}} \\
\cmidrule(lr){2-12}\cmidrule(lr){13-23}
\textbf{Metric} 
& EN & AR & ZH & JA & KO & BN & HU & SW & TE & TH & VI
& EN & AR & ZH & JA & KO & BN & HU & SW & TE & TH & VI \\
\midrule
Overlap (\%) 
& 4.66 & 5.62 & 2.68 & 9.80 & 9.74 & 18.72 & 21.69 & 20.53 & 15.67 & 5.81 & 10.28
& 5.35 & 3.83 & 3.44 & 11.84 & 9.74 & 14.40 & 21.06 & 4.11 & 8.32 & 5.42 & 8.26 \\
\#Overlap Ch. 
& 816 & 983 & 469 & 1716 & 1705 & 3277 & 3796 & 3593 & 2743 & 1017 & 1799
& 936 & 670 & 603 & 2073 & 1705 & 2520 & 3686 & 720 & 1457 & 949 & 1446 \\
\bottomrule
\end{tabular}
}
\caption{
Weight--activation top-1\% channel-mask overlap (\%) on \textbf{Qwen2.5-7B} (\textbf{Math-Instruct vs.\ Instruct}).
We report overlap ratios and the number of overlapping channels for each language and domain.
}
\label{tab:app_f5_qwen7b_math}
\end{table*}

\begin{table*}[htbp]
\centering
\small
\setlength{\tabcolsep}{4pt}
\renewcommand{\arraystretch}{1.15}
\resizebox{\textwidth}{!}{
\begin{tabular}{lccccccccccc ccccccccccc}
\toprule
& \multicolumn{11}{c}{\textbf{Math}} & \multicolumn{11}{c}{\textbf{Science}} \\
\cmidrule(lr){2-12}\cmidrule(lr){13-23}
\textbf{Metric}
& AR & BN & EN & HU & JA & KO & SW & TE & TH & VI & ZH
& AR & BN & EN & HU & JA & KO & SW & TE & TH & VI & ZH \\
\midrule
Overlap (\%)
& 10.36 & 12.43 & 5.00 & 25.15 & 5.76 & 5.49 & 17.05 & 10.54 & 9.55 & 7.15 & 3.94
& 8.92 & 14.35 & 5.34 & 25.54 & 6.08 & 6.55 & 14.30 & 11.69 & 8.60 & 8.55 & 4.25 \\
\#Overlap Ch.
& 1813 & 2176 & 876 & 4402 & 1009 & 961 & 2985 & 1845 & 1672 & 1252 & 690
& 1561 & 2512 & 935 & 4470 & 1065 & 1146 & 2503 & 2046 & 1505 & 1496 & 744 \\
\bottomrule
\end{tabular}
}
\caption{
Weight--activation top-1\% channel-mask overlap (\%) on \textbf{Qwen2.5-7B} (\textbf{Coder-Instruct vs.\ Instruct}).
We report overlap ratios and the number of overlapping channels for each language and domain.
}
\label{tab:app_f5_qwen7b_coder}
\end{table*}

\begin{table*}[htbp]
\centering
\small
\setlength{\tabcolsep}{4pt}
\renewcommand{\arraystretch}{1.15}
\resizebox{\textwidth}{!}{
\begin{tabular}{lccccccccccc ccccccccccc}
\toprule
& \multicolumn{11}{c}{\textbf{Math}} & \multicolumn{11}{c}{\textbf{Science}} \\
\cmidrule(lr){2-12}\cmidrule(lr){13-23}
\textbf{Metric} 
& EN & AR & ZH & JA & KO & BN & HU & SW & TE & TH & VI
& EN & AR & ZH & JA & KO & BN & HU & SW & TE & TH & VI \\
\midrule
Overlap (\%) 
& 6.85 & 5.44 & 7.78 & 4.97 & 5.62 & 6.39 & 9.03 & 9.29 & 6.36 & 4.00 & 4.42
& 7.15 & 5.51 & 7.58 & 5.69 & 5.71 & 6.68 & 9.89 & 7.83 & 6.70 & 4.14 & 4.95 \\
\#Overlap Ch.
& 607 & 482 & 689 & 440 & 498 & 566 & 800 & 823 & 564 & 354 & 392
& 634 & 488 & 672 & 504 & 506 & 592 & 876 & 694 & 594 & 367 & 439 \\
\bottomrule
\end{tabular}
}
\caption{
Weight--activation top-1\% channel-mask overlap (\%) on \textbf{Qwen2.5-1.5B} (\textbf{Math-Instruct vs.\ Instruct}).
We report overlap ratios and the number of overlapping channels for each language and domain.
}
\label{tab:app_f5_qwen15b_math}
\end{table*}

\begin{table*}[htbp]
\centering
\small
\setlength{\tabcolsep}{4pt}
\renewcommand{\arraystretch}{1.15}
\resizebox{\textwidth}{!}{
\begin{tabular}{lccccccccccc ccccccccccc}
\toprule
& \multicolumn{11}{c}{\textbf{Math}} & \multicolumn{11}{c}{\textbf{Science}} \\
\cmidrule(lr){2-12}\cmidrule(lr){13-23}
\textbf{Metric}
& AR & BN & EN & HU & JA & KO & SW & TE & TH & VI & ZH
& AR & BN & EN & HU & JA & KO & SW & TE & TH & VI & ZH \\
\midrule
Overlap (\%)
& 5.76 & 7.72 & 4.89 & 7.23 & 6.99 & 8.02 & 6.61 & 7.70 & 6.17 & 5.59 & 7.73
& 5.60 & 8.14 & 5.42 & 7.26 & 7.15 & 8.09 & 6.09 & 8.10 & 6.21 & 6.06 & 7.65 \\
\#Overlap Ch.
& 510 & 684 & 433 & 641 & 619 & 711 & 586 & 682 & 547 & 495 & 685
& 496 & 721 & 480 & 643 & 634 & 717 & 540 & 718 & 550 & 537 & 678 \\
\bottomrule
\end{tabular}
}
\caption{
Weight--activation top-1\% channel-mask overlap (\%) on \textbf{Qwen2.5-1.5B} (\textbf{Coder-Instruct vs.\ Instruct}).
We report overlap ratios and the number of overlapping channels for each language and domain.
}
\label{tab:app_f5_qwen15b_coder}
\end{table*}

\subsection{Additional Results for Finding~5}
\label{app:finding5}

This appendix provides full results for Finding~5 by reporting, across model families and abilities, the overlap between
(i) the \emph{activation-based} top-1\% channel masks (ranked by token-averaged cross-model activation differences) and
(ii) the \emph{weight-based} top-1\% channel masks (ranked by channel-wise $\ell_2$ weight updates).
Throughout, overlaps are measured as the fraction of shared channels (in \%) between the two top-1\% masks, together with the number of overlapping channels.

\paragraph{Weight-Based Channel Update Distributions.}
Figure~\ref{fig:weight_ccdf_by_structure} visualizes the complementary cumulative distribution functions (CCDFs) of channel-wise $\ell_2$ weight differences between Qwen2.5-7B Math-Instruct and Instruct models.
Consistent with the main text, weight updates also exhibit heavy-tailed behavior, but with substantially weaker concentration compared to activation-based differences.

\paragraph{Qwen2.5 (7B and 1.5B): multilingual results.}
Tables~\ref{tab:app_f5_qwen7b_math}--\ref{tab:app_f5_qwen15b_coder} report results for Qwen2.5 across 11 languages and two domains (math/science), for both Math-Instruct and Coder-Instruct fine-tuning.
Across all settings, overlap consistently exceeds the random baseline (1\%), but remains far from complete, indicating that channels with the largest activation shifts are not simply those with the largest weight-magnitude updates.
We also observe substantial variation across languages and domains.
In particular, some languages exhibit noticeably higher alignment (e.g., overlaps reaching $\sim$20\%--25\% in a few cases),
while others remain close to single-digit percentages, suggesting that the relationship between weight updates and activation shifts
is highly ability- and language-dependent.

\paragraph{LLaMA-2-13B family: English-only results.}
Table~\ref{tab:app_f5_llama2_weight_act} reports weight--activation overlap for English math, science, and code abilities
across three LLaMA-2-derived model pairs.
We observe that overlap levels vary substantially across fine-tuning recipes.
In particular, WizardLM-13B and Tulu-2-DPO-13B exhibit moderately higher overlap with weight-based masks, whereas WizardMath-13B shows consistently lower alignment.
This variation suggests that the correspondence between large activation shifts and large weight updates is sensitive to the specific fine-tuning objective and training procedure, even when starting from the same pretrained base model.

\begin{table*}[t]
\centering
\small
\setlength{\tabcolsep}{6pt}
\renewcommand{\arraystretch}{1.15}

\begin{subtable}[t]{0.32\textwidth}
\scriptsize
\centering
\begin{tabular}{lccc}
\toprule
\textbf{Metric} & \textbf{Math} & \textbf{Science} & \textbf{Code} \\
\midrule
Overlap (\%)     & 18.77 & 19.93 & 19.37 \\
\#Overlap Ch.    & 4845  & 5145  & 5000  \\
\bottomrule
\end{tabular}
\caption{Tulu-2-DPO-13B vs.\ LLaMA-2-13B}
\end{subtable}
\hfill
\begin{subtable}[t]{0.32\textwidth}
\scriptsize
\centering
\begin{tabular}{lccc}
\toprule
\textbf{Metric} & \textbf{Math} & \textbf{Science} & \textbf{Code} \\
\midrule
Overlap (\%)     & 30.01 & 36.24 & 32.35 \\
\#Overlap Ch.    & 7747  & 9356  & 8353  \\
\bottomrule
\end{tabular}
\caption{WizardLM-13B vs.\ LLaMA-2-13B}
\end{subtable}
\hfill
\begin{subtable}[t]{0.32\textwidth}
\scriptsize
\centering
\begin{tabular}{lccc}
\toprule
\textbf{Metric} & \textbf{Math} & \textbf{Science} & \textbf{Code} \\
\midrule
Overlap (\%)     & 9.41  & 9.21  & 9.64 \\
\#Overlap Ch.    & 2430  & 2377  & 2489 \\
\bottomrule
\end{tabular}
\caption{WizardMath-13B vs.\ LLaMA-2-13B}
\end{subtable}

\caption{
Weight--activation top-1\% channel-mask overlap (\%) within the \textbf{LLaMA-2-13B} family.
Each subtable reports the overlap between activation-based and weight-based masks for English math, science, and code.
}
\label{tab:app_f5_llama2_weight_act}
\end{table*}

\paragraph{Summary.}
Across both Qwen2.5 and LLaMA-2 families, activation-based ability channels consistently show only partial overlap with channels selected by large weight-magnitude updates.
Although overlaps are reliably above a random baseline, they remain far from complete and vary across languages, abilities, model scales, and fine-tuning recipes.
These results reinforce Finding~5: large cross-model activation differences are not simply explained by weight-update magnitude, indicating that activation-based localization captures complementary and more functionally specific information about ability-related model behavior.

\end{document}